\documentclass[sigconf]{acmart}

\usepackage{hyperref}
\usepackage{url}
\usepackage[utf8]{inputenc} % allow utf-8 input
\usepackage[T1]{fontenc}    % use 8-bit T1 fonts
\usepackage{booktabs}       % professional-quality tables
\usepackage{amsfonts}       % blackboard math symbols
\usepackage{nicefrac}       % compact symbols for 1/2, etc.
\usepackage{microtype}      % microtypography
\usepackage{xcolor}         % colors
\usepackage{graphicx}
\usepackage{wrapfig}
\usepackage{multirow}
\usepackage{epsfig}
\usepackage{amsmath}
\usepackage{subfigure}
\usepackage{subcaption}
\usepackage{marvosym}
\usepackage{color}
\usepackage{threeparttable}
\usepackage{algorithm}
\usepackage{algpseudocode}
\usepackage{setspace}
\usepackage{amsthm}
\usepackage{dutchcal}
\usepackage{makecell}
\usepackage{pifont}
\usepackage{bm}
\usepackage{ bbold }

\usepackage{tcolorbox}
\usepackage{colortbl}
\usepackage{geometry}

\newcommand{\boldres}[1]{{\textbf{\textcolor{red}{#1}}}}
\newcommand{\secondres}[1]{{\underline{\textcolor{blue}{#1}}}}

\AtBeginDocument{%
  }

\setcopyright{acmlicensed}
\copyrightyear{2025}
\acmYear{2025}
\acmDOI{XXXXXXX.XXXXXXX}
\acmConference[Under Review]{}{}{}
\acmISBN{978-1-4503-XXXX-X/18/06}

\begin{document}

\title{Unify and Anchor: A Context-Aware Transformer for Cross-Domain Time Series Forecasting}

% \author{Xiaobin Hong}
% \authornote{Equal contribution.}
% \affiliation{%
% } State Key Laboratory for Novel Software Technology

\author{Xiaobin Hong\textsuperscript{*}}
\affiliation{%
  \institution{Nanjing University}
  \city{Nanjing}
  \country{China}}
\email{xiaobinhong@smail.nju.edu.cn}

\author{Jiawen Zhang\textsuperscript{*}}
\affiliation{%
  \institution{HKUST(GZ)}
  \city{Guangzhou}
  \country{China}
}
\email{jiawe.zh@gmail.com}

\author{Wenzhong Li\textsuperscript{\textdagger}}
\affiliation{%
 \institution{Nanjing University}
 \city{Nanjing}
 \country{China}}
\email{lwz@nju.edu.cn}

\author{Sanglu Lu}
\affiliation{%
  \institution{Nanjing University}
  \city{Nanjing}
  \country{China}}
\email{sanglu@nju.edu.cn}

\author{Jia Li\textsuperscript{\textdagger}}
\affiliation{%
  \institution{HKUST(GZ)}
  \city{Guangzhou}
  \country{China}}
\email{jialee@ust.hk}

\thanks{* Equal contribution.}
\thanks{\textdagger Corresponding authors.}

% \affiliation{%

% \affiliation{%


\begin{abstract}
The rise of foundation models has revolutionized natural language processing and computer vision, yet their best practices to time series forecasting remains underexplored. Existing time series foundation models often adopt methodologies from these fields without addressing the unique characteristics of time series data. In this paper, we identify two key challenges in cross-domain time series forecasting: the complexity of temporal patterns and semantic misalignment. To tackle these issues, we propose the “Unify and Anchor” transfer paradigm, which disentangles frequency components for a unified perspective and incorporates external context as domain anchors for guided adaptation. Based on this framework, we introduce ContexTST, a Transformer-based model that employs a time series coordinator for structured representation and the Transformer blocks with a context-informed mixture-of-experts mechanism for effective cross-domain generalization. Extensive experiments demonstrate that ContexTST advances state-of-the-art forecasting performance while achieving strong zero-shot transferability across diverse domains.
% \footnote{The code for this project will be made open source.}
\end{abstract}




\maketitle

\setlength{\leftmargini}{10pt}

\section{Introduction}

\begin{figure*}[th]
    \centering
    \includegraphics[width=\linewidth]{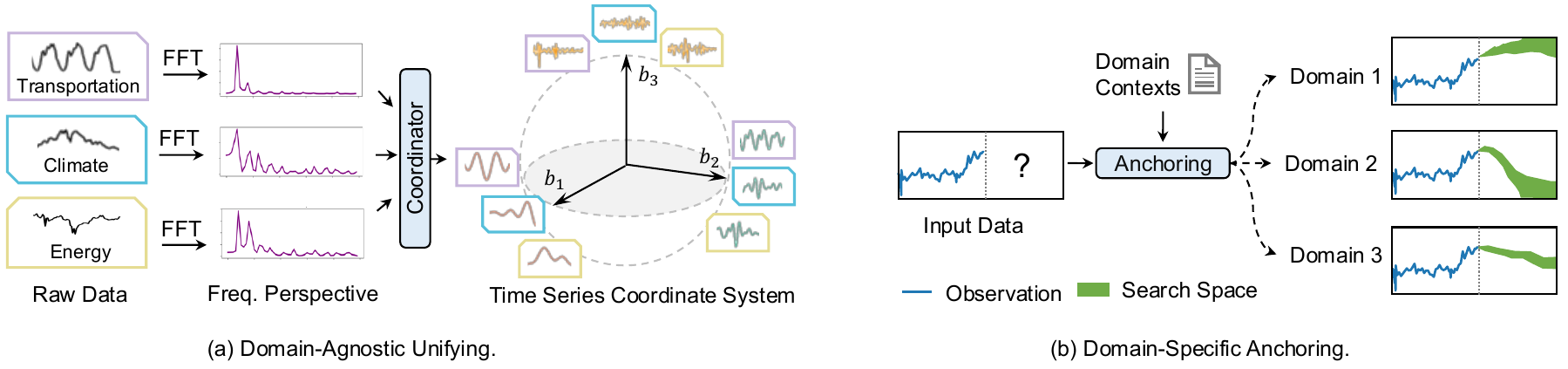}
    \vspace{-4mm}
    % \captionsetup{aboveskip=5pt}
    \caption{The \emph{Unify and Anchor} cross-domain transfer paradigm. (a) An FFT-based coordinator decomposes time series into frequency bases, establishing a unified representation that captures shared structures across domains. (b) Domain-specific contexts serve as anchors, guiding adaptation by constraining the prediction search space to align with domain characteristics.}
    \vspace{-3mm}
    \label{fig:motivation}
\end{figure*}

The rapid advancement of artificial general intelligence (AGI) has positioned foundation models as a key research direction~\cite{bommasani2021foundationmodel}, with the pre-train and transfer paradigm achieving remarkable success in natural language processing (NLP)~\cite{dubey2024llama, liu2024deepseek, brown2020gpt3, touvron2023llama} and computer vision (CV)~\cite{kirillov2023segment, Brooks2024sora, rombach2022high}. This paradigm enables models trained on large-scale datasets to generalize across tasks and domains with minimal fine-tuning or even zero-shot adaptation. However, despite these advancements, its application to time series forecasting—a crucial task in finance, transportation, retail, energy, and climate science~\cite{hou2021finance_ts, lv2014traffic_ts, smith1997traffic_ts, bose2017retail_ts, wang2019renew_energy_ts, mudelsee2019climate_ts}—remains largely unexplored, particularly in the context of cross-domain transfer.

While recent efforts have introduced time series foundation models (TSFMs), many simply repurpose methodologies from NLP~\cite{ansari2024chronos,rasul2023lag,das2023decoder,gao2024units,liu2024timer} and CV~\cite{liu2024moirai,goswami2024moment,ekambaram2024ttms} without accounting for the distinct nature of time series data. However, these direct adaptations are often suboptimal. Unlike text and images, time series lack inherent semantics and exhibit unbounded value ranges, making conventional representation learning techniques less effective in capturing their underlying patterns. Moreover, many time series foundation models rely on multi-domain pre-training with mixed datasets to enhance generalization. Yet, naively merging heterogeneous time series without proper alignment can introduce domain conflicts, leading to negative transfer~\cite{xu2024specialized, zhang2022negative_transfer}. For example, stock market data exhibit irregular, event-driven fluctuations, whereas meteorological data follow periodic seasonal cycles. Training a single model on both domains without explicitly addressing these discrepancies can degrade predictive performance, as the model fails to differentiate domain-specific temporal structures.

% challenge 1: 序列内部模式的多样性与复杂性
To bridge this gap, we identify two key challenges in cross-domain time series forecasting. \textbf{The first challenge is the complexity of temporal patterns.} Time series data often contain superimposed periodic components, where different frequency bands encode distinct semantic information~\cite{zhou2022fedformer, wang2024rose} and are often entangled with noise. Furthermore, datasets from different domains exhibit diverse statistical distributions, and the relative importance of low- and high-frequency components varies across applications~\cite{zhang2024not}. This creates a fundamental trade-off: while expressive feature extraction is necessary to capture domain-specific dynamics, excessive reliance on such patterns impairs generalization. Although prior studies have explored cross-domain generalization in time series forecasting~\cite{wang2024rose, deng2024domain, jin2022domain}, they typically assume stationary distributions and shared attributes across domains. These assumptions rarely hold in real-world scenarios, where time series data exhibit high variability and heterogeneity.

% challenge 2: 领域间语义空间的对齐
\textbf{The second challenge is semantic misalignment.} Identical numerical patterns can represent entirely different underlying phenomena for distinct domains, leading to inconsistencies in learned representations. For example, a daily periodic spike in numerical values may indicate peak electricity consumption in the energy sector but signify traffic congestion in transportation. Without domain awareness, a model trained in one domain may misinterpret such patterns when transferred to another, significantly impacting performance. This highlights the necessity of robust domain alignment mechanisms to ensure effective cross-domain adaptation. Recently, increasing efforts have explored incorporating contextual and external knowledge to improve model adaptability~\cite{zhou2023one,jin2023time,liu2024unitime,liu2024taming,williams2024context}. While most existing approaches employ simplistic mechanisms for integrating contextual information, they reveal the vast potential of leveraging external signals to enhance cross-domain forecasting.

% intuition
In response to these challenges, we propose a novel \emph{``Unify and Anchor”} paradigm for cross-domain time series forecasting, as illustrated in Figure~\ref{fig:motivation}. To handle the complexity of temporal patterns across diverse domains, we establish a \emph{unified perspective} that disentangles coupled semantic information. Specifically, we decompose the raw input series in the frequency domain and construct a time series coordinate system based on extracted frequency bases. This decomposition effectively isolates distinct frequency components, enabling the model to capture shared structures across domains while achieving a balance between feature extraction and generalization. Furthermore, to mitigate the issue of semantic misalignment, we incorporate external contextual information as a \emph{domain anchor} to guide model adaptation. By embedding domain-specific biases, this approach reduces the search space for predictions, facilitating more effective cross-domain generalization.

% contribution
Building on the \emph{``Unify and Anchor"} framework, we present \textbf{Contex}t-aware \textbf{T}ime \textbf{S}eries \textbf{T}ransformer (ContexTST), a Transformer variant designed for cross-domain time series forecasting. The architecture begins with a time series coordinator, which decomposes raw sequences into orthogonal frequency components, forming a unified representation. To enrich the inherently numeric data with semantic context, we automatically generate global- and variable-level textual descriptions as domain anchors. In addition to integrating variable-level anchors into the input representation, we introduce a context-informed mixture-of-experts (CI-MoE) mechanism. This module dynamically selects the most suitable expert for encoding data from different domains, further refining adaptation and improving forecasting accuracy. Extensive experiments demonstrate the superiority of ContexTST in comparison with the state-of-the-art baseline models.
% In summary, %our key contributions are:

The key contributions of this paper are summarized as follows:
\begin{itemize}
    \item We introduce the \emph{``Unify and Anchor"} paradigm, an effective framework for cross-domain time series forecasting.
    \item We develop ContexTST, a Transformer-based model that unifies temporal perspectives through a time series coordinator and anchors domain semantics via the context-aware Transformer.
    \item We conduct extensive experiments demonstrating ContexTST’s superior performance and strong zero-shot transferability across diverse domains.
\end{itemize}

\section{Related Works}
\label{sec:related_works}

\textbf{\emph{Deep Time Series Forecasting.}}
Deep learning models have been extensively explored for time series analysis, particularly in forecasting tasks \cite{liu2022scinet, wu2022timesnet, zhao2017lstm, lai2018modeling, oreshkin2019n, das2023long, zhang2023probts}. Transformers~\cite{vaswani2017attention}, as foundational architectures, have gained significant traction in time series forecasting due to their ability to capture long-term dependencies and intricate temporal patterns \cite{patchtst, liu2023itransformer,wen2022transformers}. 
However, while state-of-the-art deep forecasting models demonstrate strong performance on individual datasets, they often fail to generalize to new datasets—even within the same domain—due to overfitting to domain-specific patterns.
To enhance generalization, decomposition-based methods \cite{wang2024timemixer, wang2024timemixer++, wu2021autoformer, deng2024parsimony} apply signal processing techniques to extract intrinsic periodicity and fundamental structures. While effective in seen-domain forecasting, these methods struggle with cross-domain transfer, limiting their applicability to unseen datasets.

\textbf{\emph{Time Series Foundation Models.}}
Inspired by scaling laws in large language models (LLMs) \cite{kaplan2020scaling}, TSFMs have emerged to improve generalized forecasting. These models fall into two main categories: NLP-inspired architectures \cite{ansari2024chronos, rasul2023lag, das2023decoder, liu2024timer}, which treat patched time series as discrete sequences similar to LLaMA~\cite{touvron2023llama} or BERT~\cite{devlin2018bert}, and CV-inspired architectures, which leverage MLP-Mixer-like structures~\cite{tolstikhin2021mlp} or masking-based models~\cite{Brooks2024sora} to capture temporal dependencies and inter-variable correlations.

% limitation
However, these adaptations often overlook the unique characteristics of time series compared to modalities like text or images, failing to incorporate time series–specific inductive biases. Many TSFMs rely on multi-domain pre-training over mixed datasets to enhance generalization, but the lack of inherent semantics in numerical data can cause conflicts and negative transfer. These challenges underscore the need for tailored strategies to enable effective cross-domain transfer in time series forecasting.

\textbf{\emph{Time Series Cross-Domain Transfer.}}
Cross-domain time series transfer requires a model to be pre-trained on a source domain and perform zero-shot inference on an unseen target domain. To enable this, previous studies have explored cross-domain generalization by learning shared patterns inherent in time series data \cite{wang2024rose, deng2024domain, jin2022domain}. However, these approaches often assume time series from different domains share common attributes and experience no abrupt distribution shifts. In practice, these assumptions rarely hold, as real-world time series data exhibit highly dynamic and heterogeneous behaviors, limiting the applicability of such methods.

Another line of research introduces textual modality as semantic enrichment to enhance cross-domain time series forecasting \cite{zhou2023one, jin2023time, liu2024unitime, liu2024taming, liu2024timecma,williams2024context}.  
For example, CIK \cite{williams2024context} establishes a forecasting benchmark with carefully crafted textual context, demonstrating impressive zero-shot transfer capabilities. However, its reliance on extensive labeling and training efforts poses a significant challenge. UniTime \cite{liu2024unitime}, a counterpart to this study, enhances cross-domain generalization by incorporating a leveraging domain prompts during training. This approach primarily emphasizes domain anchoring while overlooking the importance of providing a unified representation for time series across different domains.

To address these gaps, we present the “Unify and Anchor” paradigm, which simultaneously introduces a unified frequency perspective to capture fundamental time series characteristics while integrating contextual information as domain anchors to guide adaptation. In the following sections, we detail ContexTST, our proposed model built upon this framework.

% \clearpage

% \section{Preliminaries}
% \label{sec:preli}

% \textbf{Problem Definition.} In this work, we focus on the multivariate time series cross-domain forecasting task. To this end, we define the observed history time series from domain $\tau$ as $X^{(\tau)}_{1:L}=\{x^{(\tau)}_1, x^{(\tau)}_2, \cdots, x^{(\tau)}_L\} \in \mathbb{R}^{L\times c_{\tau}}$, where $L$ is the look back horizon, and $c_{\tau}$ denotes the number of channels/variates of input time series within domain $\tau$. In the context of cross-domain time series forecasting, the channel numbers $\tau$ vary across domains, and the predicted future time series can be defined as $X^{(\tau)}_{L+1:L+H}=\{x^{(\tau)}_{L+1}, x^{(\tau)}_{L+2}, \cdots, x^{(\tau)}_{L+H}\}\in \mathbb{R}^{H\times c_{\tau}}$, where $H$ is the length of predicted time series.

% \textbf{Global Prompt.} We utilize the dataset-level and task-level textual information as the global prompt for time series cross-domain transferring in the Mixture-of-Experts (MoE) architecture.

% \clearpage

\section{Methodology}
\label{sec:method}

\begin{figure*}
    \centering
    \includegraphics[width=0.98\linewidth]{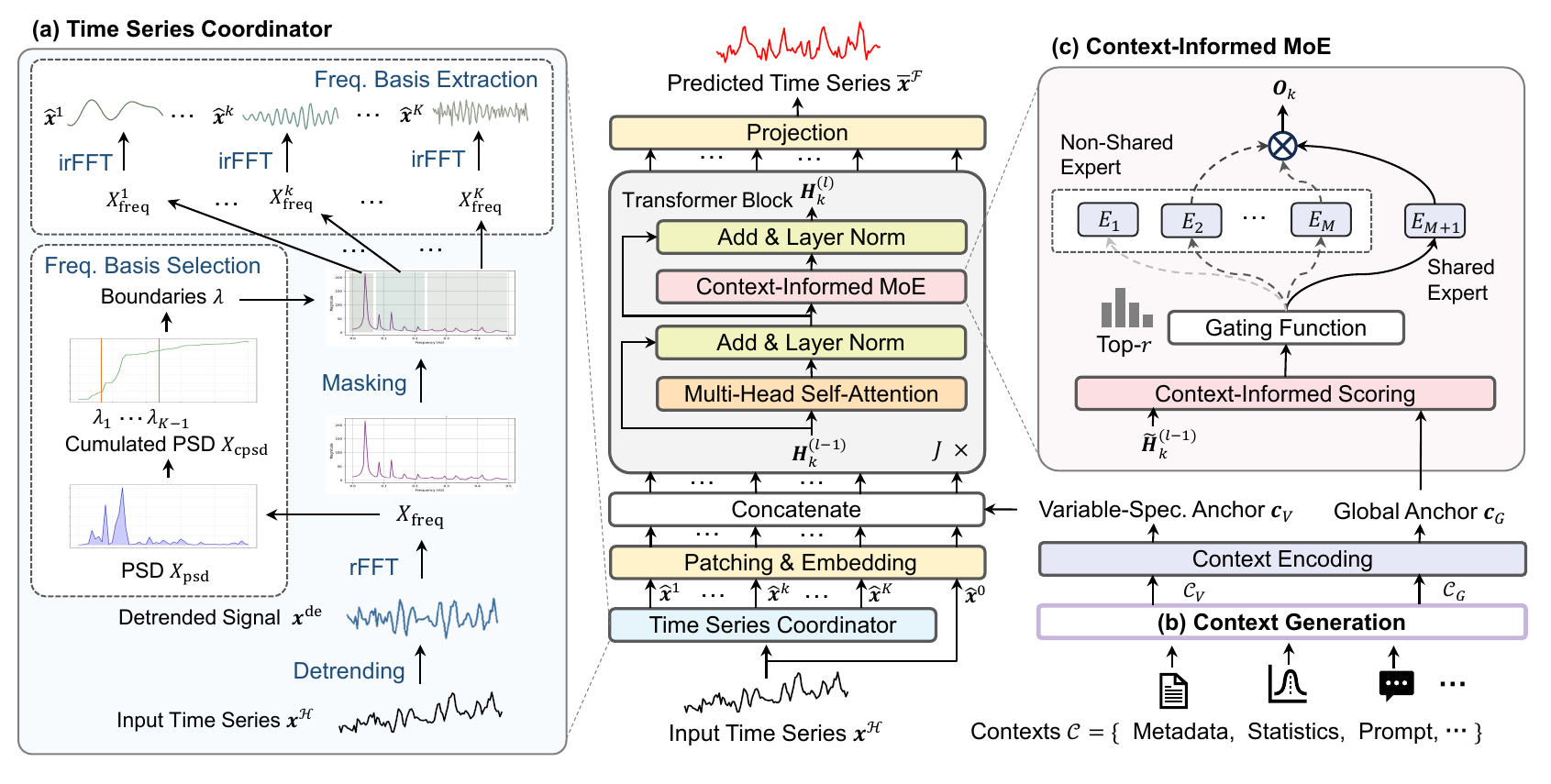}
    \vspace{-5mm}
    \caption{Overview of the proposed ContexTST framework. The architecture begins with the (a) Time Series Coordinator, which decomposes raw time series into orthogonal frequency components for structured representation. To complement temporal features, (b) Context Generation produces global- and variable-level textual descriptions, enriching the time series with external knowledge. Furthermore, the (c) Context-Informed MoE mechanism integrates contextual information, enabling effective generalization across domains.}
    \vspace{-4mm}
    \label{fig:framework}
\end{figure*}

Following the channel-independent setting~\cite{patchtst}, we define a univariate time series as $\bm{x}_{1:T} = \{ x_n \}_{n=1}^{T} $, where $x_n \in \mathbb{R}$.
Given an observation history of length $L$, denoted as $\bm{x}^\mathcal{H}_t = \bm{x}_{t-L:t}$, along with auxiliary contextual information $\mathcal{C}$ of arbitrary nature, the objective of time series forecasting is to is to predict length-$T$ future values $\bm{x}^\mathcal{F}_t = \bm{x}_{t+1:t+T}$:
% \vspace{-2mm}
\begin{align}
    \max_\phi \mathbb{E}_{\bm{x} \sim p(\mathcal{D}), t \sim p(\mathcal{T})} \left[ \log p_\phi (\bm{x}^\mathcal{F}_t | \bm{x}^\mathcal{H}_t, \mathcal{C})  \right],
\end{align}
% \vspace{-2mm}
where $p(\mathcal{D})$ represents the distribution of time series samples, and $p(\mathcal{T})$ defines the distribution from which prediction timestamps are sampled.
In the cross-domain setting, training is performed on a set of source domain datasets $\mathcal{D}_s$, while evaluation is conducted on an unseen target domain dataset $\mathcal{D}_t$.

\textbf{\emph{Overview.}}
Figure~\ref{fig:framework} provides an overview of ContexTST. The framework begins with the Time Series Coordinator (Section~\ref{sec:coordinator}), which decomposes raw time series into orthogonal frequency bases, enabling a structured and unified representation of complex patterns. These decomposed signals are then anchored with contextual information, as detailed in Section~\ref{sec:context_gen}. 
The Context-Aware Transformer (Section~\ref{sec:context-aware-transformer}) processes this combined input for frequency-specific feature extraction, leveraging a Context-Informed Mixture-of-Experts (CI-MoE) to adaptively select domain-specific experts. Finally, all frequency bases are aggregated in the Projection layer (Section~\ref{sec:agg_proj}) and mapped to the final time series prediction.

\subsection{Time Series Coordinator}
\label{sec:coordinator}

The Time Series Coordinator is a key module in the “Unify and Anchor” paradigm. It decomposes time series into multiple orthogonal frequency bases, establishing a unified perspective through a newly constructed coordinate system.

% detrend
\textbf{\emph{Detrending.}} 
To mitigate the influence of amplitude variations in different data domains, we first apply a standard moving-average method \cite{hyndman2018forecasting} to separate the input time series into a trend-cycle signal $\bm{x}^{\textrm{trend}}$ and a detrended signal $\bm{x}^{\textrm{de}}$. Given an observed time series $\bm{x}^\mathcal{H}$, we compute\footnote{For brevity, we omit the prediction timepoint $t$ from $\bm{x}^\mathcal{H}_t$ in the following sections.}:
\begin{align}
    x^{\textrm{de}}_t = x_t - x^{\textrm{trend}}_t, \quad x^{\textrm{trend}}_t = \frac{1}{2\kappa +1} \sum^\kappa_{j=-\kappa} x_{t+j}, \quad x_t \in \bm{x}^\mathcal{H},
\end{align}
where $\kappa$ is the kernel size for local averaging. 

\textbf{\emph{Switching to Frequency Perspective.}}
The detrended signal $\bm{x}^{\textrm{de}}$ contains both high- and low-frequency patterns, which can lead to overfitting to complex fluctuations. Therefore, we apply a spectral transform tool, Real Fast Fourier Transform (rFFT),
% , which is also used in \cite{wang2024rose}, 
to convert the time-domain signal into its frequency-domain representation, enabling a clearer decomposition of frequency components.
\begin{align}
    X_{\textrm{freq}} = \text{rFFT}(\bm{x}^{\textrm{de}}).
\end{align}
Here, $X_{\textrm{freq}} \in \mathbb{C}^{L/2 +1}$ represents the frequency-domain coefficients.

\textbf{\emph{Adaptive Frequency Basis Selection.}}
Different domains exhibit distinct frequency distributions, as illustrated in Figure~\ref{fig:motivation}. To ensure each recomputed frequency basis aligns with the dominant frequencies in the data, we employ Power Spectral Density (PSD)~\cite{stein2000psd}, which quantifies the power of each frequency component, facilitating the identification of key frequency patterns. Given the frequency-domain representation $X_{\text{freq}}$, the PSD is computed as:
\begin{align}
    X_{\text{psd}}[p] =
\begin{cases}
\frac{|X_{\text{freq}}[p]|^2}{L}, & p = 0 \text{ or } p = \frac{L}{2}, \\
\frac{|X_{\text{freq}}[p]|^2}{L/2}, & 1 \leq p \leq \frac{L}{2} - 1.
\end{cases}
\end{align}
where $p=0,1,\cdots,L/2$ denotes the frequency index and $|X_{\text{freq}}[p]|^2$ denotes the squared magnitude of the frequency component at index $p$.
Next, we partition the frequency spectrum based on the cumulative energy percentage, ensuring that each frequency band retains a similar proportion of the total signal energy. The cumulative PSD (CPSD) is then computed as:
\begin{align}
    X_{\text{cpsd}}[p] = \frac{\sum_{j=0}^{p} X_{\text{psd}}[j]}{\sum_{j=0}^{\lfloor L/2\rfloor} X_{\text{psd}}[j]}
\end{align}
where $X_{\text{cpsd}}[p]$ represents the cumulative energy percentage up to frequency index $p$. 
We select $K$ representative frequency components as the basis, with frequency boundaries $\{\lambda_0, \lambda_1, \cdots, \lambda_K\}$ defined as:
\begin{align}
\lambda_k := \arg \min_{p} \{ X_{\text{cpsd}}[p] \geq \frac{k}{K} \}, \quad k=1,\cdots, K-1.
\end{align}
This adaptive partitioning allows each selected frequency component to maintain a proportional contribution to the overall signal structure.

\textbf{\emph{Frequency Basis Extraction.}}
For frequency basis decomposition, we employ a frequency-domain masking approach to isolate distinct frequency components. Specifically, we define a mask matrix  $M^k \in \{0,1\}^{(\frac{L}{2} +1)}$  to extract the $k$-th frequency basis within the range $[\lambda_{k-1}, \lambda_{k})$. The mask values are assigned as:
\begin{align}
m_j^k =
\begin{cases}
1, & \lambda_{k-1} \leq j < \lambda_k, \\
0, & \text{otherwise}.
\end{cases}
\end{align}
Here, $m_j^k$ indicates whether the $j$-th frequency component is retained ($m_j^k=1$) or masked ($m_j^k=0$) for basis $k$.
Using the mask matrix, we extract the $k$-th frequency component from the full frequency-domain representation $x_{\text{freq}}$ as:
\begin{equation}
X_{\text{freq}}^k = X_{\text{freq}} \odot M^k, \quad X^k_{\text{freq}} \in \mathbb{C}^{\frac{L}{2} + 1},
\end{equation}
where $\odot$ is element-wise Hadamard product. This operation isolates the target frequency band while nullifying all other components.

Next, we apply the inverse rFFT (irFFT) to reconstruct the time-domain representation of the $k$-th frequency component:
\begin{align}
\hat{\bm{x}}^k =\text{irFFT}( X_{\text{freq}}^k), \quad \hat{\bm{x}}^k \in \mathbb{R}^{L}.
\end{align}
The resulted $\hat{\bm{x}}^k$ represents the decomposed time series corresponding to the  $k$-th frequency basis.

\textbf{\emph{Patching and Embedding.}}  
Time series data lack meaningful semantics at individual time points, thus we adopt patching techniques~\cite{patchtst}, segmenting the input sequence $\bm{x} \in \mathbb{R}^L$ into non-overlapping patches. 
Given a token length $P$, the input sequence is divided into $N = \left\lceil \frac{L}{P} \right\rceil$ patches, forming $\bm{x}^p \in \mathbb{R}^{N \times P}$. To maintain proper segmentation, we pad the sequence by duplicating its last value before applying the patching operation.  
This process is applied to both the original sequence $\hat{\bm{x}}^0 = \bm{x}^{\mathcal{H}}$ and each decomposed series $\{\hat{\bm{x}}^1, \dots, \hat{\bm{x}}^K\}$, yielding a patched representation $\hat{\bm{X}} \in \mathbb{R}^{(K+1) \times N \times P}$. Finally, a trainable linear projection $\mathbf{H}^{\text{ts}} = \text{Embed}(\hat{\bm{X}})$ is applied, where $\text{Embed}(\cdot): \mathbb{R}^{P} \rightarrow \mathbb{R}^{D}$ maps patches into a latent space of dimension $D$. The resulting representation $\mathbf{H}^{\text{ts}} \in \mathbb{R}^{(K+1) \times N \times D}$ forms the time series tokens (TS-tokens) for subsequent model processing.

\subsection{Context Generation \& Encoding}
\label{sec:context_gen}

Contextual information acts as a domain anchor, enhancing cross-domain generalization in time series forecasting. Since many time series datasets lack textual descriptions, we propose an automated method to generate them, capturing key aspects of the underlying processes and providing complementary knowledge akin to expert annotations or external information sources. 

\textbf{\emph{Context Generation}.} 
We generate two levels of natural language descriptions: \emph{Global-Level Description} ($\mathcal{C}_{G}$) and \emph{Variable-Level Description} ($\mathcal{C}_V$), with examples provided in Appendix~\ref{app:context_generate}.  
The global-level description $\mathcal{C}_{G}$ offers a comprehensive summary of the dataset and task, including key metadata such as domain type, sampling frequency, timestamp range, number of variables, and task-specific instructions.  
The variable-level description $\mathcal{C}_{V}$ provides feature-specific explanations for individual time series variables. To generate these descriptions, we query a general-purpose large language model (LLM) using a structured prompt incorporating metadata from $\mathcal{C}_{G}$ and variable names. We use GPT-4o in our experiments for this purpose. Additionally, we compute basic statistical features (e.g., minimum, maximum, mean, and median) from the training data in each domain and append them to variable descriptions, enriching contextual information with domain-specific insights.

\textbf{\emph{Context Encoding.}}
We encode textual descriptions using a pre-trained large language model (LLM). Specifically, we employ Sentence-BERT~\cite{reimers2019sentence} to convert the global and variable-level descriptions into numerical representations: 
\vspace{-1mm}
\begin{align}
\mathbf{c}_{G} = \text{LLM}(\mathcal{C}_{G}) \in \mathbb{R}^{D_{C}}, \quad
\mathbf{c}_{V} = \text{LLM}(\mathcal{C}_{V}) \in \mathbb{R}^{D_{C}},
\end{align}
\vspace{-1mm}
where $\mathbf{c}_{G}$ serves as the global context anchor, and $\mathbf{c}_{V}$ represents the variable-specific context anchors.

For variate-level anchoring, we concatenate $\mathbf{c}_{V}$ with the time series tokens. At the input stage, the variable-level context anchor $\mathbf{c}_{V}$ is replicated across all $K+1$ components to form the context matrix $\mathbf{C}_{V} \in \mathbb{R}^{(K+1)\times D_C}$. Since the context anchors and TS tokens originate from different modalities with potentially mismatched latent dimensions, we introduce a modality alignment module $\text{Align}(\cdot)$, consisting of two linear layers with an activation function:  
\begin{align}
    \mathbf{H} = \text{Concat}(\mathbf{H}^\text{ts}, \text{Align}(\mathbf{C}_V)), \quad \mathbf{H} \in \mathbb{R}^{(K+1) \times (N+1) \times D}.
\end{align}
The global context anchor $\mathbf{c}_{G}$ is integrated into the Context-Aware Transformer, which is detailed in Section \ref{sec:context-aware-transformer}.

\subsection{Context-Aware Transformer}
\label{sec:context-aware-transformer}

The Context-Aware Transformer follows an encoder-only architecture. Unlike the standard Transformer, it incorporates Component-Wise Attention and a Context-Informed Mixture-of-Experts to enhance adaptability in cross-domain transfer scenarios. These mechanisms are fundamental to the anchoring process, enabling more efficient and flexible domain adaptation.

\textbf{\emph{Component-Wise Attention}}
The component-wise attention mechanism operates on the patch embeddings of original and decomposed time-series signals, processing each component independently. This design allows the model to learn component-specific representations without interference from other frequency bands, enhancing its ability to capture distinct temporal patterns.
For the $k$-th component, denoted as $\mathbf{H}_{k} \in \mathbb{R}^{(N+1) \times D}$, where $\mathbf{H}$ represents the full set of patched embeddings, we apply multi-head self-attention ($\textrm{MHSA} (\cdot)$) within a post layer normalization framework:
\begin{equation}
\tilde{\mathbf{H}}^{(l-1)}_{k} = \textrm{LayerNorm} \left( \mathbf{H}^{(l-1)}_{k} + \textrm{MHSA} (\mathbf{H}^{(l-1)}_{k}) \right),
\end{equation}
where $l \in \{1,\dots,J\}$ denotes the $l$-th Transformer block.

\textbf{\emph{Context-Informed Mixture-of-Experts.}}
To enhance generalization and mitigate overfitting, we introduce the Context-Informed Mixture-of-Experts (CI-MoE) within the context-aware Transformer block. CI-MoE replaces the standard feedforward network (FFN) with a MoE layer that dynamically selects specialized experts based on contextual information.  

The CI-MoE layer comprises $M$ domain-specific expert networks $\{ E_1, \dots, E_M\}$, which capture domain-dependent patterns, and a shared expert $E_{M+1}$, which learns generalizable knowledge across contexts. Each expert is implemented as a two-layer multilayer perceptron (MLP).  
A gating function $g(\cdot)$ dynamically allocates experts by taking both the component representations $\tilde{\bm{H}}^{(l-1)}_{k}$ and the global context anchor $\mathbf{C}_G$ as input. It activates the top-$r$ experts with the highest scores:  
\begin{equation}
    g(\tilde{\bm{H}}^{(l-1)}_{k},\mathbf{C}_G ) = \text{softmax}\left(\text{Top-}r( f( \tilde{\bm{H}}^{(l-1)}_{k}, \mathbf{C}_G))\right).
\end{equation} 
Here, $f(\cdot)$ is the context-informed scoring function, which assigns a score to each expert based on both the global context and time series representations. We define $f(\bm{H}, \bm{C}) = (\text{Align}(\bm{C}) W^g)^\top \bm{H}$, where $W^g \in \mathbb{R}^{1 \times E}$ is a learnable parameter matrix. 
Following prior works~\cite{lepikhin2020gshard,rajbhandari2022deepspeed,liu2024moirai}, we set the number of activated experts to $r=2$.

The CI-MoE layer aggregates the outputs of the selected experts along with the shared expert $E_{M+1}$ and non-shared experts:
\begin{equation}
    \bm{O}_{k} = E_{M+1}(\tilde{\bm{H}}^{(l-1)}_{k}) + \sum_{e=1}^M g(\tilde{\bm{H}}^{(l-1)}_{k},\mathbf{C}_G ) E_e(\tilde{\bm{H}}^{(l-1)}_{k}).
\end{equation}

To stabilize training and facilitate gradient flow, we apply a residual connection followed by layer normalization:
\begin{equation}
    \bm{H}^{(l)}_{k} = \text{LayerNorm}\left( \bm{O}_{k} + \tilde{\bm{H}}^{(l-1)}_{k} \right).
\end{equation}

\subsection{Projection}
\label{sec:agg_proj}

To generate the final time series prediction, we integrate the decomposed signal components and map them to the forecast output. Given the hidden states $\bm{H}^{(J)}_k$ for each component, where $k = \{ 0, 1, \dots, K\}$, we apply mean aggregation to obtain a unified representation $\bar{\bm{H}}$. 
% We use a linear projector to map $\bar{\bm{H}}$ to final length-$T$ forecast:
The aggregated representation $\bar{\bm{H}}$ is then passed through a projection head, a learnable transformation function that maps it to the final forecast:  
\begin{equation}
\hat{\bm{x}}^{\mathcal{F}} = \text{Projection}(\bar{\bm{H}}), \quad \hat{\bm{x}}^{\mathcal{F}} \in \mathbb{R}^{T},
\end{equation}

\subsection{Optimization}
\label{sec:optimization}

We employ a Huber loss~\cite{huber1992robust} for prediction error minimization. The Huber loss provides robustness to outliers while maintaining training stability, combining the benefits of both L1 and L2 loss functions:
\begin{align}
\mathcal{L}_{\text{pred}} &= \frac{1}{T} \sum^{T}_{t=1} \mathcal{L}_{\text{huber}} (x^{\mathcal{F}}_t, \bar{x}^{\mathcal{F}}_t), \\
    \mathcal{L}_{\text{huber}} (x^{\mathcal{F}}_t, \bar{x}^{\mathcal{F}}_t) &= 
    \begin{cases}
    \frac{1}{2} (x^{\mathcal{F}} - \bar{x}^{\mathcal{F}})^2, & \text{if } |x^{\mathcal{F}} - \bar{x}^{\mathcal{F}}| \leq \delta, \\
    \delta |x^{\mathcal{F}} - \bar{x}^{\mathcal{F}}| - \frac{1}{2} \delta^2 , & \text{otherwise},
    \end{cases}
\end{align}
where $\delta$ is a hyperparameter that controls the transition between quadratic loss (for small errors) and linear loss (for large errors).

\begin{table*}[th]
  \caption{The average forecasting results from 4 different prediction length \{96, 192, 336, 720\}. The input length is set to 96. The best results are highlighted in \textcolor{red}{red}, and the second best is in \textcolor{blue}{blue}. Full results are reported in Table~\ref{tab:full_forecasting_results}.}
  \vspace{-3mm}
  \label{tab:brif_indomain_forecasting}
  \vskip 0.05in
  \centering
  \resizebox{1.0\textwidth}{!}{
  \begin{threeparttable}
  \begin{small}
  \renewcommand{\multirowsetup}{\centering}
  \setlength{\tabcolsep}{1pt}
  \begin{tabular}{c|cc|cc|cc|cc|cc|cc|cc|cc|cc|cc|cc}
    \toprule
    \multirow{2}{*}{Models} &
    \multicolumn{2}{c}{\rotatebox{0}{\scalebox{0.8}{\textbf{ContexTST}}}} &
    \multicolumn{2}{c}{\rotatebox{0}{\scalebox{0.8}{TimeXer}}} &
    \multicolumn{2}{c}{\rotatebox{0}{\scalebox{0.8}{TimeMixer}}} &
    \multicolumn{2}{c}{\rotatebox{0}{\scalebox{0.8}{iTransformer}}} &
    \multicolumn{2}{c}{\rotatebox{0}{\scalebox{0.8}{GPT4TS}}} &
    \multicolumn{2}{c}{\rotatebox{0}{\scalebox{0.8}{PatchTST}}} & 
    \multicolumn{2}{c}{\rotatebox{0}{\scalebox{0.8}{DLinear}}} &
    \multicolumn{2}{c}{\rotatebox{0}{\scalebox{0.8}{MICN}}} & 
    \multicolumn{2}{c}{\rotatebox{0}{\scalebox{0.8}{TimesNet}}} & 
    \multicolumn{2}{c}{\rotatebox{0}{\scalebox{0.8}{FEDformer}}} &
    \multicolumn{2}{c}{\rotatebox{0}{\scalebox{0.8}{Autoformer}}} \\
    
     & \multicolumn{2}{c}{\scalebox{0.8}{(\textbf{Ours})}} &
    \multicolumn{2}{c}{\scalebox{0.8}{\citeyear{wang2024timexer}}} &
    \multicolumn{2}{c}{\scalebox{0.8}{\citeyear{wang2024timemixer}}}&
    \multicolumn{2}{c}{\scalebox{0.8}{\citeyear{liu2023itransformer}}}&
    \multicolumn{2}{c}{\scalebox{0.8}{\citeyear{zhou2023one}}}&
    \multicolumn{2}{c}{\scalebox{0.8}{\citeyear{patchtst}}}&
    \multicolumn{2}{c}{\scalebox{0.8}{\citeyear{dlinear}}}&
    \multicolumn{2}{c}{\scalebox{0.8}{\citeyear{wang2023micn}}}&
    \multicolumn{2}{c}{\scalebox{0.8}{\citeyear{wu2022timesnet}}}&
    \multicolumn{2}{c}{\scalebox{0.8}{\citeyear{zhou2022fedformer}}}&
    \multicolumn{2}{c}{\scalebox{0.8}{\citeyear{wu2021autoformer}}}
    \\
    
    \cmidrule(lr){2-3} \cmidrule(lr){4-5}\cmidrule(lr){6-7} \cmidrule(lr){8-9}\cmidrule(lr){10-11}\cmidrule(lr){12-13}\cmidrule(lr){14-15}\cmidrule(lr){16-17}\cmidrule(lr){18-19} \cmidrule(lr){20-21} \cmidrule(lr){22-23}
    Metric & \scalebox{0.78}{MSE} & \scalebox{0.78}{MAE} & \scalebox{0.78}{MSE} & \scalebox{0.78}{MAE} & \scalebox{0.78}{MSE} & \scalebox{0.78}{MAE} & \scalebox{0.78}{MSE} & \scalebox{0.78}{MAE} & \scalebox{0.78}{MSE} & \scalebox{0.78}{MAE} & \scalebox{0.78}{MSE} & \scalebox{0.78}{MAE} & \scalebox{0.78}{MSE} & \scalebox{0.78}{MAE} & \scalebox{0.78}{MSE} & \scalebox{0.78}{MAE} & \scalebox{0.78}{MSE} & \scalebox{0.78}{MAE} & \scalebox{0.78}{MSE} & \scalebox{0.78}{MAE} & \scalebox{0.78}{MSE} & \scalebox{0.78}{MAE}\\
    \toprule

    \scalebox{0.85}{ETTh1} & \boldres{\scalebox{0.85}{0.430}}&\boldres{\scalebox{0.85}{0.431}}&\secondres{\scalebox{0.85}{0.437}}&\secondres{\scalebox{0.85}{0.437}}&\scalebox{0.85}{0.447}&\scalebox{0.85}{0.441}&\scalebox{0.85}{0.454}&\scalebox{0.85}{0.448}&\scalebox{0.85}{0.457}&\scalebox{0.85}{0.450}&\scalebox{0.85}{0.469}&\scalebox{0.85}{0.455}&\scalebox{0.85}{0.456}&\scalebox{0.85}{0.453}&\scalebox{0.85}{0.475}&\scalebox{0.85}{0.481}&\scalebox{0.85}{0.461}&\scalebox{0.85}{0.450}&\scalebox{0.85}{0.498}&\scalebox{0.85}{0.484}&\scalebox{0.85}{0.496}&\scalebox{0.85}{0.487} \\

    \scalebox{0.85}{ETTh2} & \boldres{\scalebox{0.85}{0.365}}&\boldres{\scalebox{0.85}{0.392}}&\scalebox{0.85}{0.370}&\scalebox{0.85}{0.398}&\secondres{\scalebox{0.85}{0.365}}&\secondres{\scalebox{0.85}{0.395}}&\scalebox{0.85}{0.383}&\scalebox{0.85}{0.407}&\scalebox{0.85}{0.398}&\scalebox{0.85}{0.416}&\scalebox{0.85}{0.388}&\scalebox{0.85}{0.410}&\scalebox{0.85}{0.559}&\scalebox{0.85}{0.515}&\scalebox{0.85}{0.574}&\scalebox{0.85}{0.532}&\scalebox{0.85}{0.414}&\scalebox{0.85}{0.427}&\scalebox{0.85}{0.436}&\scalebox{0.85}{0.449}&\scalebox{0.85}{0.450}&\scalebox{0.85}{0.459}\\

    \scalebox{0.85}{ETTm1} & \boldres{\scalebox{0.85}{0.380}}&\boldres{\scalebox{0.85}{0.394}}&\scalebox{0.85}{0.382}&\scalebox{0.85}{0.403}&\secondres{\scalebox{0.85}{0.381}}&\secondres{\scalebox{0.85}{0.396}}&\scalebox{0.85}{0.409}&\scalebox{0.85}{0.420}&\scalebox{0.85}{0.396}&\scalebox{0.85}{0.401}&\scalebox{0.85}{0.398}&\scalebox{0.85}{0.411}&\scalebox{0.85}{0.403}&\scalebox{0.85}{0.417}&\scalebox{0.85}{0.423}&\scalebox{0.85}{0.422}&\scalebox{0.85}{0.400}&\scalebox{0.85}{0.420}&\scalebox{0.85}{0.448}&\scalebox{0.85}{0.452}&\scalebox{0.85}{0.588}&\scalebox{0.85}{0.517}\\

    \scalebox{0.85}{ETTm2} & \secondres{\scalebox{0.85}{0.279}}&\secondres{\scalebox{0.85}{0.323}}&\boldres{\scalebox{0.85}{0.274}}&\boldres{\scalebox{0.85}{0.322}}&\scalebox{0.85}{0.280}&\scalebox{0.85}{0.326}&\scalebox{0.85}{0.288}&\scalebox{0.85}{0.332}&\scalebox{0.85}{0.294}&\scalebox{0.85}{0.339}&\scalebox{0.85}{0.281}&\scalebox{0.85}{0.326}&\scalebox{0.85}{0.350}&\scalebox{0.85}{0.408}&\scalebox{0.85}{0.353}&\scalebox{0.85}{0.402}&\scalebox{0.85}{0.291}&\scalebox{0.85}{0.333}&\scalebox{0.85}{0.304}&\scalebox{0.85}{0.349}&\scalebox{0.85}{0.327}&\scalebox{0.85}{0.371}\\

    \scalebox{0.85}{Electricity} & \boldres{\scalebox{0.85}{0.176}}&\boldres{\scalebox{0.85}{0.270}}&\scalebox{0.85}{0.188}&\scalebox{0.85}{0.284}&\secondres{\scalebox{0.85}{0.182}}&\secondres{\scalebox{0.85}{0.273}}&\scalebox{0.85}{0.189}&\scalebox{0.85}{0.282}&\scalebox{0.85}{0.217}&\scalebox{0.85}{0.307}&\scalebox{0.85}{0.216}&\scalebox{0.85}{0.304}&\scalebox{0.85}{0.217}&\scalebox{0.85}{0.300}&\scalebox{0.85}{0.196}&\scalebox{0.85}{0.309}&\scalebox{0.85}{0.193}&\scalebox{0.85}{0.295}&\scalebox{0.85}{0.213}&\scalebox{0.85}{0.327}&\scalebox{0.85}{0.227}&\scalebox{0.85}{0.338}\\

    \scalebox{0.85}{Weather} & \boldres{\scalebox{0.85}{0.240}}&\boldres{\scalebox{0.85}{0.266}}&\scalebox{0.85}{0.241}&\scalebox{0.85}{0.272}&\secondres{\scalebox{0.85}{0.240}}&\secondres{\scalebox{0.85}{0.272}}&\scalebox{0.85}{0.258}&\scalebox{0.85}{0.278}&\scalebox{0.85}{0.279}&\scalebox{0.85}{0.297}&\scalebox{0.85}{0.259}&\scalebox{0.85}{0.280}&\scalebox{0.85}{0.265}&\scalebox{0.85}{0.317}&\scalebox{0.85}{0.268}&\scalebox{0.85}{0.321}&\scalebox{0.85}{0.262}&\scalebox{0.85}{0.287}&\scalebox{0.85}{0.309}&\scalebox{0.85}{0.360}&\scalebox{0.85}{0.338}&\scalebox{0.85}{0.382}\\

    \scalebox{0.85}{Traffic} & \secondres{\scalebox{0.85}{0.454}}&\scalebox{0.85}{0.307}&\scalebox{0.85}{0.466}&\scalebox{0.85}{0.309}&\scalebox{0.85}{0.485}&\scalebox{0.85}{0.307}&\boldres{\scalebox{0.85}{0.428}}&\boldres{\scalebox{0.85}{0.282}}&\scalebox{0.85}{0.490}&\scalebox{0.85}{0.330}&\scalebox{0.85}{0.481}&\secondres{\scalebox{0.85}{0.304}}&\scalebox{0.85}{0.625}&\scalebox{0.85}{0.383}&\scalebox{0.85}{0.593}&\scalebox{0.85}{0.356}&\scalebox{0.85}{0.620}&\scalebox{0.85}{0.349}&\scalebox{0.85}{0.609}&\scalebox{0.85}{0.376}&\scalebox{0.85}{0.628}&\scalebox{0.85}{0.379}\\
    \bottomrule
  \end{tabular}
    \end{small}
  \end{threeparttable}
  }
  % \vspace{-3mm}
\end{table*}

When training models with an MoE architecture, unbalanced expert utilization can lead to suboptimal learning, as some experts may receive disproportionately fewer updates~\cite{shazeer2017outrageously}. To mitigate this issue, we utilize an auxiliary load balancing loss:
\begin{align}
\begin{split}
\mathcal{L}_{\text{load}}  &= M \sum_{e=1}^{M} \mathcal{F}_e \mathcal{P}_e, \\
\mathcal{F}_e &= \frac{1}{rKL} \sum_{k=1}^{K} \sum_{t=1}^{L} \mathbb{1} \{ \hat{x}^k_t \text{ selects Expert }e\}, \\
\mathcal{P}_e &= \frac{1}{KL} \sum_{k=1}^{K} \sum_{t=1}^L g(\tilde{\bm{H}}^{(l-1)}_{k},\mathbf{C}_G )[e].
\end{split}
\end{align}
Here, $\mathcal{F}_e$ represents the fraction of tokens assigned to expert $e$, $\mathcal{P}_e$ denotes the gating probability proportion allocated to expert $e$, and $\mathbb{1}$ is an indicator function that tracks token-to-expert assignments. This loss encourages the gating mechanism to distribute computations more evenly across experts, ensuring all experts receive sufficient updates.

The overall loss function combines the prediction loss and the load balancing loss, weighted by a hyperparameter $\alpha$:
\begin{equation}
    \mathcal{L} = \mathcal{L}_{\text{pred}} + \alpha \mathcal{L}_{\text{load}}.
\end{equation}

\section{Experiments}
\label{sec:experiments}

We conduct extensive experiments to evaluate the effectiveness of the proposed model. Specifically, we address the following key research questions:
\begin{itemize}
    \item \textbf{RQ~1.} How effectively does ContexTST model time series in in-domain forecasting scenarios?
    \item \textbf{RQ~2.} To what extent does ContexTST exhibit zero-shot transfer capability?
    \item \textbf{RQ~3.} Can ContexTST maintain its competitive advantage in the more challenging one-to-one zero-shot transfer setting?
    \item \textbf{RQ~4.} How does ContexTST compare to foundation models pre-trained on large-scale datasets in terms of zero-shot transfer performance?
\item \textbf{RQ~5.} What are the contributions of the proposed modules to the overall performance of ContexTST?
\end{itemize}

Additional results, including full in-domain and cross-domain evaluations (Appendices \ref{app:in_domain}, \ref{app:cross_domain}), efficiency analysis (Appendix \ref{app:model_mem}), parameter sensitivity studies (Appendix \ref{app:param_analysis}), and visualization insights (Appendix \ref{app:viz_analysis}), are provided in the appendix.

\subsection{Experimental Setup}
\label{sec:exp_setup}

\textbf{\emph{Datasets.}} 
We evaluate ContexTST on seven real-world benchmark datasets spanning diverse time series applications, following existing studies \cite{liu2024unitime,patchtst,liu2023itransformer}. 
% Including several distinct domains, like Energy (i.e., ETT-archive~\cite{haoyietal-informer-2021}), Electricity (i.e., Electricity\footnote{https://archive.ics.uci.edu/ml/datasets/ElectricityLoadDiagrams20112014}), Climate (i.e., Weather\footnote{https://www.bgc-jena.mpg.de/wetter/}), and Transport (i.e., Traffic\footnote{https://pems.dot.ca.gov/}).
Details are provided in Table~\ref{tab:dataset}. 

\begin{itemize}
    \item \textbf{ETT}~\cite{haoyietal-informer-2021}: The \textbf{Electricity Transformer Temperature (ETT)} dataset records oil temperature in electrical transformers and six external power load features in China from July 2016 to July 2018. It includes four subsets based on sampling frequency: \textbf{ETTh1} and \textbf{ETTh2} (hourly), and \textbf{ETTm1} and \textbf{ETTm2} (15-minute).  
    \item \textbf{Electricity}\footnote{https://archive.ics.uci.edu/ml/datasets/ElectricityLoadDiagrams20112014}: Records hourly electricity consumption (kWh) of 321 customers from July 2016 to July 2019.  
    \item \textbf{Weather}\footnote{https://www.bgc-jena.mpg.de/wetter/}: Contains 21 meteorological variables collected every 10 minutes from the Max Planck Biogeochemistry Institute's weather station in 2020.  
    \item \textbf{Traffic}\footnote{https://pems.dot.ca.gov/}: Hourly road occupancy data from California of Transportation sensors on San Francisco Bay Area freeways.  
\end{itemize}

\begin{figure*}[t]
    \centering
    \subfigure[ETTh1$\rightarrow$Weather]{\includegraphics[width=0.24\textwidth]{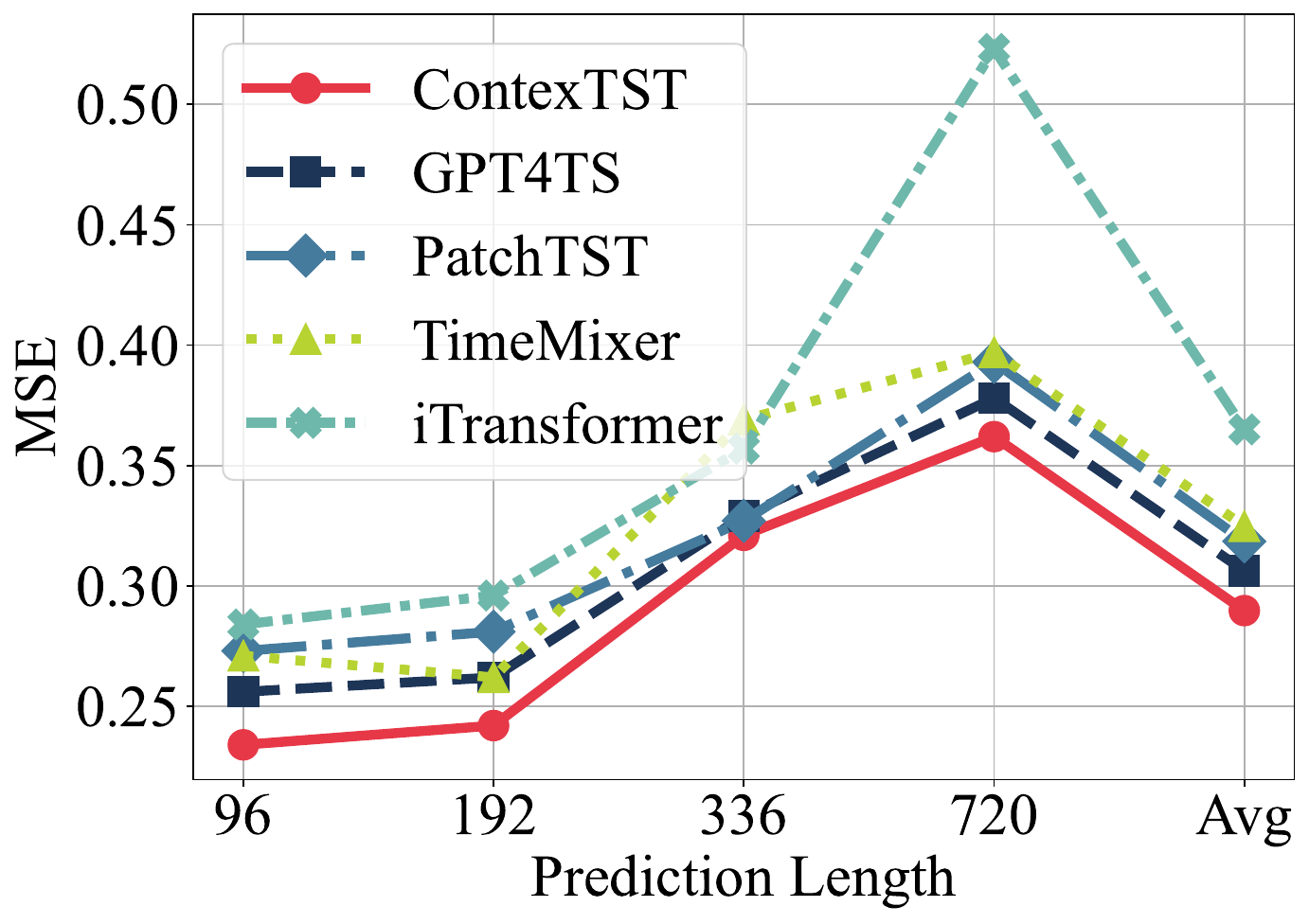}\label{fig:etth1_weather_mse}}
    \hfill
    \subfigure[ETTh1$\rightarrow$Electricity]{\includegraphics[width=0.24\textwidth]{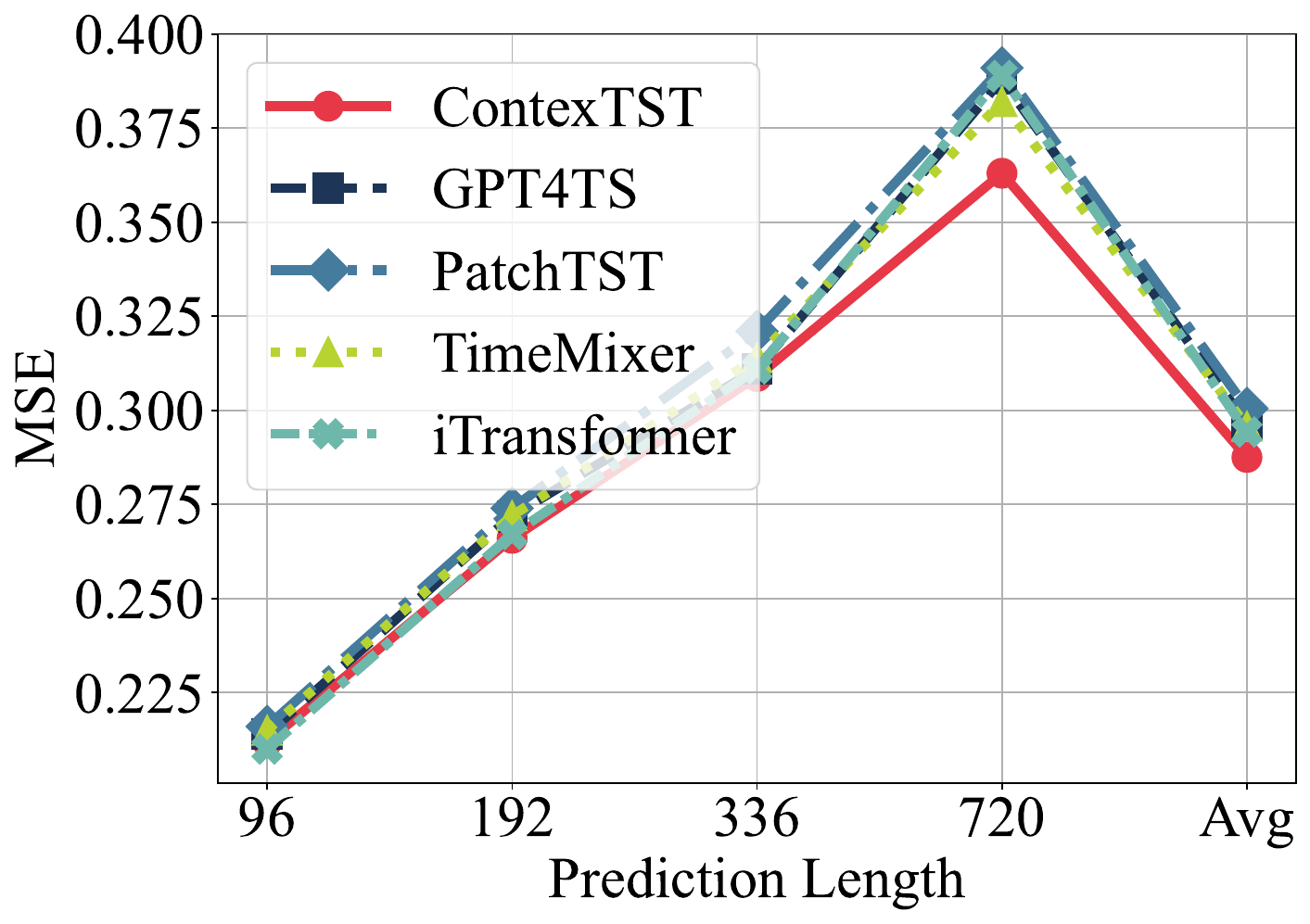}\label{fig:etth1_electricity_mse}}
    \subfigure[Electricity$\rightarrow$ETTh1]{\includegraphics[width=0.24\textwidth]{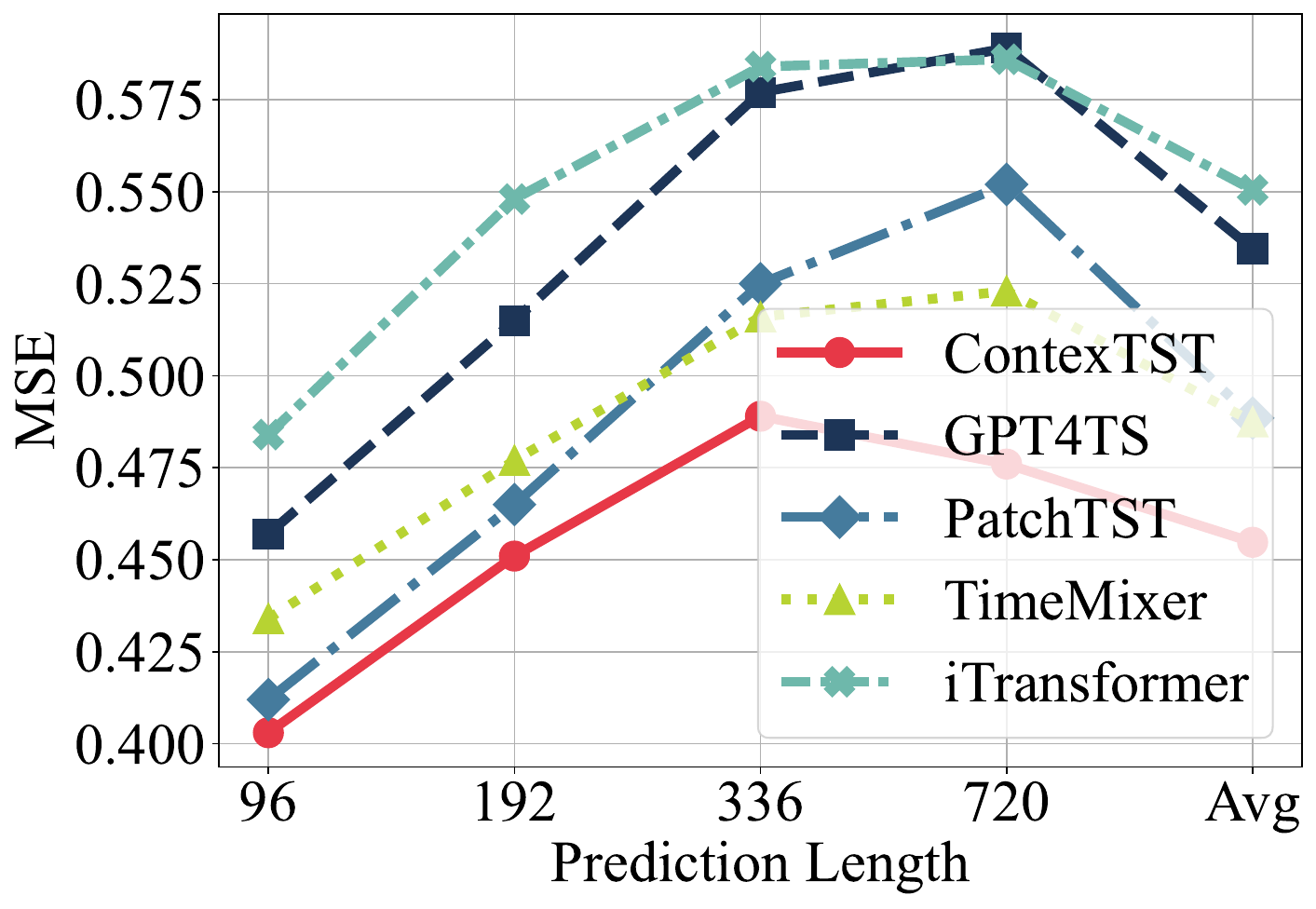}\label{fig:electricity_etth1_mse}}
    \hfill
  \subfigure[Electricity$\rightarrow$Weather]{\includegraphics[width=0.24\textwidth]{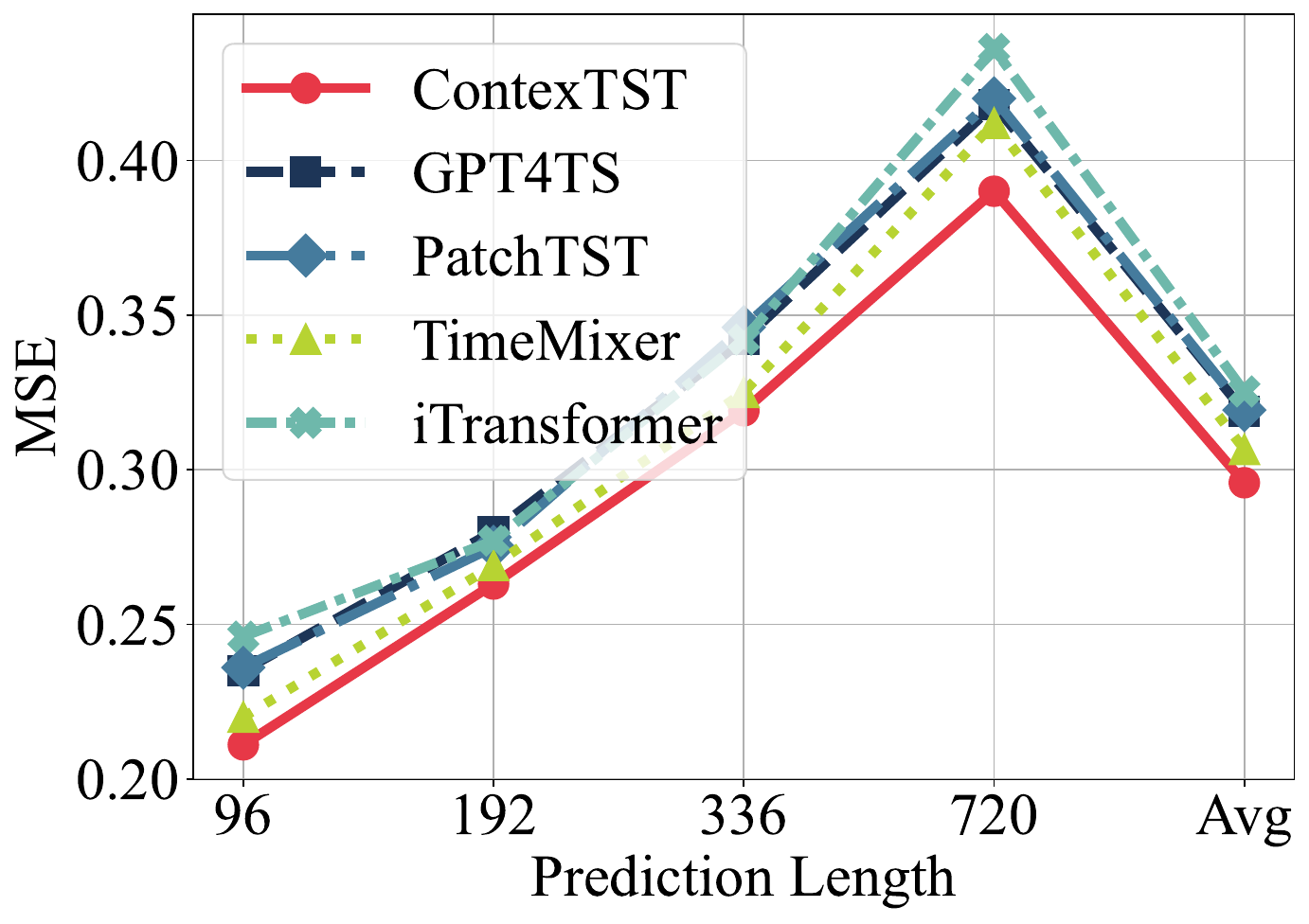}\label{fig:electricity_weather_mse}}
  \vspace{-3mm}
    \caption{One-to-one zero-shot forecasting. We set ETTh1 and Electricity as the source datasets for model training and transfer ContexTST to multiple target domains. The look-back window is set to 96 for all models. Find the full results in Appendix~\ref{sec:ap_cross_domain}}
    \vspace{-3mm}
    \label{fig:one-to-one_transfer}
\end{figure*}

\noindent \textbf{\emph{Baselines.}} 
We evaluate ContexTST against the following baselines, categorized into three groups: (1) Classical models without context information, including TimeXer \cite{wang2024timexer}, TimeMixer \cite{wang2024timemixer}, PatchTST \cite{patchtst}, iTransformer \cite{liu2023itransformer}, DLinear \cite{dlinear}, MICN \cite{wang2023micn}, TimesNet \cite{wu2022timesnet}, FEDformer \cite{zhou2022fedformer}, and Autoformer \cite{wu2021autoformer}. (2) Forecasting models incorporating contextual information, including UniTime\cite{liu2024unitime} and GPT4TS\cite{zhou2023one}. (3) Time series foundation models, including Chronos\cite{ansari2024chronos}, TimesFM\cite{das2023decoder}, and Moirai~\cite{liu2024moirai}. Further details on the baselines are provided in Appendix \ref{app:data_info}.

\noindent \textbf{\emph{Implementation.}} We implement ContexTST using PyTorch 2.4.0, both training and inference are performed on CUDA 12.2. We utilize the Adam optimizer with an initial learning rate of $10^{-3}$. All experiments are conducted on 2*NVIDIA A800-SXM4-80GB GPUs. We use MSE and MAE as our evaluation metrics. To ensure fair comparisons with other baselines, we explore the hyperparameter search space around their best-reported configurations. The detailed model configuration can be found in Appendix Table~\ref{tab:model_parameters}.

\subsection{In-Domain Time Series Forecasting (RQ1)}
\label{sec:exp_indomain}

To assess ContexTST’s ability to capture temporal patterns, we compare it against state-of-the-art forecasting models in in-domain settings. We evaluate performance across four forecasting horizons, with average results summarized in Table~\ref{tab:brif_indomain_forecasting}. Full results are provided in Appendix Table~\ref{tab:full_forecasting_results}.

As shown in Table~\ref{tab:brif_indomain_forecasting}, ContexTST consistently demonstrates strong performance across seven benchmark datasets. It achieves the best results on the ETT, Electricity, and Weather datasets while remaining highly competitive on Traffic. On average, ContexTST reduces MSE by 3.5\% compared to TimeMixer \cite{wang2024timemixer} and by 12.7\% compared to GPT4TS \cite{zhou2023one}. These results underscore ContexTST’s effectiveness in modeling complex temporal patterns, highlighting its expressiveness and capability for time-series forecasting.

\vspace{-1mm}
\subsection{Zero-Shot Transferability (RQ2)}
\label{sec:exp_multi}

Following the experimental setup of UniTime~\cite{liu2024unitime}, we conduct a zero-shot transfer analysis to evaluate ContexTST’s transferability. In this experiment, models are pre-trained on multiple source domains and then directly evaluated on unseen target domains without further fine-tuning.
Specifically, we first train the models on ETTh1, ETTm1, and ETTm2 datasets, and then assess their performance in both in-domain (ETTh2) and out-of-domain (Electricity and Weather) transfer scenarios under zero-shot testing.

As shown in Table~\ref{tab:merge_train}, ContexTST outperforms all baseline models across all target datasets and prediction horizons. In the in-domain transfer setting (ETTh2), ContexTST achieves the lowest average prediction error, demonstrating its strong ability to generalize within similar domains. In the out-of-domain transfer setting on Electricity, ContexTST surpasses all baselines, achieving a 3.6\% lower MSE than the state-of-the-art time-series transfer model UniTime~(\citeyear{liu2024unitime}).
We speculate that this performance gain stems from the complex superposition of periodic patterns in electricity consumption data. Different from UniTime, which primarily leverages contextual information for adaptation, ContexTST benefits from a unified frequency-domain perspective, which proves advantageous in such cross-domain transfer scenarios.

\begin{figure*}[t]
    \centering
    \subfigure[ETTh1]{\includegraphics[width=0.24\textwidth]{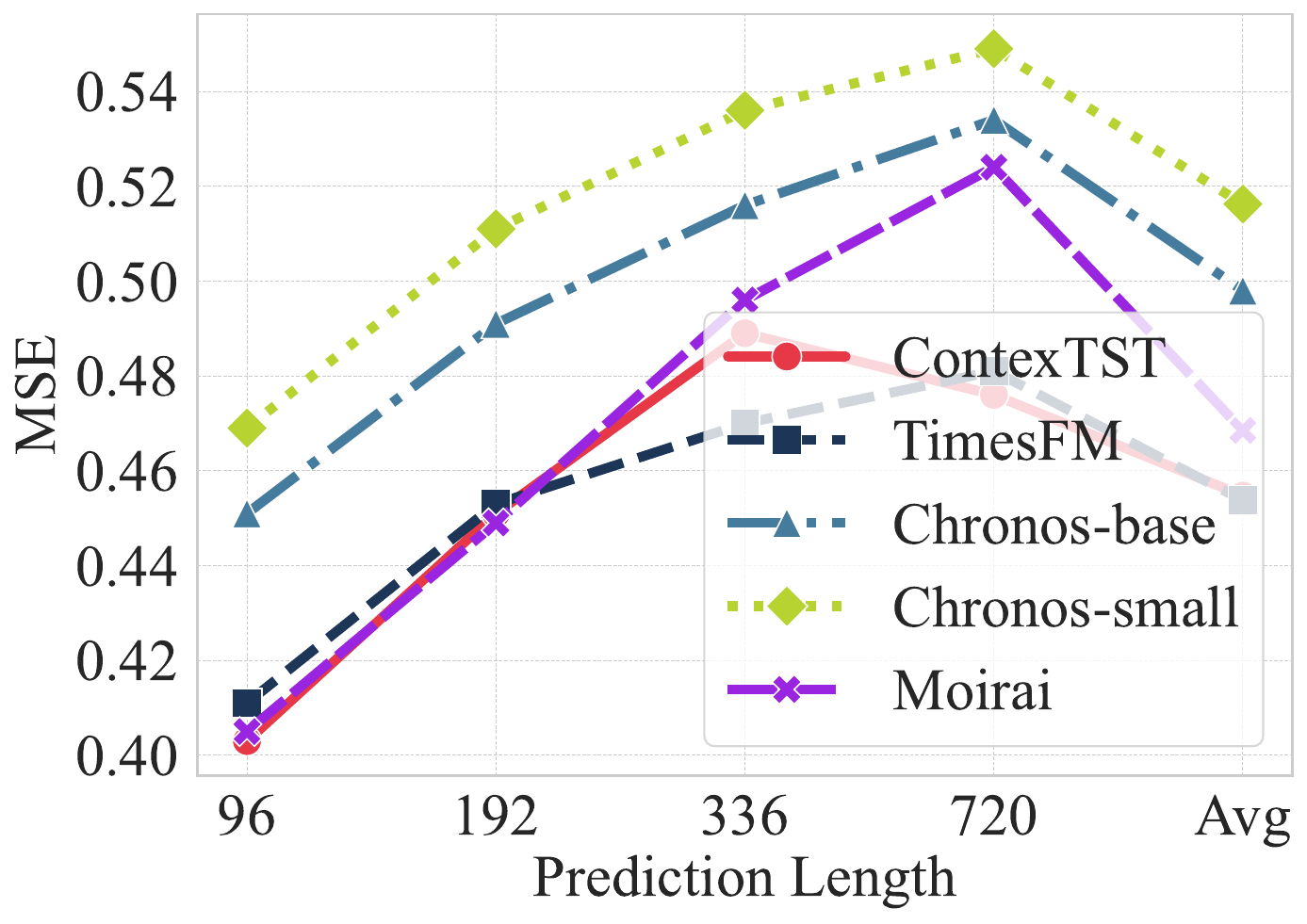}\label{fig:etth1_mse}}
    \hfill
    \subfigure[ETTh2]{\includegraphics[width=0.24\textwidth]{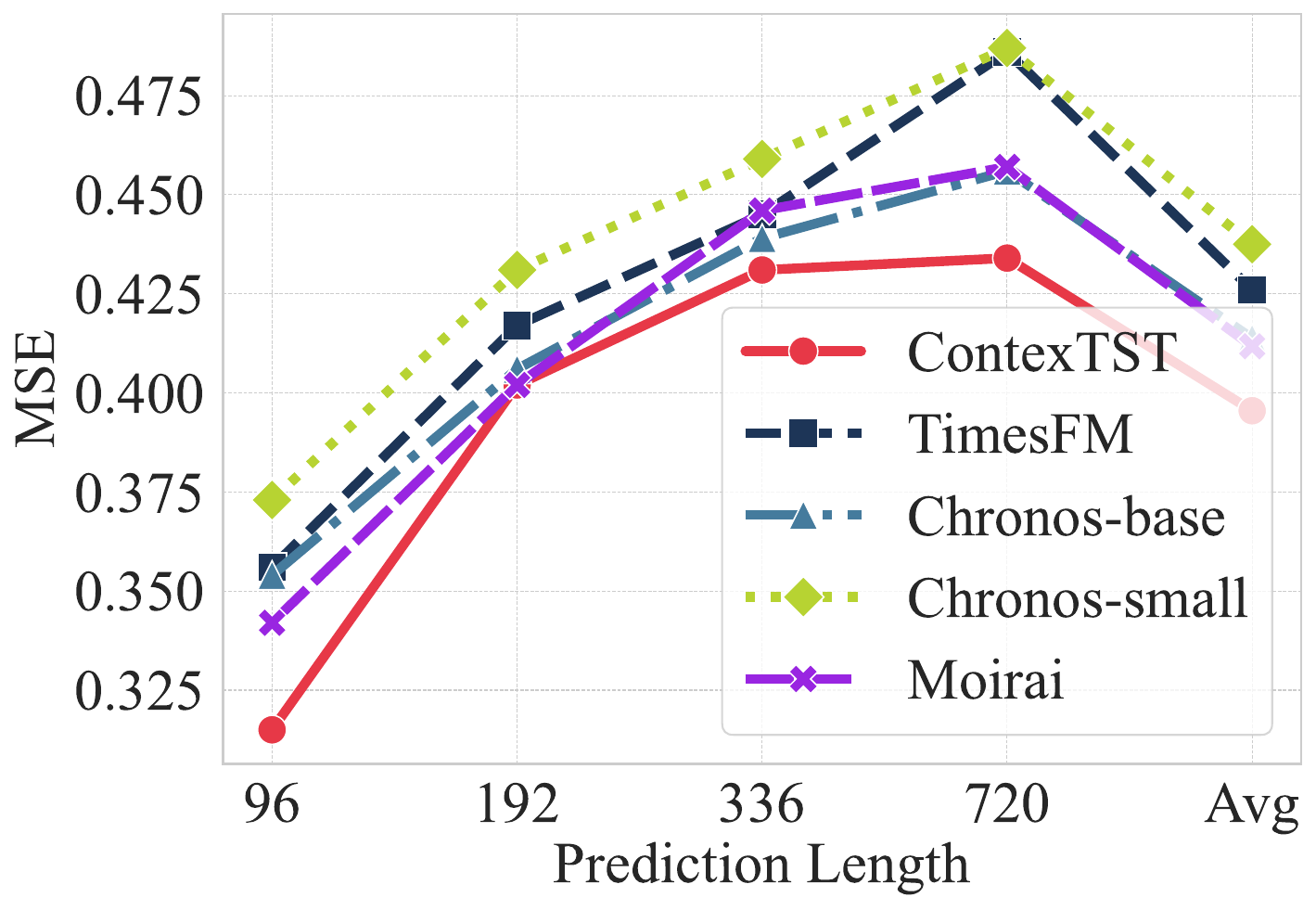}\label{fig:etth2_mse}}
    \hfill
    \subfigure[ETTm1]{\includegraphics[width=0.24\textwidth]{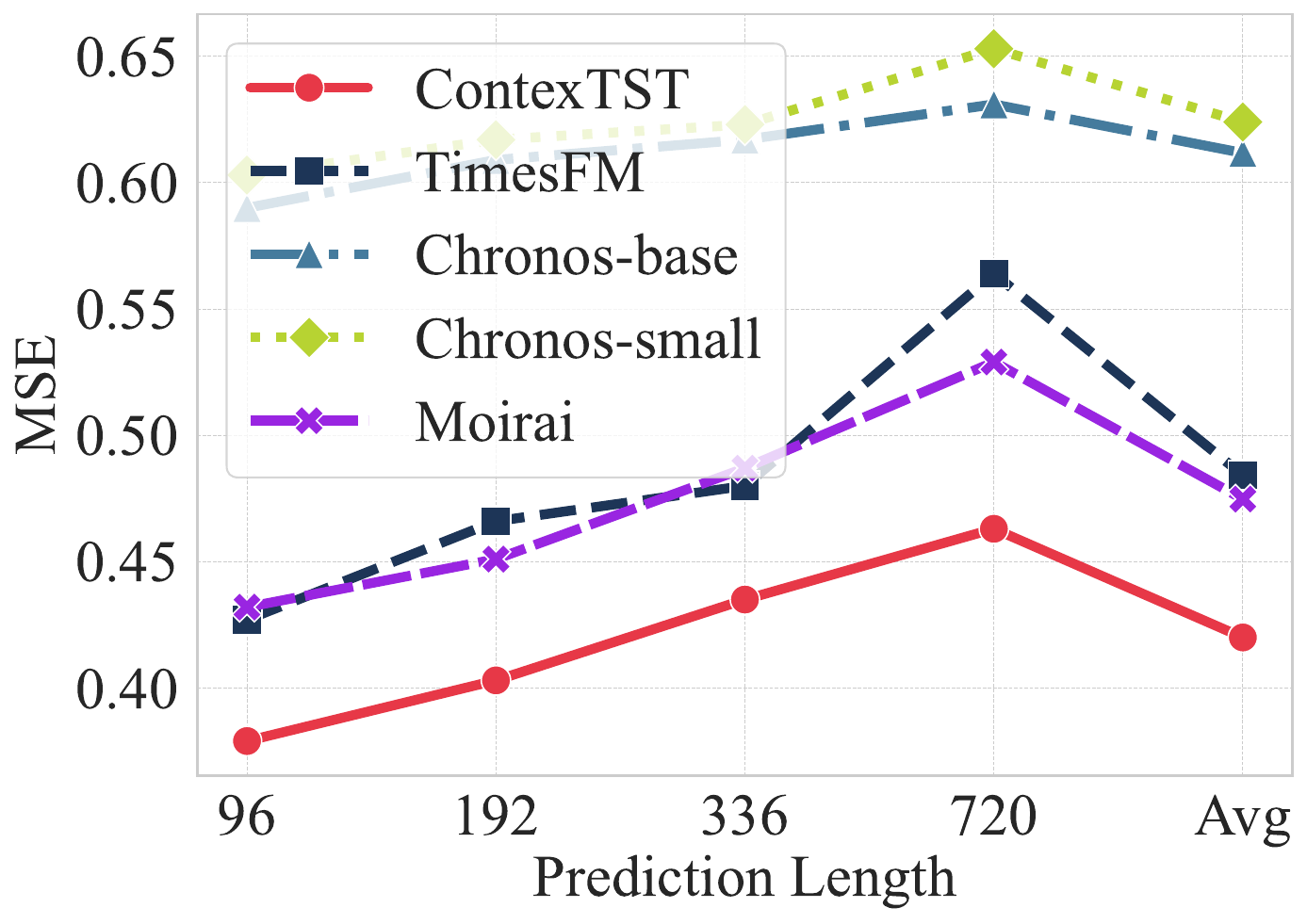}\label{fig:ettm1_mse}}
    \hfill
    \subfigure[ETTm2]{\includegraphics[width=0.24\textwidth]{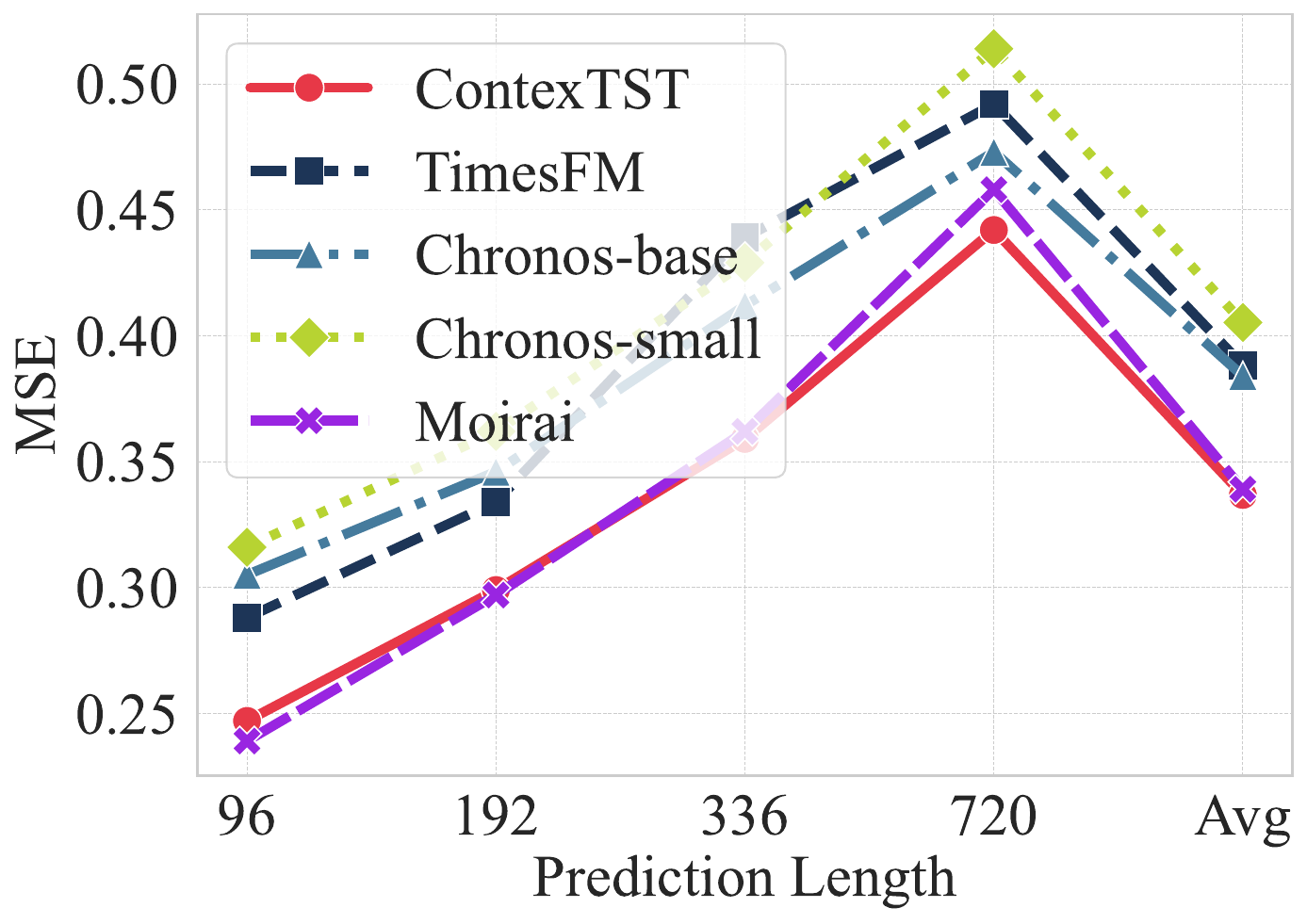}\label{fig:ettm2_mse}}
    \vspace{-3mm}
    \caption{Comparison with foundational time series models on diverse prediction horizons. The input sequence length is set to 96 for all models. For each foundation model, we exclude the evaluation results on its pre-trained datasets, and ContexTST is pre-trained on Electricity datasets then zero-shot inference in 4 target domains.}
    \vspace{-3mm}
    \label{fig:foundation_res}
\end{figure*}

\begin{figure*}[th]
    \centering
    
    \subfigure[ETTh1$\rightarrow$Weather]{\includegraphics[width=0.24\textwidth]{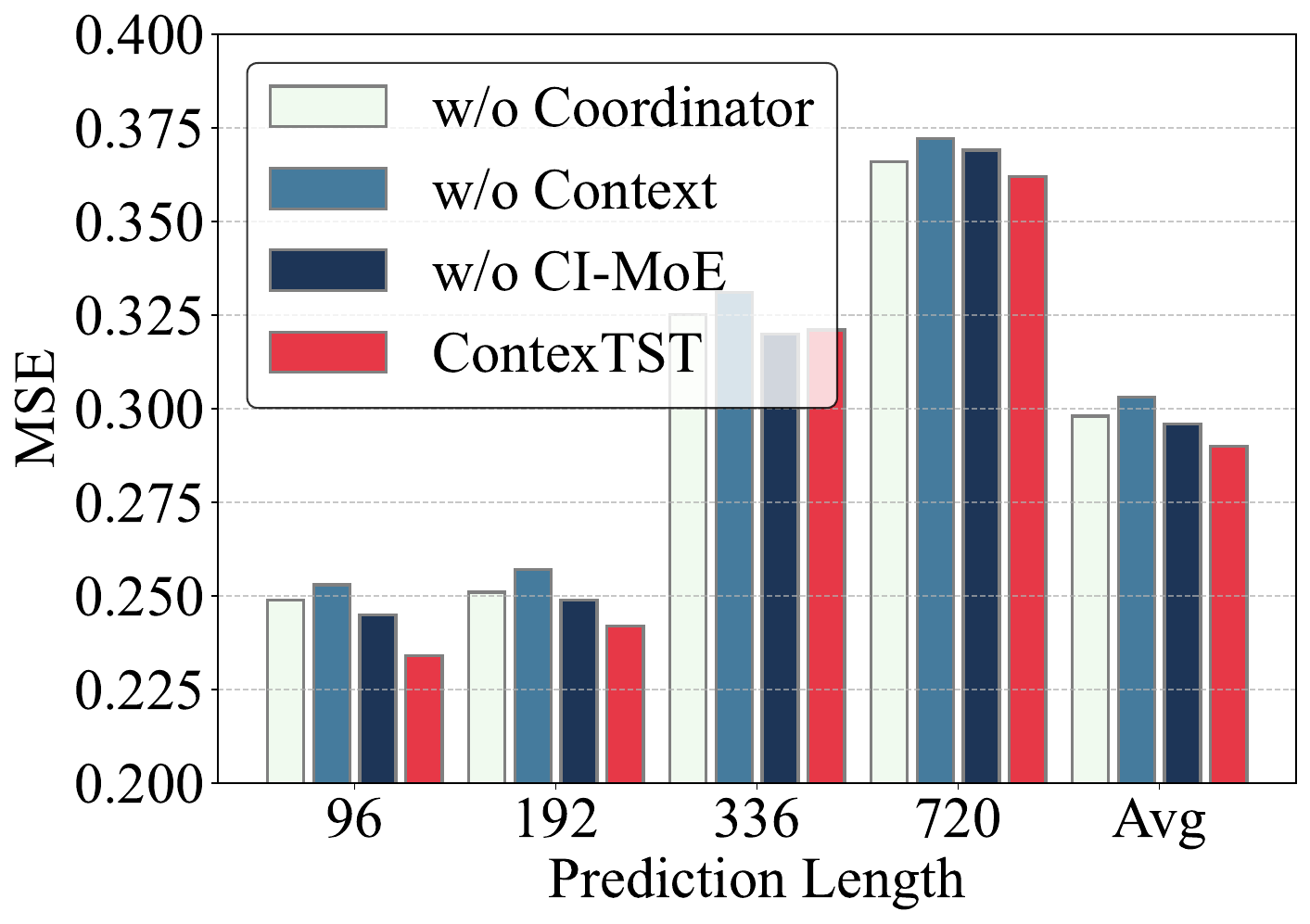}\label{fig:ab_etth1_weather_mse}}
    \hfill
    \subfigure[ETTh1$\rightarrow$Electricity]{\includegraphics[width=0.24\textwidth]{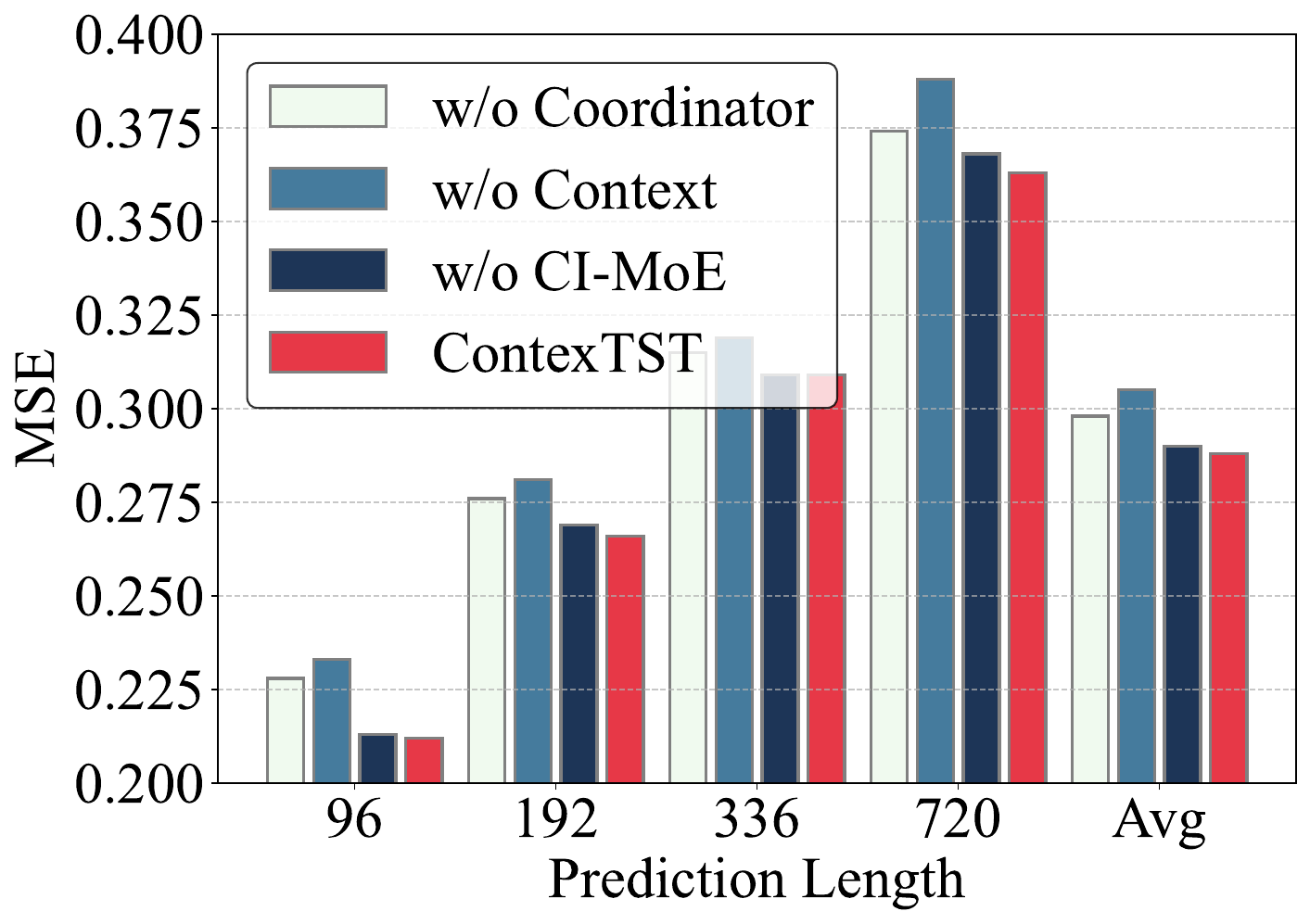}\label{fig:ab_etth1_ecl_mse}}
    \subfigure[Electricity$\rightarrow$ETTh1]{\includegraphics[width=0.24\textwidth]{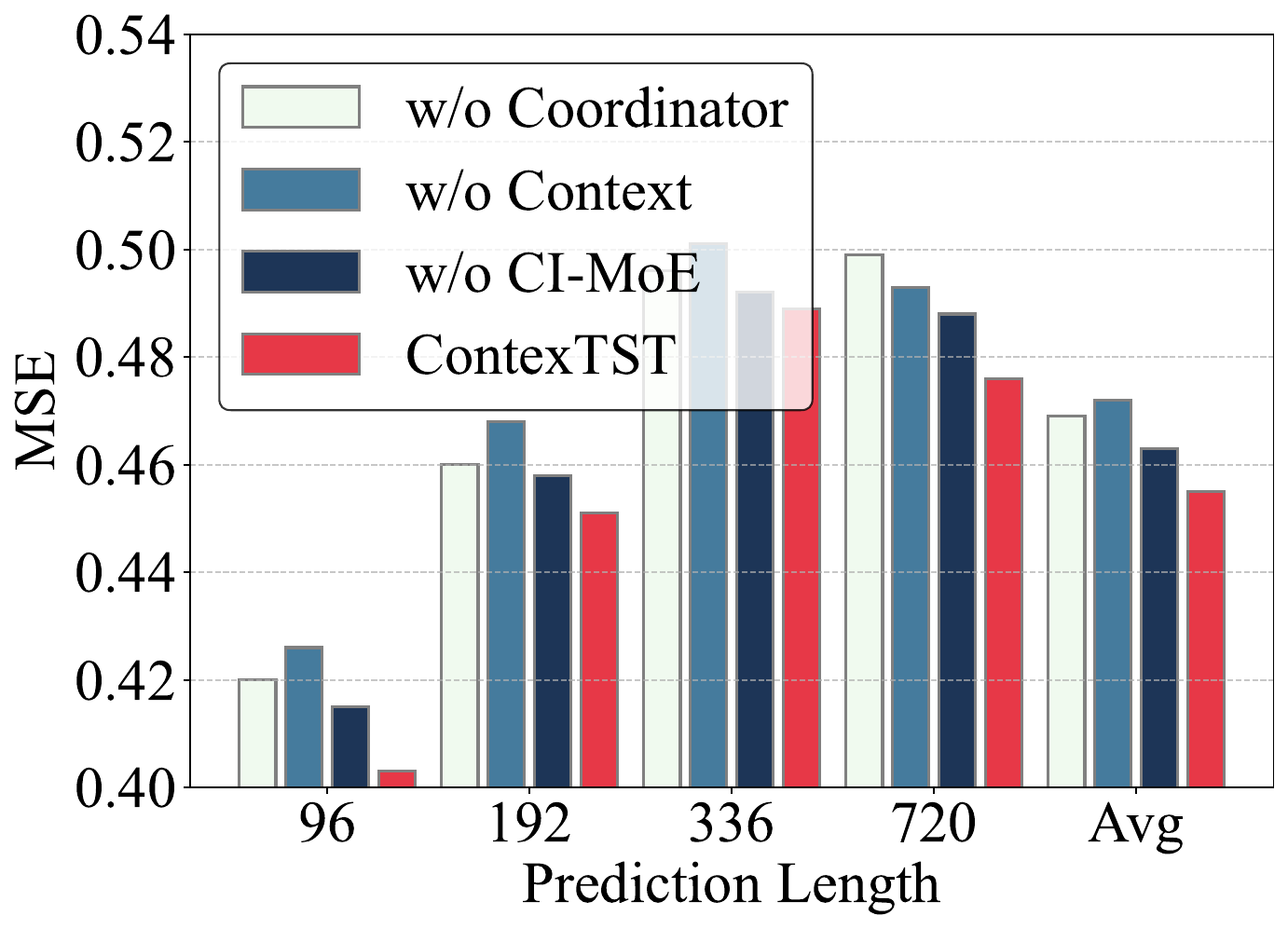}\label{fig:ab_ecl_etth1_mse}}
    \hfill
    \subfigure[Electricity$\rightarrow$Weather]{\includegraphics[width=0.24\textwidth]{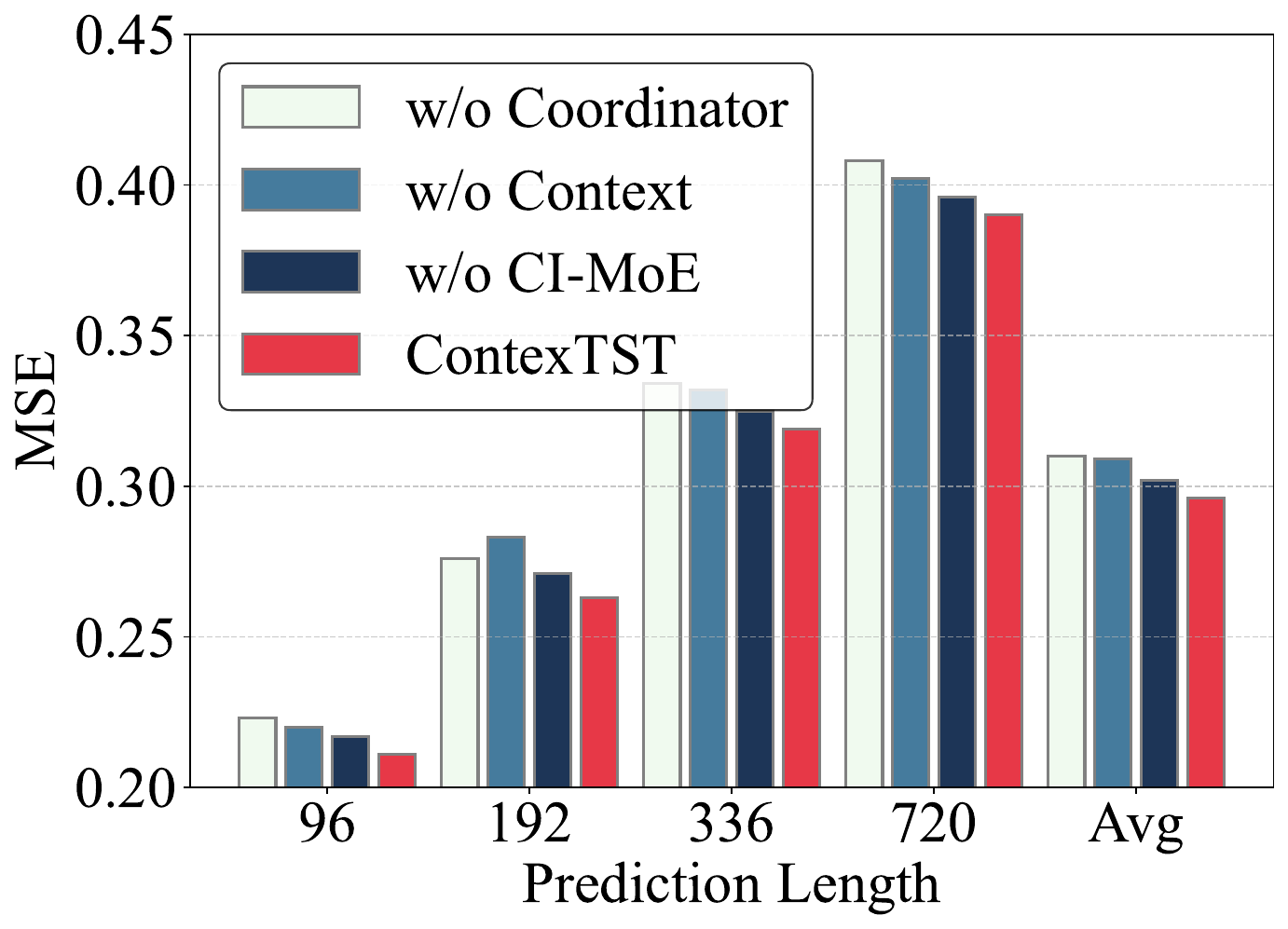}\label{fig:ab_ecl_weather_mse}}
    \vspace{-3mm}
    \caption{The ablation studies about our proposed coordinator, context, and CI-MoE modules in cross-domain transfer scenario. We report here 4 ablation experiments on challenging domain transfer tasks, while for additional cross-domain and in-domain results please refer to appendix Figure~ \ref{fig:ablation_studies} and \ref{fig:ap_cross_ablation_studies}.}
    \vspace{-3mm}
    \label{fig:cross_ablation_studies}
\end{figure*}

\vspace{-1mm}
\subsection{One-to-One Zero-Shot Transfer (RQ3)}
\label{sec:exp_zero}

With the aim of further assessing ContexTST’s transferability, we adopt a more challenging experimental setup: training on a single source dataset and directly inferring on a different target dataset, a setting we refer to as one-to-one transfer. The results of this experiment are presented in Figure~\ref{fig:one-to-one_transfer}.

Surprisingly, some models not explicitly designed for cross-domain transfer tasks perform better than expected. For instance, PatchTST, despite not leveraging additional contextual information, retains a certain level of predictive capability even under complete domain shifts.
Among them, ContexTST consistently achieves the lowest forecasting errors across all prediction horizons, with particularly strong performance in longer horizons (e.g., 720), making it a robust choice for zero-shot forecasting tasks.

\begin{table}[t]
  \caption{Multi-datasets merge pre-trained and zero-shot inference results. \emph{Avg} is averaged from all four prediction lengths, that is 96, 192, 336, 720. The best is in \textbf{bold}.}\label{tab:merge_train}
  \vspace{-2mm}
  \centering
  \resizebox{1.0\linewidth}{!}{
  \begin{threeparttable}
  % \begin{small}
  \renewcommand{\multirowsetup}{\centering}
  \begin{tabular}{c|c|cc|cc|cc|cc}
  \toprule
    \multicolumn{2}{c}{\multirow{2}{*}{Models}} &
    \multicolumn{2}{c}{\rotatebox{0}{\textbf{ContexTST}}} &
    \multicolumn{2}{c}{\rotatebox{0}{UniTime}} &
    \multicolumn{2}{c}{\rotatebox{0}{GPT4TS}} &
    \multicolumn{2}{c}{\rotatebox{0}{PatchTST}}\\ 

    \multicolumn{2}{c}{} & \multicolumn{2}{c}{(\textbf{Ours})} &
    \multicolumn{2}{c}{\citeyear{liu2024unitime}} &
    \multicolumn{2}{c}{\citeyear{zhou2023one}}&
    \multicolumn{2}{c}{\citeyear{patchtst}}\\

    \cmidrule(lr){3-4} \cmidrule(lr){5-6}\cmidrule(lr){7-8} \cmidrule(lr){9-10} 
    \multicolumn{2}{c}{Metric} & MSE & MAE & MSE & MAE & MSE & MAE & MSE & MAE \\
    \toprule

    \multirow{5}{*}{\rotatebox{90}{ETTh2}} 
    & 96 & \textbf{0.295} & \textbf{0.332} & 0.306 & 0.352 & 0.316 & 0.361 & 0.332 & 0.371 \\
    & 192 & \textbf{0.372} & \textbf{0.364} &	0.389 &	0.401 &	0.400 &	0.410 &	0.422 &	0.421 \\
    & 336 & \textbf{0.420} & \textbf{0.419} &	0.424 &	0.434 &	0.430 &	0.439 &	0.462 &	0.455 \\
    & 720 & \textbf{0.423} & \textbf{0.433} &	0.433 &	0.450 &	0.442 &	0.461 &	0.467 &	0.469 \\
    \cmidrule(lr){2-10}
    & Avg & \textbf{0.378} & \textbf{0.387} &	0.388 &	0.409 &	0.397 &	0.418 &	0.421 &	0.429 \\
    \midrule

    \multirow{5}{*}{\rotatebox{90}{Electricity}}
    & 96 & \textbf{0.383} &	\textbf{0.436} &	0.409 &	0.481 &	0.448 &	0.520 &	0.529 &	0.562 \\
    & 192 & \textbf{0.396} &	\textbf{0.458} &	0.410 &	0.484 &	0.443 &	0.517 &	0.507 &	0.550 \\
    & 336 & \textbf{0.429} & \textbf{0.484} & 0.439 & 0.504 & 0.462 & 0.526 & 0.536 & 0.566 \\
    & 720 & \textbf{0.476} &	\textbf{0.528} &	0.487 &	0.531 &	0.494 &	0.542 &	0.563 &	0.581 \\
    \cmidrule(lr){2-10}
    & Avg & \textbf{0.421} &	\textbf{0.477} &	0.436 &	0.500 &	0.462 &	0.526 &	0.534 &	0.565 \\
    \midrule

    \multirow{5}{*}{\rotatebox{90}{Weather}}
    & 96 & \textbf{0.187} &	\textbf{0.251} &	0.210 &	0.262 &	0.223 &	0.271 &	0.235 &	0.277 \\
    & 192 & \textbf{0.254} &	\textbf{0.289} &	0.264 &	0.303 &	0.287 &	0.319 &	0.293 &	0.320 \\
    & 336 & \textbf{0.306} &	\textbf{0.326} &	0.326 &	0.334 &	0.347 &	0.357 &	0.351 &	0.356 \\
    & 720 & \textbf{0.388} &	\textbf{0.362} &	0.402 &	0.382 &	0.432 &	0.409 &	0.427 &	0.404 \\
    \cmidrule(lr){2-10}
    & Avg & \textbf{0.284} &	\textbf{0.307} &	0.301 &	0.320 &	0.322 &	0.339 &	0.327 &	0.339 \\  
    \bottomrule
  \end{tabular}
    % \end{small}
  \end{threeparttable}
  }
  \vspace{-6mm}
\end{table}

Notably, even in the Electricity $\rightarrow$ Weather scenario—where the source and target domains are highly uncorrelated—ContexTST outperforms all baselines, demonstrating its adaptability. We attribute this ability to our “Unify and Anchor” paradigm. By projecting multi-domain time series signals into a common coordinate system using spectral tools, ContexTST effectively encodes temporal dependencies and multi-periodic patterns. Additionally, domain-specific anchors serve as semantic supplements, allowing the model to extract richer contextual information from the target domain and reduce the prediction search space.
These findings highlight ContexTST’s effectiveness in adapting to unseen datasets without retraining, making it a promising solution for cross-domain time series forecasting.

\subsection{Comparison with Foundation Models (RQ4)}
\label{sec:exp_tsfm}

We compare ContexTST’s zero-shot transfer capability against leading foundation models pre-trained on large-scale datasets, including Chronos-small, Chronos-base\cite{ansari2024chronos}, Moirai\cite{woo2024unified}, and TimesFM\cite{das2023decoder}. To ensure a fair evaluation, we select ETT datasets that these models have not been pre-trained on and conduct zero-shot inference, with ContexTST pre-trained solely on the Electricity dataset. The results, measured by MSE, are reported in Figure\ref{fig:foundation_res}.

Despite being trained on a single dataset, ContexTST always achieves lower or comparable MSE compared to TSFMs in zero-shot settings. It significantly outperforms baselines on ETTh2 and ETTm1, particularly for longer forecasting horizons (e.g., 336 and 720 steps), while maintaining competitive performance on ETTh1 and ETTm2. This observation suggests that while pre-training on diverse datasets enhances cross-domain generalization, numerical time-series data lacks explicit semantic information, leading to potential performance degradation when zero-shot transfer encounters semantic misalignment. In contrast, ContexTST leverages contexts as the domain anchor, effectively mitigating this issue and compensating for the limitations of training on a single dataset.

Additionally, we observe that TSFMs excel in short-term forecasting (e.g., 12-step predictions) when utilizing extremely long look-back windows (e.g., 5000 steps), as demonstrated by Moirai’s default configuration~\cite{liu2024moirai}. However, ContexTST achieves effective cross-domain forecasting without relying on excessively long look-back windows. We attribute this advantage to the domain anchoring mechanism, as well, where contextual information provides additional semantic cues, reducing the search space and enhancing predictive accuracy.

\subsection{Ablation Study (RQ5)}
\label{sec:exp_ablation}

To evaluate the effectiveness of ContexTST’s core components in cross-domain time series forecasting, we conduct ablation studies on three key variants of our model. Specifically, we examine the impact of removing the Time Series Coordinator, contextual information (where CI-MoE is degraded into standard MoE), and CI-MoE module across four one-to-one domain transfer scenarios at various prediction lengths. The results are summarized in Figure~\ref{fig:cross_ablation_studies}. For more ablation studies under in-domain scenario, please refer to Appendix Sec~\ref{sec:app_ab1}.  

The experimental results highlight the critical role of each component in ContexTST. First, the removal of the coordinate system significantly impairs zero-shot transfer forecasting performance, confirming its importance in structuring cross-domain temporal patterns. Additionally, the absence of contextual information leads to a noticeable decline in predictive accuracy across all datasets. For instance, in the Weather dataset (Figure~\ref{fig:ab_etth1_weather_mse}), the average MAE increases from 0.288 to 0.305, demonstrating that context-aware representations enhance ContexTST’s forecasting capability.
Furthermore, we observe that the CI-MoE module plays an indispensable role in cross-domain adaptation, as its removal results in a clear performance drop. The full ContexTST model consistently achieves the best performance, illustrating the synergistic effect of these components in capturing complex temporal patterns and enabling robust generalization across diverse forecasting tasks.
\section{Conclusion}
\label{sec:conclusion}

We propose the ``Unify and Anchor" paradigm for cross-domain time series forecasting, which disentangles frequency components for a unified representation and integrates external context as domain anchors for adaptation. Building on this framework, we introduce ContexTST, a Transformer varient, and validate its effectiveness through extensive experiments, including in-domain forecasting, cross-domain transfer, and ablation studies.

While ContexTST demonstrates strong performance, it has limitations. Its FFT-based decomposition assumes fixed-length, regularly sampled inputs, limiting its applicability to variable-length and irregularly sampled time series. Additionally, while ContexTST exemplifies the “Unify and Anchor” approach, alternative implementations remain unexplored. Finally, our evaluations are conducted on relatively small-scale datasets, and scaling to large datasets with optimized computational efficiency is a crucial next step. Addressing these challenges will further advance time series foundation models and broaden their real-world impact.

% In this paper, we propose ContexTST with the insight of ``Unified and then Anchor" for time series cross-domain forecasting. We introduce a time series coordinator based on the perspective of frequency decomposition, which constructs a unified space to facilitate multiple domain series learning. To tackle the semantic misalignment during domain transfer, we propose to utilize the context information as domain anchors for expert routing. Extensive experiments validate the effectiveness of the proposed model, covering both in-domain long-term time series forecasting, cross-domain transfer tasks, and ablation studies. Compared to several state-of-the-art domain-specific, transfer-adapted, and time series foundation models, ContexTST consistently outperforms its counterparts. The ablation results illustrate the effectiveness of the time series coordinator and context-informed MoE. The comparison results demonstrate the superior performance, robustness, and generalization capability of the context-aware transformer in ContexTST.

\bibliographystyle{ACM-Reference-Format}
\bibliography{reference}

\clearpage
\appendix

\section{Details of Datasets and Baselines}
\label{app:data_info}
\noindent \textbf{Dataset.} We evaluate the performance of different models on 7 well-established datasets from multiple real-world application domains, including Temperature (i.e., ETT), Climate (i.e., Weather), Transposition (i.e., Traffic), and Energy (i.e., Electricity). The detailed information of these datasets is summarized in Table~\ref{tab:dataset}.
(1) \textbf{ETT}~\cite{haoyietal-informer-2021} is the Electricity Transformer Temperature (ETT) series recording the oil temperature of electrical transformers and
six external power load features in a region in China, covering the period from July 2016 to July 2018. According to the sampling frequency, ETT contains 4 datasets: \textbf{ETTh1} and \textbf{ETTh2} are in 1-hour frequency, \textbf{ETTm1} and \textbf{ETTm2} are in 15-minute frequency. (2) \textbf{Electricity} is an electricity-consuming load dataset, recording the hourly electricity consumption (Kwh) of 321 customers collected from 2016/7/1 to 2019/7/2. (3) \textbf{Weather} includes 21 meteorological factors collected every 10 minutes from the
Weather Station of the Max Planck Biogeochemistry Institute in 2020. (4) \textbf{Traffic} is a collection of hourly data from the California Department of Transportation,
which describes the road occupancy rates measured by different sensors on San Francisco Bay area freeways.

\noindent \textbf{Metric.} We utilize the Mean Square Error (MSE) and Mean Absolute Error (MAE) as evaluation metrics for time series forecasting. The MSE is a quantitative metric used to measure the average squared difference between the observed actual value and forecasts. It is defined mathematically as follows:
\begin{align}
    \text{MSE} = \frac{1}{C\times T}\sum_{c=1}^{C}\sum_{t=1}^{T}(x_{t}^{c}-\hat{x}_{t}^{c})^2,
\end{align}
where $c$ denotes the variable and $T$ is the prediction length. The MAE quantifies the average absolute deviation between the forecasts and the true values. Since it averages the absolute errors, MAE is robust to outliers. Its mathematical formula is given by:
\begin{align}
    \text{MAE} =  \frac{1}{C\times T}\sum_{c=1}^{C}\sum_{t=1}^{T}|x_{t}^{c}-\hat{x}_{t}^{c}|.
\end{align}

\noindent \textbf{Baselines.} To fully evaluate the performance of the proposed model in in-domain time series forecasting and cross-domain transfer scenarios, we conduct extensive experiments comparing current state-of-the-art methods. The baseline methods can be categorized into three groups: (1) In-domain time series forecasting methods: TimeXer~\cite{wang2024timexer}, TimeMixer~\cite{wang2024timemixer}, PatchTST~\cite{patchtst}, iTransformer~\cite{liu2023itransformer}, DLinear~\cite{dlinear}, MICN~\cite{wang2023micn}, TimesNet~\cite{wu2022timesnet}, FEDformer~\cite{zhou2022fedformer}, and Autoformer~\cite{wu2021autoformer}. (2) Methods in cross-domain transfer scenarios, including the specifically designed for cross-domain time series forecasting approach, UniTime~\cite{liu2024unitime}, and other methods adapted to be applied to cross-domain transfer scenarios: GPT4TS~\cite{zhou2023one}, PatchTST~\cite{patchtst}, TimeMixer~\cite{wang2024timemixer}, and iTransformer~\cite{liu2023itransformer}. (3) The time series foundation models which pre-trained on a large time series corpus: TimesFM~\cite{das2023decoder}, Chornos~\cite{ansari2024chronos}, and Moirai~\cite{liu2024moirai}.

\begin{table*}[thbp]
  % \vspace{-10pt}
  \caption{Dataset detailed descriptions. The dataset size is organized in (Train, Validation, Test).}\label{tab:dataset}
  % \vskip 0.05in
  \centering
   \resizebox{0.85\textwidth}{!}{
  \begin{threeparttable}
  \begin{small}
  \renewcommand{\multirowsetup}{\centering}
  \setlength{\tabcolsep}{3.8pt}
  \begin{tabular}{c|c|c|c|c|c|c}
    \toprule
    Dataset & Dim & Series Length & Dataset Size &Frequency &Forecastability$\ast$ & Domain \\
    \toprule
    \cmidrule{1-7}
     ETTh1 & 7 & \{96, 192, 336, 720\} & (8545, 2881, 2881) & 15 min &0.38 & Temperature \\

     \cmidrule{1-7}
     ETTh2 & 7 & \{96, 192, 336, 720\} & (8545, 2881, 2881) & 15 min &0.45 & Temperature \\
     
     \cmidrule{1-7}
     ETTm1 & 7 & \{96, 192, 336, 720\} & (34465, 11521, 11521)  & 15min &0.46 & Temperature\\
     
    \cmidrule{1-7}
    ETTm2 & 7 & \{96, 192, 336, 720\} & (34465, 11521, 11521)  & 15min &0.55 & Temperature\\
    
    \cmidrule{1-7}
    Electricity & 321 & \{96, 192, 336, 720\} & (18317, 2633, 5261) & Hourly &0.77 & Electricity \\

    \cmidrule{1-7}
     Weather & 21 & \{96, 192, 336, 720\} & (36792, 5271, 10540)  &10 min &0.75  & Weather \\
    
    \cmidrule{1-7}
    Traffic & 862 & \{96, 192, 336, 720\} & (12185, 1757, 3509) & Hourly &0.68 & Transportation \\
    
    % \cmidrule{2-8}
    % & Solar-Energy & 137  & \scalebox{0.8}{\{96, 192, 336, 720\}}  & (36601, 5161, 10417)& 10min &0.33 & \scalebox{0.8}{Electricity} \\
    % \midrule
    % & PEMS03 & 358 & 12 & (15617,5135,5135) & 5min &0.65 & \scalebox{0.8}{Transportation}\\
    % \cmidrule{2-8}
    % & PEMS04 & 307 & 12 & (10172,3375,3375) & 5min &0.45 & \scalebox{0.8}{Transportation}\\
    % \cmidrule{2-8}
    % & PEMS07 & 883 & 12 & (16911,5622,5622) & 5min &0.58 & \scalebox{0.8}{Transportation}\\
    % \cmidrule{2-8}
    % Short-term & PEMS08 & 170 & 12 & (10690,3548,265) & 5min &0.52 & \scalebox{0.8}{Transportation}\\
    % \cmidrule{2-8}
    % Forecasting & M4-Yearly & 1 & 6 & (23000, 0, 23000) &Yearly &0.43 &\scalebox{0.8}{Demographic} \\
    % \cmidrule{2-8}
    %  & M4-Quarterly & 1 & 8 & (24000, 0, 24000) &Quarterly &0.47 & \scalebox{0.8}{Finance} \\
    % \cmidrule{2-8}
    % & M4-Monthly & 1 & 18 & (48000, 0, 48000) & Monthly &0.44 & \scalebox{0.8}{Industry} \\
    % \cmidrule{2-8}
    % & M4-Weakly & 1 & 13 & (359, 0, 359) & Weakly &0.43 & \scalebox{0.8}{Macro} \\
    % \cmidrule{2-8}
    %  & M4-Daily & 1 & 14 & (4227, 0, 4227) &Daily &0.44 & \scalebox{0.8}{Micro} \\
    % \cmidrule{2-8}
    %  & M4-Hourly & 1 &48 & (414, 0, 414) & Hourly &0.46 & \scalebox{0.8}{Other} \\
    \bottomrule
    \end{tabular}
     \begin{tablenotes}
        \item $\ast$ The forecastability is calculated by one minus the entropy of Fourier decomposition of time series \citep{goerg2013forecastable}. A larger value indicates better predictability.
    \end{tablenotes}
    \end{small}
  \end{threeparttable}
  }
  \vspace{-5pt}
\end{table*}

\section{Details of Context Generation and Embedding}
\label{app:context_generate}
In this work, we propose to utilize the context information as the anchors in the unified time series coordinate, which is constructed by spectral decomposition. As illustrated in Section~\ref{sec:context_gen}, the used 
\begin{tcolorbox}[title = {global-level description template}] 
\label{box_global}
The \{\textcolor{blue}{dataset name}\} dataset is sourced from the field of \{\textcolor{blue}{domain}\} with a frequency of \{\textcolor{blue}{sampling frequency}\}. The dataset is \{\textcolor{blue}{dataset metadata}\}. The dataset consists of \{\textcolor{blue}{variable numbers}\} variables. The current task objective is to forecast time series over \{\textcolor{blue}{prediction length}\} future time steps using historical time series spanning \{\textcolor{blue}{look-back windows}\} time steps. 
\tcblower 
\textbf{Example: ETTh1-96-96.}\\
The ETTh1 dataset is sourced from the field of Temperature with a frequency of 1 hour. The dataset is an electricity transformer temperature dataset, recording electrical transformers' oil temperature and corresponding external power load features in a region in China between July 2016 and July 2018. The dataset consists of 7 variables. The current task objective is to forecast time series over 96 future time steps using historical time series spanning 96 time steps.
\end{tcolorbox}
\begin{tcolorbox}[title = {variable-level description template}] 
\label{box_variable}
\textbf{prompt to GPT-4o:}\\
``role": ``system", \\
``content": ``You are an expert in the field of \{\textcolor{blue}{domain}\}. Given detailed information about a dataset and the names of the variables in this dataset, you are asked to generate descriptive text for each variable based on your own domain knowledge, containing the meaning of the variable and the interrelationships between the variables, and the text should be one paragraph in length for a variable, with no other extraneous instruction."\\
``role": ``user", \\
``content": ``Here are the dataset details: <\textcolor{blue}{\{messages['dataset']\}}>. The dataset contains the following variables: <\textcolor{blue}{\{messages['var']\}}>. Please generate descriptive text for each variable based on your domain knowledge."\\
\textbf{Query: ETTh1}\\
messages = \{
            ``dataset": ``ETTh1'',
            ``var": [``High UseFul Load", ``High UseLess Load", ``Middle UseFul Load", ``Middle UseLess Load", ``Low UseFul Load", ``Low UseLess Load", ``Oil Temperature"]
        \}
\tcblower 
\textbf{Response:}\\
The ``High UseFul Load" variable refers to the measure of the electrical load at a high voltage level that is actively being utilized to meet consumer demand or operate electrical machinery. ... Its impact on transformer oil temperature, since increased useful load can contribute to higher efficiency ... In the history statistics of this variable, the minimum value is -4.07995, the maximum value is 46.00699, the median value is 11.39599, the mean value is 13.32467, and the overall trend is up.
\end{tcolorbox}
\noindent context includes the \textit{Global-Level Description} and \textit{Variable-Level Description}.

For global-level description, we use a template to generate, which includes the information with: dataset metadata, domain instruction, sampling frequency, and variable numbers. In addition, we appended a description of the current task at the end of the text to assist the model in perceiving between tasks. 

For variable-level description, we prompt a large expert LLM (in our experiment, we use GPT-4o) to generate based on the information of variable names and their general knowledge. In addition, we compute the statics of the history series and append them to the variable description for overall trend modeling. 

For the generated global and variable text corpus, we utilize an open-sourced LLM (in our experiment, we use Sentence-BERT~\cite{reimers2019sentence}) for text embedding.

\section{Full Results in In-domain Setting}
\label{app:in_domain}
\begin{table*}[htbp]
  \caption{Full results for the in-domain long-term forecasting task. We compare extensive competitive models under different prediction lengths. \emph{Avg} is averaged from all four prediction lengths, that is 96, 192, 336, 720.}\label{tab:full_forecasting_results}
  \vskip 0.05in
  \centering
  \resizebox{1.0\textwidth}{!}{
  \begin{threeparttable}
  \begin{small}
  \renewcommand{\multirowsetup}{\centering}
  \setlength{\tabcolsep}{1pt}
  \begin{tabular}{c|c|cc|cc|cc|cc|cc|cc|cc|cc|cc|cc|cc}
    \toprule
    \multicolumn{2}{c}{\multirow{2}{*}{Models}} &
    \multicolumn{2}{c}{\rotatebox{0}{\scalebox{0.8}{\textbf{ContexTST}}}} &
    \multicolumn{2}{c}{\rotatebox{0}{\scalebox{0.8}{TimeXer}}} &
    \multicolumn{2}{c}{\rotatebox{0}{\scalebox{0.8}{TimeMixer}}} &
    \multicolumn{2}{c}{\rotatebox{0}{\scalebox{0.8}{iTransformer}}} &
    \multicolumn{2}{c}{\rotatebox{0}{\scalebox{0.8}{GPT4TS}}} &
    \multicolumn{2}{c}{\rotatebox{0}{\scalebox{0.8}{PatchTST}}} & 
    \multicolumn{2}{c}{\rotatebox{0}{\scalebox{0.8}{DLinear}}} &
    \multicolumn{2}{c}{\rotatebox{0}{\scalebox{0.8}{MICN}}} & 
    \multicolumn{2}{c}{\rotatebox{0}{\scalebox{0.8}{TimesNet}}} & 
    \multicolumn{2}{c}{\rotatebox{0}{\scalebox{0.8}{FEDformer}}} &
    \multicolumn{2}{c}{\rotatebox{0}{\scalebox{0.8}{Autoformer}}} \\
    \multicolumn{2}{c}{} & \multicolumn{2}{c}{\scalebox{0.8}{(\textbf{Ours})}} &
    \multicolumn{2}{c}{\scalebox{0.8}{\citeyear{wang2024timexer}}} &
    \multicolumn{2}{c}{\scalebox{0.8}{\citeyear{wang2024timemixer}}}&
    \multicolumn{2}{c}{\scalebox{0.8}{\citeyear{liu2023itransformer}}}&
    \multicolumn{2}{c}{\scalebox{0.8}{\citeyear{zhou2023one}}}&
    \multicolumn{2}{c}{\scalebox{0.8}{\citeyear{patchtst}}}&
    \multicolumn{2}{c}{\scalebox{0.8}{\citeyear{dlinear}}}&
    \multicolumn{2}{c}{\scalebox{0.8}{\citeyear{wang2023micn}}}&
    \multicolumn{2}{c}{\scalebox{0.8}{\citeyear{wu2022timesnet}}}&
    \multicolumn{2}{c}{\scalebox{0.8}{\citeyear{zhou2022fedformer}}}&
    \multicolumn{2}{c}{\scalebox{0.8}{\citeyear{wu2021autoformer}}}
    \\
    
    \cmidrule(lr){3-4} \cmidrule(lr){5-6}\cmidrule(lr){7-8} \cmidrule(lr){9-10}\cmidrule(lr){11-12}\cmidrule(lr){13-14}\cmidrule(lr){15-16}\cmidrule(lr){17-18}\cmidrule(lr){19-20} \cmidrule(lr){21-22} \cmidrule(lr){23-24}
    \multicolumn{2}{c}{Metric} & \scalebox{0.78}{MSE} & \scalebox{0.78}{MAE} & \scalebox{0.78}{MSE} & \scalebox{0.78}{MAE} & \scalebox{0.78}{MSE} & \scalebox{0.78}{MAE} & \scalebox{0.78}{MSE} & \scalebox{0.78}{MAE} & \scalebox{0.78}{MSE} & \scalebox{0.78}{MAE} & \scalebox{0.78}{MSE} & \scalebox{0.78}{MAE} & \scalebox{0.78}{MSE} & \scalebox{0.78}{MAE} & \scalebox{0.78}{MSE} & \scalebox{0.78}{MAE} & \scalebox{0.78}{MSE} & \scalebox{0.78}{MAE} & \scalebox{0.78}{MSE} & \scalebox{0.78}{MAE} & \scalebox{0.78}{MSE} & \scalebox{0.78}{MAE}\\
    \toprule

 \multirow{5}{*}{\rotatebox{90}{\scalebox{0.95}{ETTh1}}} 
    & \scalebox{0.85}{96} &\boldres{\scalebox{0.85}{0.372}}&\boldres{\scalebox{0.85}{0.395}}&\scalebox{0.85}{0.382}&\scalebox{0.85}{0.403}&\secondres{\scalebox{0.85}{0.375}}&\secondres{\scalebox{0.85}{0.400}}&\scalebox{0.85}{0.386}&\scalebox{0.85}{0.405}&\scalebox{0.85}{0.398}&\scalebox{0.85}{0.424}&\scalebox{0.85}{0.414}&\scalebox{0.85}{0.419}&\scalebox{0.85}{0.388}&\scalebox{0.85}{0.403}&\scalebox{0.85}{0.426}&\scalebox{0.85}{0.446}&\scalebox{0.85}{0.395}&\scalebox{0.85}{0.402}&\scalebox{0.85}{0.395}&\scalebox{0.85}{0.424}&\scalebox{0.85}{0.449}&\scalebox{0.85}{0.459} \\
    
    & \scalebox{0.85}{192} &\boldres{\scalebox{0.85}{0.418}}&\boldres{\scalebox{0.85}{0.421}}&\secondres{\scalebox{0.85}{0.429}}&\scalebox{0.85}{0.435}&\scalebox{0.85}{0.429}&\secondres{\scalebox{0.85}{0.422}}&\scalebox{0.85}{0.441}&\scalebox{0.85}{0.436}&\scalebox{0.85}{0.449}&\scalebox{0.85}{0.427}&\scalebox{0.85}{0.460}&\scalebox{0.85}{0.445}&\scalebox{0.85}{0.437}&\scalebox{0.85}{0.432}&\scalebox{0.85}{0.454}&\scalebox{0.85}{0.464}&\scalebox{0.85}{0.436}&\scalebox{0.85}{0.429}&\scalebox{0.85}{0.469}&\scalebox{0.85}{0.470}&\scalebox{0.85}{0.500}&\scalebox{0.85}{0.482}\\
    
    & \scalebox{0.85}{336} &\boldres{\scalebox{0.85}{0.464}}&\secondres{\scalebox{0.85}{0.452}}&\secondres{\scalebox{0.85}{0.468}}&\boldres{\scalebox{0.85}{0.448}}&\scalebox{0.85}{0.484}&\scalebox{0.85}{0.458}&\scalebox{0.85}{0.487}&\scalebox{0.85}{0.458}&\scalebox{0.85}{0.492}&\scalebox{0.85}{0.466}&\scalebox{0.85}{0.501}&\scalebox{0.85}{0.466}&\scalebox{0.85}{0.481}&\scalebox{0.85}{0.459}&\scalebox{0.85}{0.493}&\scalebox{0.85}{0.487}&\scalebox{0.85}{0.491}&\scalebox{0.85}{0.469}&\scalebox{0.85}{0.530}&\scalebox{0.85}{0.499}&\scalebox{0.85}{0.521}&\scalebox{0.85}{0.496}\\
    
    & \scalebox{0.85}{720} &\boldres{\scalebox{0.85}{0.467}}&\boldres{\scalebox{0.85}{0.458}}&\secondres{\scalebox{0.85}{0.469}}&\secondres{\scalebox{0.85}{0.461}}&\scalebox{0.85}{0.498}&\scalebox{0.85}{0.482}&\scalebox{0.85}{0.503}&\scalebox{0.85}{0.491}&\scalebox{0.85}{0.487}&\scalebox{0.85}{0.483}&\scalebox{0.85}{0.500}&\scalebox{0.85}{0.488}&\scalebox{0.85}{0.519}&\scalebox{0.85}{0.516}&\scalebox{0.85}{0.526}&\scalebox{0.85}{0.526}&\scalebox{0.85}{0.521}&\scalebox{0.85}{0.500}&\scalebox{0.85}{0.598}&\scalebox{0.85}{0.544}&\scalebox{0.85}{0.514}&\scalebox{0.85}{0.512} 
    \\
    \cmidrule(lr){2-24}
    & \scalebox{0.85}{Avg} &\boldres{\scalebox{0.85}{0.430}}&\boldres{\scalebox{0.85}{0.431}}&\secondres{\scalebox{0.85}{0.437}}&\secondres{\scalebox{0.85}{0.437}}&\scalebox{0.85}{0.447}&\scalebox{0.85}{0.441}&\scalebox{0.85}{0.454}&\scalebox{0.85}{0.448}&\scalebox{0.85}{0.457}&\scalebox{0.85}{0.450}&\scalebox{0.85}{0.469}&\scalebox{0.85}{0.455}&\scalebox{0.85}{0.456}&\scalebox{0.85}{0.453}&\scalebox{0.85}{0.475}&\scalebox{0.85}{0.481}&\scalebox{0.85}{0.461}&\scalebox{0.85}{0.450}&\scalebox{0.85}{0.498}&\scalebox{0.85}{0.484}&\scalebox{0.85}{0.496}&\scalebox{0.85}{0.487} \\
    \midrule
    
    \multirow{5}{*}{\rotatebox{90}{\scalebox{0.95}{ETTh2}}} 
    & \scalebox{0.85}{96} &\boldres{\scalebox{0.85}{0.279}}&\boldres{\scalebox{0.85}{0.331}}&\secondres{\scalebox{0.85}{0.286}}&\secondres{\scalebox{0.85}{0.338}}&\scalebox{0.85}{0.289}&\scalebox{0.85}{0.341}&\scalebox{0.85}{0.297}&\scalebox{0.85}{0.349}&\scalebox{0.85}{0.314}&\scalebox{0.85}{0.361}&\scalebox{0.85}{0.308}&\scalebox{0.85}{0.355}&\scalebox{0.85}{0.333}&\scalebox{0.85}{0.387}&\scalebox{0.85}{0.372}&\scalebox{0.85}{0.424}&\scalebox{0.85}{0.340}&\scalebox{0.85}{0.374}&\scalebox{0.85}{0.358}&\scalebox{0.85}{0.397}&\scalebox{0.85}{0.346}&\scalebox{0.85}{0.388}
    \\
    
    & \scalebox{0.85}{192} &\boldres{\scalebox{0.85}{0.360}}&\boldres{\scalebox{0.85}{0.383}}&\secondres{\scalebox{0.85}{0.363}}&\secondres{\scalebox{0.85}{0.389}}&\scalebox{0.85}{0.372}&\scalebox{0.85}{0.392}&\scalebox{0.85}{0.380}&\scalebox{0.85}{0.400}&\scalebox{0.85}{0.407}&\scalebox{0.85}{0.411}&\scalebox{0.85}{0.388}&\scalebox{0.85}{0.405}&\scalebox{0.85}{0.477}&\scalebox{0.85}{0.476}&\scalebox{0.85}{0.492}&\scalebox{0.85}{0.492}&\scalebox{0.85}{0.402}&\scalebox{0.85}{0.414}&\scalebox{0.85}{0.429}&\scalebox{0.85}{0.439}&\scalebox{0.85}{0.456}&\scalebox{0.85}{0.452}
    \\
    & \scalebox{0.85}{336} &\secondres{\scalebox{0.85}{0.407}}&\secondres{\scalebox{0.85}{0.415}}&\scalebox{0.85}{0.414}&\scalebox{0.85}{0.423}&\boldres{\scalebox{0.85}{0.386}}&\boldres{\scalebox{0.85}{0.414}}&\scalebox{0.85}{0.428}&\scalebox{0.85}{0.432}&\scalebox{0.85}{0.437}&\scalebox{0.85}{0.443}&\scalebox{0.85}{0.426}&\scalebox{0.85}{0.433}&\scalebox{0.85}{0.594}&\scalebox{0.85}{0.541}&\scalebox{0.85}{0.607}&\scalebox{0.85}{0.555}&\scalebox{0.85}{0.452}&\scalebox{0.85}{0.452}&\scalebox{0.85}{0.496}&\scalebox{0.85}{0.487}&\scalebox{0.85}{0.482}&\scalebox{0.85}{0.486}\\
    
    & \scalebox{0.85}{720} &\secondres{\scalebox{0.85}{0.415}}&\secondres{\scalebox{0.85}{0.437}}&\scalebox{0.85}{0.418}&\scalebox{0.85}{0.443}&\boldres{\scalebox{0.85}{0.412}}&\boldres{\scalebox{0.85}{0.434}}&\scalebox{0.85}{0.427}&\scalebox{0.85}{0.445}&\scalebox{0.85}{0.434}&\scalebox{0.85}{0.448}&\scalebox{0.85}{0.431}&\scalebox{0.85}{0.446}&\scalebox{0.85}{0.831}&\scalebox{0.85}{0.657}&\scalebox{0.85}{0.824}&\scalebox{0.85}{0.655}&\scalebox{0.85}{0.462}&\scalebox{0.85}{0.468}&\scalebox{0.85}{0.463}&\scalebox{0.85}{0.474}&\scalebox{0.85}{0.515}&\scalebox{0.85}{0.511}
    \\
    
    \cmidrule(lr){2-24}
    & \scalebox{0.85}{Avg} &\boldres{\scalebox{0.85}{0.365}}&\boldres{\scalebox{0.85}{0.392}}&\scalebox{0.85}{0.370}&\scalebox{0.85}{0.398}&\secondres{\scalebox{0.85}{0.365}}&\secondres{\scalebox{0.85}{0.395}}&\scalebox{0.85}{0.383}&\scalebox{0.85}{0.407}&\scalebox{0.85}{0.398}&\scalebox{0.85}{0.416}&\scalebox{0.85}{0.388}&\scalebox{0.85}{0.410}&\scalebox{0.85}{0.559}&\scalebox{0.85}{0.515}&\scalebox{0.85}{0.574}&\scalebox{0.85}{0.532}&\scalebox{0.85}{0.414}&\scalebox{0.85}{0.427}&\scalebox{0.85}{0.436}&\scalebox{0.85}{0.449}&\scalebox{0.85}{0.450}&\scalebox{0.85}{0.459}\\
    \midrule

    \multirow{5}{*}{\rotatebox{90}{\scalebox{0.95}{ETTm1}}} 
    & \scalebox{0.85}{96} &\boldres{\scalebox{0.85}{0.317}}&\boldres{\scalebox{0.85}{0.346}}&\secondres{\scalebox{0.85}{0.318}}&\secondres{\scalebox{0.85}{0.356}}&\scalebox{0.85}{0.320}&\scalebox{0.85}{0.357}&\scalebox{0.85}{0.334}&\scalebox{0.85}{0.368}&\scalebox{0.85}{0.335}&\scalebox{0.85}{0.369}&\scalebox{0.85}{0.339}&\scalebox{0.85}{0.367}&\scalebox{0.85}{0.345}&\scalebox{0.85}{0.372}&\scalebox{0.85}{0.365}&\scalebox{0.85}{0.387}&\scalebox{0.85}{0.338}&\scalebox{0.85}{0.375}&\scalebox{0.85}{0.379}&\scalebox{0.85}{0.419}&\scalebox{0.85}{0.505}&\scalebox{0.85}{0.475}\\
    
    & \scalebox{0.85}{192} &\boldres{\scalebox{0.85}{0.353}}&\boldres{\scalebox{0.85}{0.379}}&\scalebox{0.85}{0.362}&\scalebox{0.85}{0.383}&\secondres{\scalebox{0.85}{0.361}}&\secondres{\scalebox{0.85}{0.381}}&\scalebox{0.85}{0.387}&\scalebox{0.85}{0.391}&\scalebox{0.85}{0.374}&\scalebox{0.85}{0.385}&\scalebox{0.85}{0.378}&\scalebox{0.85}{0.385}&\scalebox{0.85}{0.380}&\scalebox{0.85}{0.389}&\scalebox{0.85}{0.403}&\scalebox{0.85}{0.408}&\scalebox{0.85}{0.374}&\scalebox{0.85}{0.387}&\scalebox{0.85}{0.426}&\scalebox{0.85}{0.441}&\scalebox{0.85}{0.553}&\scalebox{0.85}{0.496}\\
    
    & \scalebox{0.85}{336} &\scalebox{0.85}{0.401}&\scalebox{0.85}{0.408}&\secondres{\scalebox{0.85}{0.395}}&\scalebox{0.85}{0.407}&\boldres{\scalebox{0.85}{0.390}}&\boldres{\scalebox{0.85}{0.404}}&\scalebox{0.85}{0.426}&\scalebox{0.85}{0.432}&\scalebox{0.85}{0.407}&\secondres{\scalebox{0.85}{0.406}}&\scalebox{0.85}{0.399}&\scalebox{0.85}{0.410}&\scalebox{0.85}{0.413}&\scalebox{0.85}{0.413}&\scalebox{0.85}{0.436}&\scalebox{0.85}{0.431}&\scalebox{0.85}{0.410}&\scalebox{0.85}{0.421}&\scalebox{0.85}{0.445}&\scalebox{0.85}{0.459}&\scalebox{0.85}{0.621}&\scalebox{0.85}{0.537}\\
    
    & \scalebox{0.85}{720} &\boldres{\scalebox{0.85}{0.448}}&\scalebox{0.85}{0.443}&\secondres{\scalebox{0.85}{0.452}}&\scalebox{0.85}{0.465}&\scalebox{0.85}{0.454}&\boldres{\scalebox{0.85}{0.441}}&\scalebox{0.85}{0.491}&\scalebox{0.85}{0.489}&\scalebox{0.85}{0.469}&\secondres{\scalebox{0.85}{0.442}}&\scalebox{0.85}{0.474}&\scalebox{0.85}{0.482}&\scalebox{0.85}{0.474}&\scalebox{0.85}{0.493}&\scalebox{0.85}{0.489}&\scalebox{0.85}{0.462}&\scalebox{0.85}{0.478}&\scalebox{0.85}{0.498}&\scalebox{0.85}{0.543}&\scalebox{0.85}{0.490}&\scalebox{0.85}{0.671}&\scalebox{0.85}{0.561}
    \\
    \cmidrule(lr){2-24}
    & \scalebox{0.85}{Avg} &\boldres{\scalebox{0.85}{0.380}}&\boldres{\scalebox{0.85}{0.394}}&\scalebox{0.85}{0.382}&\scalebox{0.85}{0.403}&\secondres{\scalebox{0.85}{0.381}}&\secondres{\scalebox{0.85}{0.396}}&\scalebox{0.85}{0.409}&\scalebox{0.85}{0.420}&\scalebox{0.85}{0.396}&\scalebox{0.85}{0.401}&\scalebox{0.85}{0.398}&\scalebox{0.85}{0.411}&\scalebox{0.85}{0.403}&\scalebox{0.85}{0.417}&\scalebox{0.85}{0.423}&\scalebox{0.85}{0.422}&\scalebox{0.85}{0.400}&\scalebox{0.85}{0.420}&\scalebox{0.85}{0.448}&\scalebox{0.85}{0.452}&\scalebox{0.85}{0.588}&\scalebox{0.85}{0.517}\\
    \midrule

    \multirow{5}{*}{\rotatebox{90}{\scalebox{0.95}{ETTm2}}} 
    & \scalebox{0.85}{96} &\secondres{\scalebox{0.85}{0.172}}&\secondres{\scalebox{0.85}{0.257}}&\boldres{\scalebox{0.85}{0.171}}&\boldres{\scalebox{0.85}{0.256}}&\scalebox{0.85}{0.175}&\scalebox{0.85}{0.258}&\scalebox{0.85}{0.180}&\scalebox{0.85}{0.264}&\scalebox{0.85}{0.190}&\scalebox{0.85}{0.275}&\scalebox{0.85}{0.175}&\scalebox{0.85}{0.259}&\scalebox{0.85}{0.193}&\scalebox{0.85}{0.292}&\scalebox{0.85}{0.197}&\scalebox{0.85}{0.296}&\scalebox{0.85}{0.187}&\scalebox{0.85}{0.267}&\scalebox{0.85}{0.203}&\scalebox{0.85}{0.287}&\scalebox{0.85}{0.255}&\scalebox{0.85}{0.339}
    \\
    
    & \scalebox{0.85}{192} &\boldres{\scalebox{0.85}{0.232}}&\boldres{\scalebox{0.85}{0.291}}&\secondres{\scalebox{0.85}{0.237}}&\secondres{\scalebox{0.85}{0.299}}&\scalebox{0.85}{0.237}&\scalebox{0.85}{0.299}&\scalebox{0.85}{0.250}&\scalebox{0.85}{0.309}&\scalebox{0.85}{0.253}&\scalebox{0.85}{0.313}&\scalebox{0.85}{0.241}&\scalebox{0.85}{0.302}&\scalebox{0.85}{0.284}&\scalebox{0.85}{0.362}&\scalebox{0.85}{0.284}&\scalebox{0.85}{0.361}&\scalebox{0.85}{0.249}&\scalebox{0.85}{0.309}&\scalebox{0.85}{0.269}&\scalebox{0.85}{0.328}&\scalebox{0.85}{0.281}&\scalebox{0.85}{0.340}
    \\
    & \scalebox{0.85}{336} &\scalebox{0.85}{0.308}&\scalebox{0.85}{0.344}&\boldres{\scalebox{0.85}{0.296}}&\boldres{\scalebox{0.85}{0.338}}&\secondres{\scalebox{0.85}{0.298}}&\secondres{\scalebox{0.85}{0.340}}&\scalebox{0.85}{0.311}&\scalebox{0.85}{0.348}&\scalebox{0.85}{0.321}&\scalebox{0.85}{0.360}&\scalebox{0.85}{0.305}&\scalebox{0.85}{0.343}&\scalebox{0.85}{0.369}&\scalebox{0.85}{0.427}&\scalebox{0.85}{0.381}&\scalebox{0.85}{0.429}&\scalebox{0.85}{0.321}&\scalebox{0.85}{0.351}&\scalebox{0.85}{0.325}&\scalebox{0.85}{0.366}&\scalebox{0.85}{0.339}&\scalebox{0.85}{0.372}
    \\
    & \scalebox{0.85}{720} &\scalebox{0.85}{0.404}&\secondres{\scalebox{0.85}{0.400}}&\boldres{\scalebox{0.85}{0.392}}&\boldres{\scalebox{0.85}{0.394}}&\scalebox{0.85}{0.409}&\scalebox{0.85}{0.406}&\scalebox{0.85}{0.412}&\scalebox{0.85}{0.407}&\scalebox{0.85}{0.411}&\scalebox{0.85}{0.406}&\secondres{\scalebox{0.85}{0.402}}&\scalebox{0.85}{0.400}&\scalebox{0.85}{0.554}&\scalebox{0.85}{0.552}&\scalebox{0.85}{0.549}&\scalebox{0.85}{0.522}&\scalebox{0.85}{0.408}&\scalebox{0.85}{0.403}&\scalebox{0.85}{0.421}&\scalebox{0.85}{0.415}&\scalebox{0.85}{0.433}&\scalebox{0.85}{0.432} 
    \\
    \cmidrule(lr){2-24}
    & \scalebox{0.85}{Avg} &\secondres{\scalebox{0.85}{0.279}}&\secondres{\scalebox{0.85}{0.323}}&\boldres{\scalebox{0.85}{0.274}}&\boldres{\scalebox{0.85}{0.322}}&\scalebox{0.85}{0.280}&\scalebox{0.85}{0.326}&\scalebox{0.85}{0.288}&\scalebox{0.85}{0.332}&\scalebox{0.85}{0.294}&\scalebox{0.85}{0.339}&\scalebox{0.85}{0.281}&\scalebox{0.85}{0.326}&\scalebox{0.85}{0.350}&\scalebox{0.85}{0.408}&\scalebox{0.85}{0.353}&\scalebox{0.85}{0.402}&\scalebox{0.85}{0.291}&\scalebox{0.85}{0.333}&\scalebox{0.85}{0.304}&\scalebox{0.85}{0.349}&\scalebox{0.85}{0.327}&\scalebox{0.85}{0.371}\\
    \midrule

    \multirow{5}{*}{\rotatebox{90}{\scalebox{0.95}{Electricity}}} 
    & \scalebox{0.85}{96} &\boldres{\scalebox{0.85}{0.144}}&\boldres{\scalebox{0.85}{0.234}}&\scalebox{0.85}{0.158}&\scalebox{0.85}{0.259}&\scalebox{0.85}{0.153}&\scalebox{0.85}{0.247}&\secondres{\scalebox{0.85}{0.148}}&\secondres{\scalebox{0.85}{0.243}}&\scalebox{0.85}{0.197}&\scalebox{0.85}{0.290}&\scalebox{0.85}{0.195}&\scalebox{0.85}{0.285}&\scalebox{0.85}{0.197}&\scalebox{0.85}{0.282}&\scalebox{0.85}{0.180}&\scalebox{0.85}{0.293}&\scalebox{0.85}{0.168}&\scalebox{0.85}{0.272}&\scalebox{0.85}{0.193}&\scalebox{0.85}{0.308}&\scalebox{0.85}{0.201}&\scalebox{0.85}{0.317}\\
    
    & \scalebox{0.85}{192} &\boldres{\scalebox{0.85}{0.161}}&\secondres{\scalebox{0.85}{0.259}}&\scalebox{0.85}{0.173}&\scalebox{0.85}{0.271}&\scalebox{0.85}{0.166}&\boldres{\scalebox{0.85}{0.256}}&\secondres{\scalebox{0.85}{0.165}}&\scalebox{0.85}{0.264}&\scalebox{0.85}{0.201}&\scalebox{0.85}{0.292}&\scalebox{0.85}{0.199}&\scalebox{0.85}{0.289}&\scalebox{0.85}{0.206}&\scalebox{0.85}{0.285}&\scalebox{0.85}{0.189}&\scalebox{0.85}{0.302}&\scalebox{0.85}{0.184}&\scalebox{0.85}{0.289}&\scalebox{0.85}{0.201}&\scalebox{0.85}{0.315}&\scalebox{0.85}{0.222}&\scalebox{0.85}{0.334}
    \\
    & \scalebox{0.85}{336} &\boldres{\scalebox{0.85}{0.177}}&\boldres{\scalebox{0.85}{0.277}}&\scalebox{0.85}{0.191}&\scalebox{0.85}{0.288}&\secondres{\scalebox{0.85}{0.185}}&\secondres{\scalebox{0.85}{0.277}}&\scalebox{0.85}{0.187}&\scalebox{0.85}{0.296}&\scalebox{0.85}{0.217}&\scalebox{0.85}{0.309}&\scalebox{0.85}{0.215}&\scalebox{0.85}{0.305}&\scalebox{0.85}{0.219}&\scalebox{0.85}{0.301}&\scalebox{0.85}{0.198}&\scalebox{0.85}{0.312}&\scalebox{0.85}{0.198}&\scalebox{0.85}{0.300}&\scalebox{0.85}{0.214}&\scalebox{0.85}{0.329}&\scalebox{0.85}{0.231}&\scalebox{0.85}{0.338}
    \\
    & \scalebox{0.85}{720} &\scalebox{0.85}{0.224}&\secondres{\scalebox{0.85}{0.311}}&\scalebox{0.85}{0.228}&\scalebox{0.85}{0.318}&\scalebox{0.85}{0.225}&\boldres{\scalebox{0.85}{0.310}}&\scalebox{0.85}{0.255}&\scalebox{0.85}{0.327}&\scalebox{0.85}{0.253}&\scalebox{0.85}{0.339}&\scalebox{0.85}{0.256}&\scalebox{0.85}{0.337}&\scalebox{0.85}{0.245}&\scalebox{0.85}{0.333}&\boldres{\scalebox{0.85}{0.217}}&\scalebox{0.85}{0.330}&\secondres{\scalebox{0.85}{0.220}}&\scalebox{0.85}{0.320}&\scalebox{0.85}{0.246}&\scalebox{0.85}{0.355}&\scalebox{0.85}{0.254}&\scalebox{0.85}{0.361}
    \\
    \cmidrule(lr){2-24}
    & \scalebox{0.85}{Avg} &\boldres{\scalebox{0.85}{0.176}}&\boldres{\scalebox{0.85}{0.270}}&\scalebox{0.85}{0.188}&\scalebox{0.85}{0.284}&\secondres{\scalebox{0.85}{0.182}}&\secondres{\scalebox{0.85}{0.273}}&\scalebox{0.85}{0.189}&\scalebox{0.85}{0.282}&\scalebox{0.85}{0.217}&\scalebox{0.85}{0.307}&\scalebox{0.85}{0.216}&\scalebox{0.85}{0.304}&\scalebox{0.85}{0.217}&\scalebox{0.85}{0.300}&\scalebox{0.85}{0.196}&\scalebox{0.85}{0.309}&\scalebox{0.85}{0.193}&\scalebox{0.85}{0.295}&\scalebox{0.85}{0.213}&\scalebox{0.85}{0.327}&\scalebox{0.85}{0.227}&\scalebox{0.85}{0.338}\\
    \midrule

    \multirow{5}{*}{\rotatebox{90}{\scalebox{0.95}{Weather}}} 
    & \scalebox{0.85}{96} &\secondres{\scalebox{0.85}{0.158}}&\boldres{\scalebox{0.85}{0.202}}&\boldres{\scalebox{0.85}{0.157}}&\secondres{\scalebox{0.85}{0.205}}&\scalebox{0.85}{0.163}&\scalebox{0.85}{0.209}&\scalebox{0.85}{0.174}&\scalebox{0.85}{0.214}&\scalebox{0.85}{0.203}&\scalebox{0.85}{0.244}&\scalebox{0.85}{0.177}&\scalebox{0.85}{0.218}&\scalebox{0.85}{0.196}&\scalebox{0.85}{0.255}&\scalebox{0.85}{0.198}&\scalebox{0.85}{0.261}&\scalebox{0.85}{0.182}&\scalebox{0.85}{0.223}&\scalebox{0.85}{0.217}&\scalebox{0.85}{0.296}&\scalebox{0.85}{0.266}&\scalebox{0.85}{0.336} \\
    
    & \scalebox{0.85}{192} &\secondres{\scalebox{0.85}{0.207}}&\boldres{\scalebox{0.85}{0.246}}&\boldres{\scalebox{0.85}{0.205}}&\secondres{\scalebox{0.85}{0.250}}&\scalebox{0.85}{0.208}&\scalebox{0.85}{0.250}&\scalebox{0.85}{0.221}&\scalebox{0.85}{0.254}&\scalebox{0.85}{0.247}&\scalebox{0.85}{0.277}&\scalebox{0.85}{0.225}&\scalebox{0.85}{0.259}&\scalebox{0.85}{0.237}&\scalebox{0.85}{0.296}&\scalebox{0.85}{0.239}&\scalebox{0.85}{0.299}&\scalebox{0.85}{0.219}&\scalebox{0.85}{0.261}&\scalebox{0.85}{0.276}&\scalebox{0.85}{0.336}&\scalebox{0.85}{0.307}&\scalebox{0.85}{0.367}\\
    
    & \scalebox{0.85}{336} &\secondres{\scalebox{0.85}{0.263}}&\secondres{\scalebox{0.85}{0.288}}&\scalebox{0.85}{0.263}&\scalebox{0.85}{0.292}&\boldres{\scalebox{0.85}{0.251}}&\boldres{\scalebox{0.85}{0.287}}&\scalebox{0.85}{0.278}&\scalebox{0.85}{0.296}&\scalebox{0.85}{0.297}&\scalebox{0.85}{0.311}&\scalebox{0.85}{0.278}&\scalebox{0.85}{0.297}&\scalebox{0.85}{0.283}&\scalebox{0.85}{0.335}&\scalebox{0.85}{0.285}&\scalebox{0.85}{0.336}&\scalebox{0.85}{0.280}&\scalebox{0.85}{0.306}&\scalebox{0.85}{0.339}&\scalebox{0.85}{0.380}&\scalebox{0.85}{0.359}&\scalebox{0.85}{0.395}\\
    
    & \scalebox{0.85}{720} &\boldres{\scalebox{0.85}{0.332}}&\boldres{\scalebox{0.85}{0.327}}&\scalebox{0.85}{0.340}&\secondres{\scalebox{0.85}{0.341}}&\secondres{\scalebox{0.85}{0.339}}&\scalebox{0.85}{0.341}&\scalebox{0.85}{0.358}&\scalebox{0.85}{0.349}&\scalebox{0.85}{0.368}&\scalebox{0.85}{0.356}&\scalebox{0.85}{0.354}&\scalebox{0.85}{0.348}&\scalebox{0.85}{0.345}&\scalebox{0.85}{0.381}&\scalebox{0.85}{0.351}&\scalebox{0.85}{0.388}&\scalebox{0.85}{0.365}&\scalebox{0.85}{0.359}&\scalebox{0.85}{0.403}&\scalebox{0.85}{0.428}&\scalebox{0.85}{0.419}&\scalebox{0.85}{0.428}
    \\
    \cmidrule(lr){2-24}
    & \scalebox{0.85}{Avg} &\boldres{\scalebox{0.85}{0.240}}&\boldres{\scalebox{0.85}{0.266}}&\scalebox{0.85}{0.241}&\scalebox{0.85}{0.272}&\secondres{\scalebox{0.85}{0.240}}&\secondres{\scalebox{0.85}{0.272}}&\scalebox{0.85}{0.258}&\scalebox{0.85}{0.278}&\scalebox{0.85}{0.279}&\scalebox{0.85}{0.297}&\scalebox{0.85}{0.259}&\scalebox{0.85}{0.280}&\scalebox{0.85}{0.265}&\scalebox{0.85}{0.317}&\scalebox{0.85}{0.268}&\scalebox{0.85}{0.321}&\scalebox{0.85}{0.262}&\scalebox{0.85}{0.287}&\scalebox{0.85}{0.309}&\scalebox{0.85}{0.360}&\scalebox{0.85}{0.338}&\scalebox{0.85}{0.382}\\
    \midrule

    \multirow{5}{*}{\rotatebox{90}{\scalebox{0.95}{Traffic}}} 
    & \scalebox{0.85}{96} &\secondres{\scalebox{0.85}{0.424}}&\scalebox{0.85}{0.294}&\scalebox{0.85}{0.428}&\scalebox{0.85}{0.297}&\scalebox{0.85}{0.462}&\secondres{\scalebox{0.85}{0.285}}&\boldres{\scalebox{0.85}{0.395}}&\boldres{\scalebox{0.85}{0.268}}&\scalebox{0.85}{0.467}&\scalebox{0.85}{0.313}&\scalebox{0.85}{0.462}&\scalebox{0.85}{0.295}&\scalebox{0.85}{0.650}&\scalebox{0.85}{0.396}&\scalebox{0.85}{0.577}&\scalebox{0.85}{0.350}&\scalebox{0.85}{0.593}&\scalebox{0.85}{0.321}&\scalebox{0.85}{0.587}&\scalebox{0.85}{0.366}&\scalebox{0.85}{0.613}&\scalebox{0.85}{0.388} 
    \\
    
    & \scalebox{0.85}{192} &\secondres{\scalebox{0.85}{0.442}}&\scalebox{0.85}{0.301}&\scalebox{0.85}{0.448}&\scalebox{0.85}{0.304}&\scalebox{0.85}{0.473}&\scalebox{0.85}{0.303}&\boldres{\scalebox{0.85}{0.417}}&\boldres{\scalebox{0.85}{0.276}}&\scalebox{0.85}{0.478}&\scalebox{0.85}{0.352}&\scalebox{0.85}{0.466}&\secondres{\scalebox{0.85}{0.296}}&\scalebox{0.85}{0.598}&\scalebox{0.85}{0.370}&\scalebox{0.85}{0.589}&\scalebox{0.85}{0.356}&\scalebox{0.85}{0.617}&\scalebox{0.85}{0.336}&\scalebox{0.85}{0.604}&\scalebox{0.85}{0.373}&\scalebox{0.85}{0.616}&\scalebox{0.85}{0.382}
    \\
    & \scalebox{0.85}{336} &\secondres{\scalebox{0.85}{0.456}}&\scalebox{0.85}{0.306}&\scalebox{0.85}{0.473}&\scalebox{0.85}{0.309}&\scalebox{0.85}{0.498}&\scalebox{0.85}{0.306}&\boldres{\scalebox{0.85}{0.433}}&\boldres{\scalebox{0.85}{0.283}}&\scalebox{0.85}{0.491}&\scalebox{0.85}{0.317}&\scalebox{0.85}{0.482}&\secondres{\scalebox{0.85}{0.304}}&\scalebox{0.85}{0.605}&\scalebox{0.85}{0.373}&\scalebox{0.85}{0.594}&\scalebox{0.85}{0.358}&\scalebox{0.85}{0.629}&\scalebox{0.85}{0.363}&\scalebox{0.85}{0.621}&\scalebox{0.85}{0.383}&\scalebox{0.85}{0.622}&\scalebox{0.85}{0.337}
    \\
    & \scalebox{0.85}{720} &\secondres{\scalebox{0.85}{0.493}}&\scalebox{0.85}{0.326}&\scalebox{0.85}{0.516}&\scalebox{0.85}{0.327}&\scalebox{0.85}{0.506}&\scalebox{0.85}{0.334}&\boldres{\scalebox{0.85}{0.467}}&\boldres{\scalebox{0.85}{0.302}}&\scalebox{0.85}{0.525}&\scalebox{0.85}{0.337}&\scalebox{0.85}{0.514}&\secondres{\scalebox{0.85}{0.322}}&\scalebox{0.85}{0.645}&\scalebox{0.85}{0.394}&\scalebox{0.85}{0.613}&\scalebox{0.85}{0.361}&\scalebox{0.85}{0.640}&\scalebox{0.85}{0.375}&\scalebox{0.85}{0.626}&\scalebox{0.85}{0.382}&\scalebox{0.85}{0.660}&\scalebox{0.85}{0.408}
    \\
    \cmidrule(lr){2-24}
    & \scalebox{0.85}{Avg} &\secondres{\scalebox{0.85}{0.454}}&\scalebox{0.85}{0.307}&\scalebox{0.85}{0.466}&\scalebox{0.85}{0.309}&\scalebox{0.85}{0.485}&\scalebox{0.85}{0.307}&\boldres{\scalebox{0.85}{0.428}}&\boldres{\scalebox{0.85}{0.282}}&\scalebox{0.85}{0.490}&\scalebox{0.85}{0.330}&\scalebox{0.85}{0.481}&\secondres{\scalebox{0.85}{0.304}}&\scalebox{0.85}{0.625}&\scalebox{0.85}{0.383}&\scalebox{0.85}{0.593}&\scalebox{0.85}{0.356}&\scalebox{0.85}{0.620}&\scalebox{0.85}{0.349}&\scalebox{0.85}{0.609}&\scalebox{0.85}{0.376}&\scalebox{0.85}{0.628}&\scalebox{0.85}{0.379}\\
    \midrule
    \multicolumn{2}{c|}{\rotatebox{0}{\scalebox{0.9}{$1^{\text{st}}$ Count}}} & \boldres{\scalebox{0.85}{19}} & \boldres{\scalebox{0.85}{19}} & \secondres{\scalebox{0.85}{6}} & \scalebox{0.85}{5} & \scalebox{0.85}{4} & \secondres{\scalebox{0.85}{7}} & \scalebox{0.85}{5} & \scalebox{0.85}{5} & \scalebox{0.85}{0} & \scalebox{0.85}{0} & \scalebox{0.85}{0} & \scalebox{0.85}{0} & \scalebox{0.85}{0} & \scalebox{0.85}{0} & \scalebox{0.85}{1} & \scalebox{0.85}{0} & \scalebox{0.85}{0} & \scalebox{0.85}{0} & \scalebox{0.85}{0} & \scalebox{0.85}{0} & \scalebox{0.85}{0} & \scalebox{0.85}{0} \\
    
    % \midrule
    
    % \multicolumn{2}{c}{\scalebox{0.85}{{$1^{\text{st}}$ Count}}}&\scalebox{0.85}{65} \\
    \bottomrule
  \end{tabular}
    \end{small}
  \end{threeparttable}
  }
\end{table*}

\begin{figure*}[th]
    \centering
    
    \subfigure[ETTm1 (MSE)]{\includegraphics[width=0.24\textwidth]{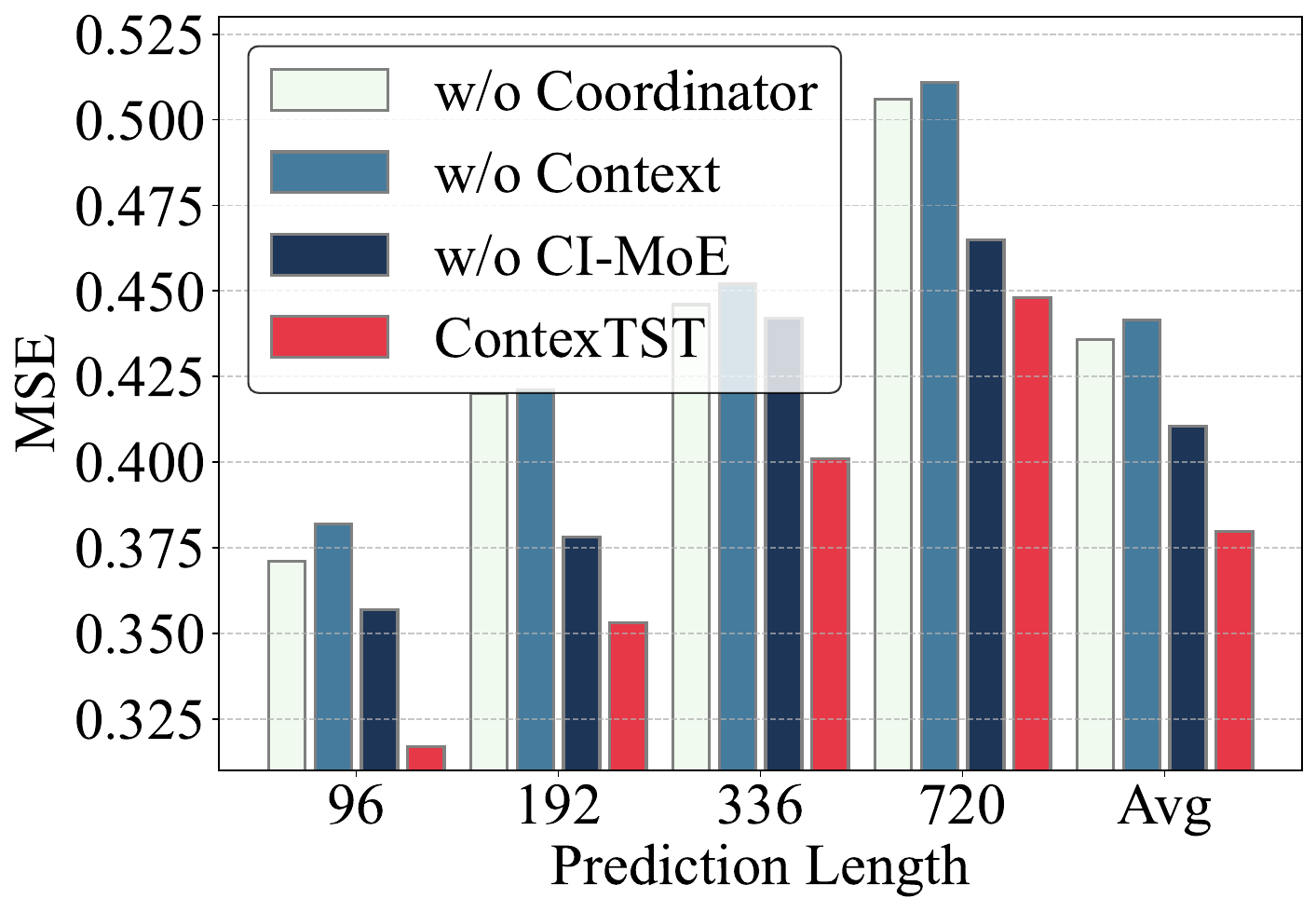}\label{fig:ab_ettm1_mse}}
    \hfill
    \subfigure[Electricity (MSE)]{\includegraphics[width=0.24\textwidth]{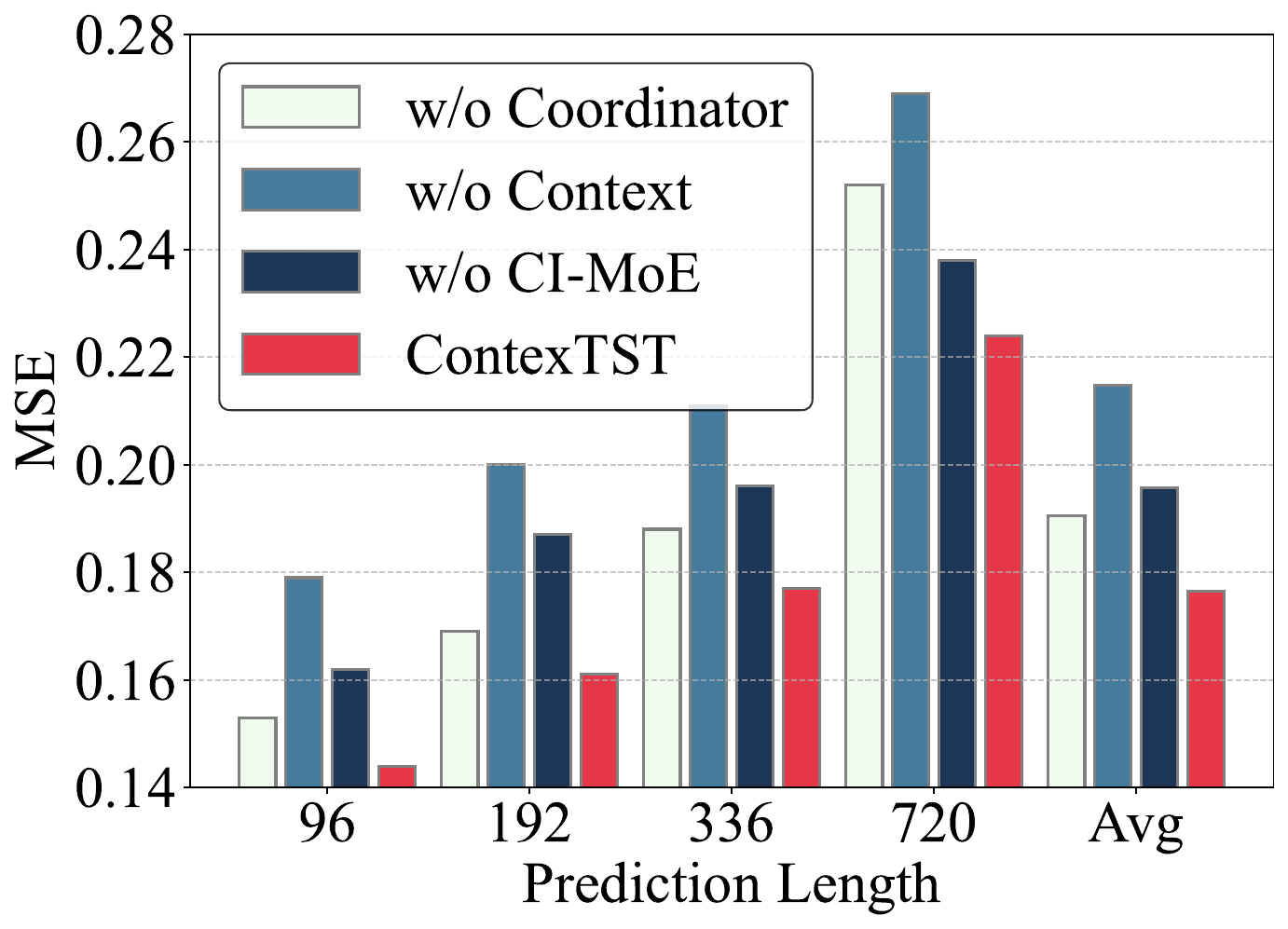}\label{fig:ab_electricity_mse}}
    \hfill
    \subfigure[Weather (MSE)]{\includegraphics[width=0.24\textwidth]{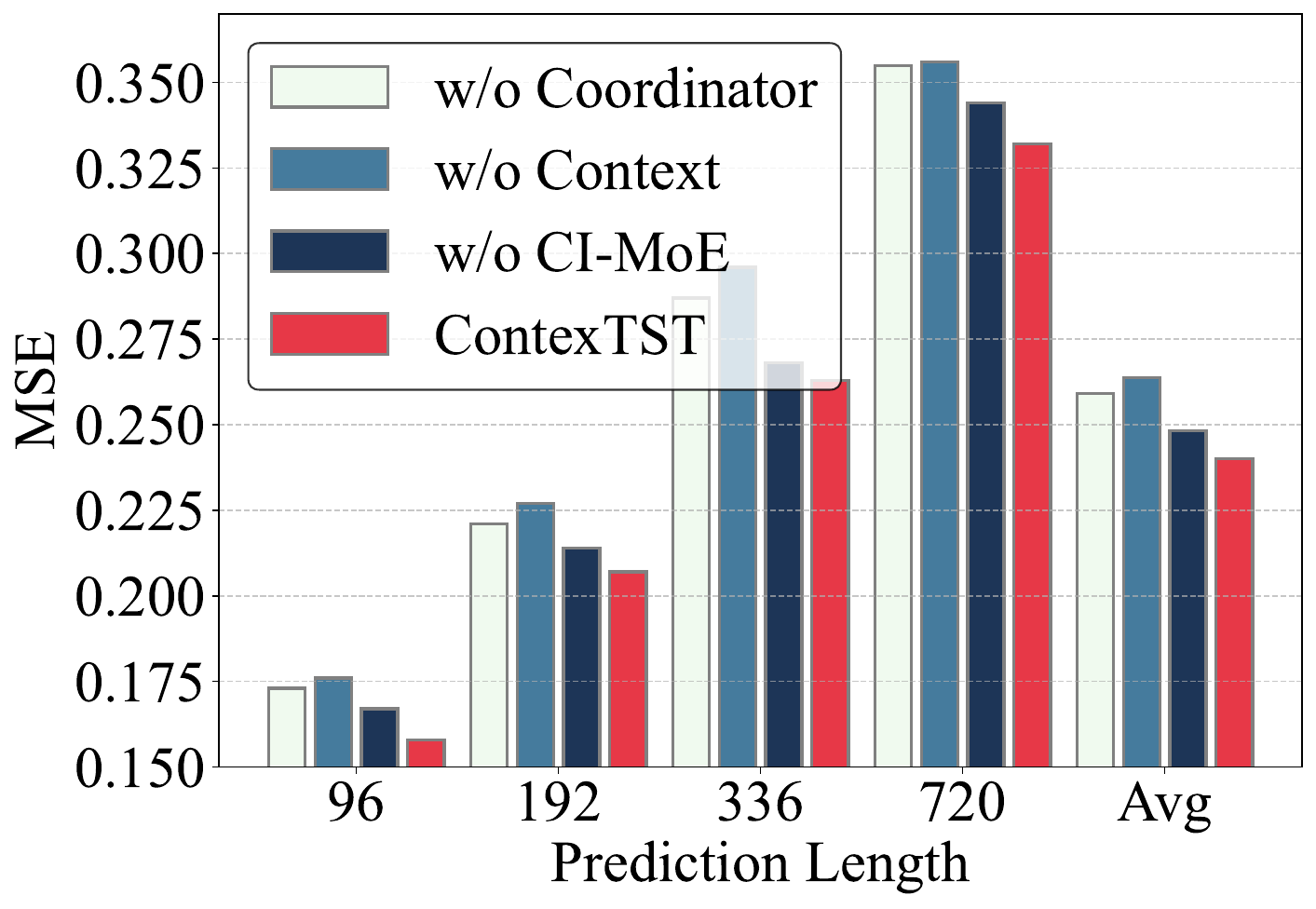}\label{fig:ab_weather_mse}}
    \hfill
    \subfigure[Traffic (MSE)]{\includegraphics[width=0.24\textwidth]{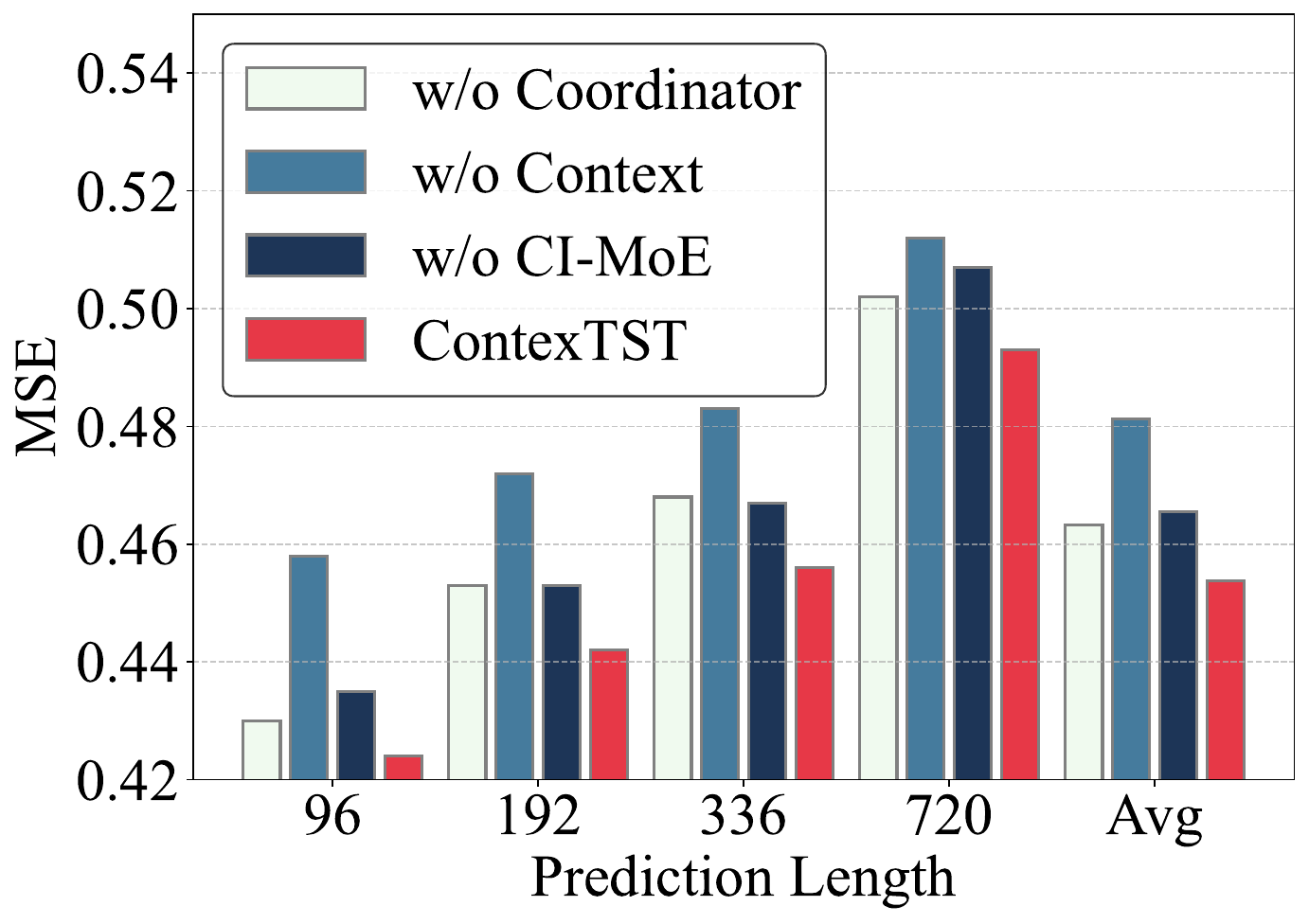}\label{fig:ab_traffic_mse}}\\

    \subfigure[ETTm1 (MAE)]{\includegraphics[width=0.24\textwidth]{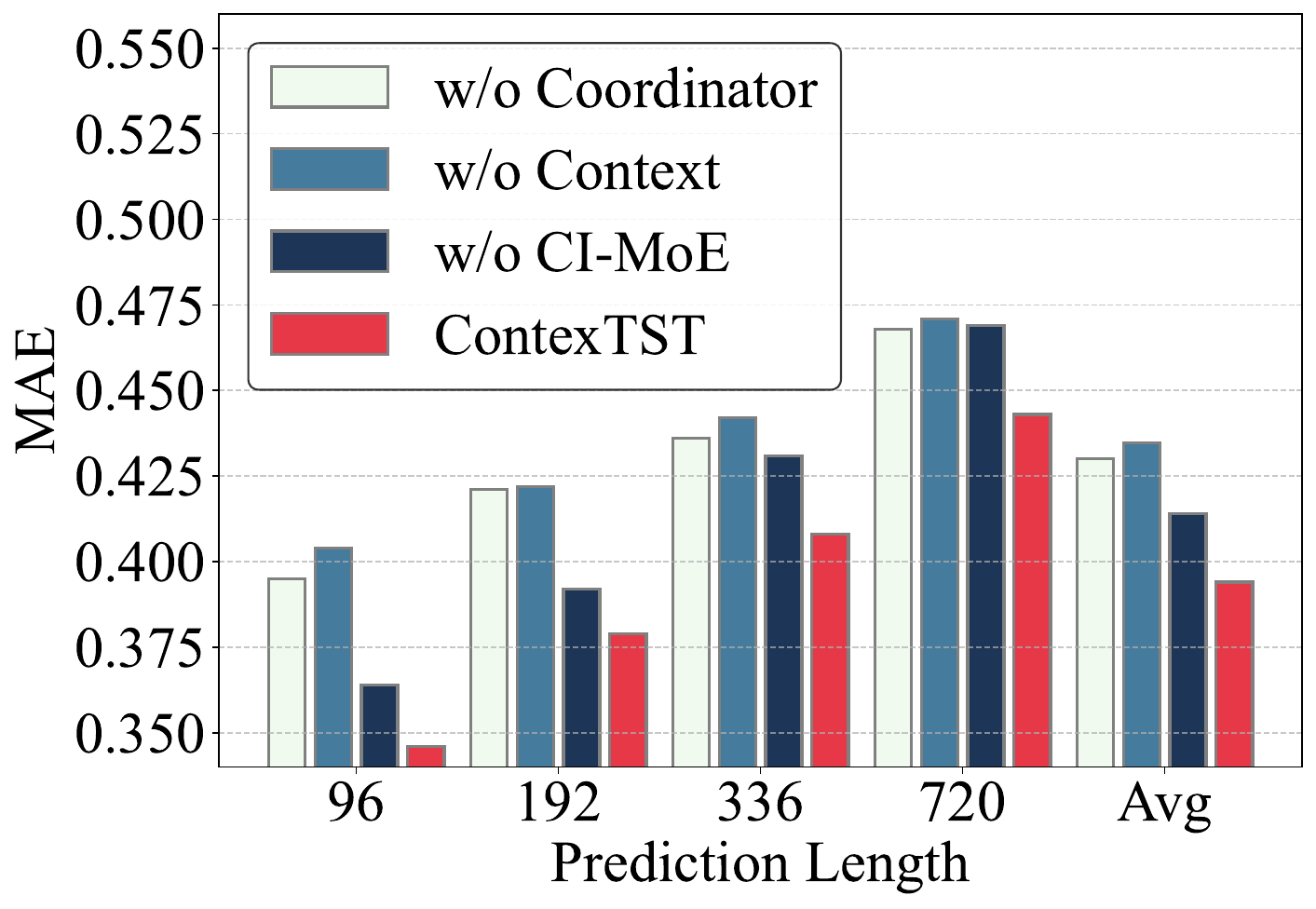}\label{fig:ab_ettm1_mae}}
    \hfill
    \subfigure[Electricity (MAE)]{\includegraphics[width=0.24\textwidth]{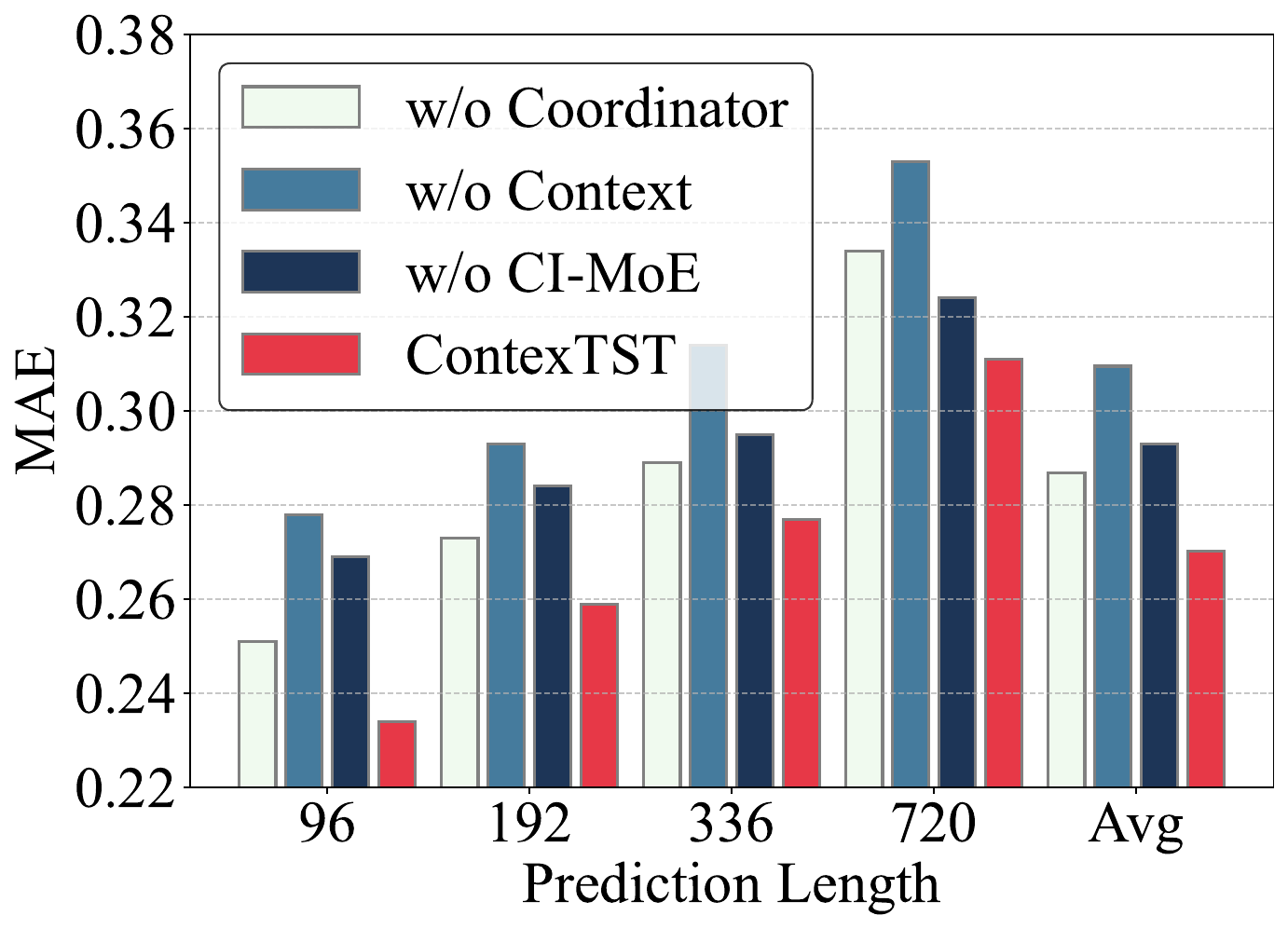}\label{fig:ab_electricity_mae}}
    \hfill
    \subfigure[Weather (MAE)]{\includegraphics[width=0.24\textwidth]{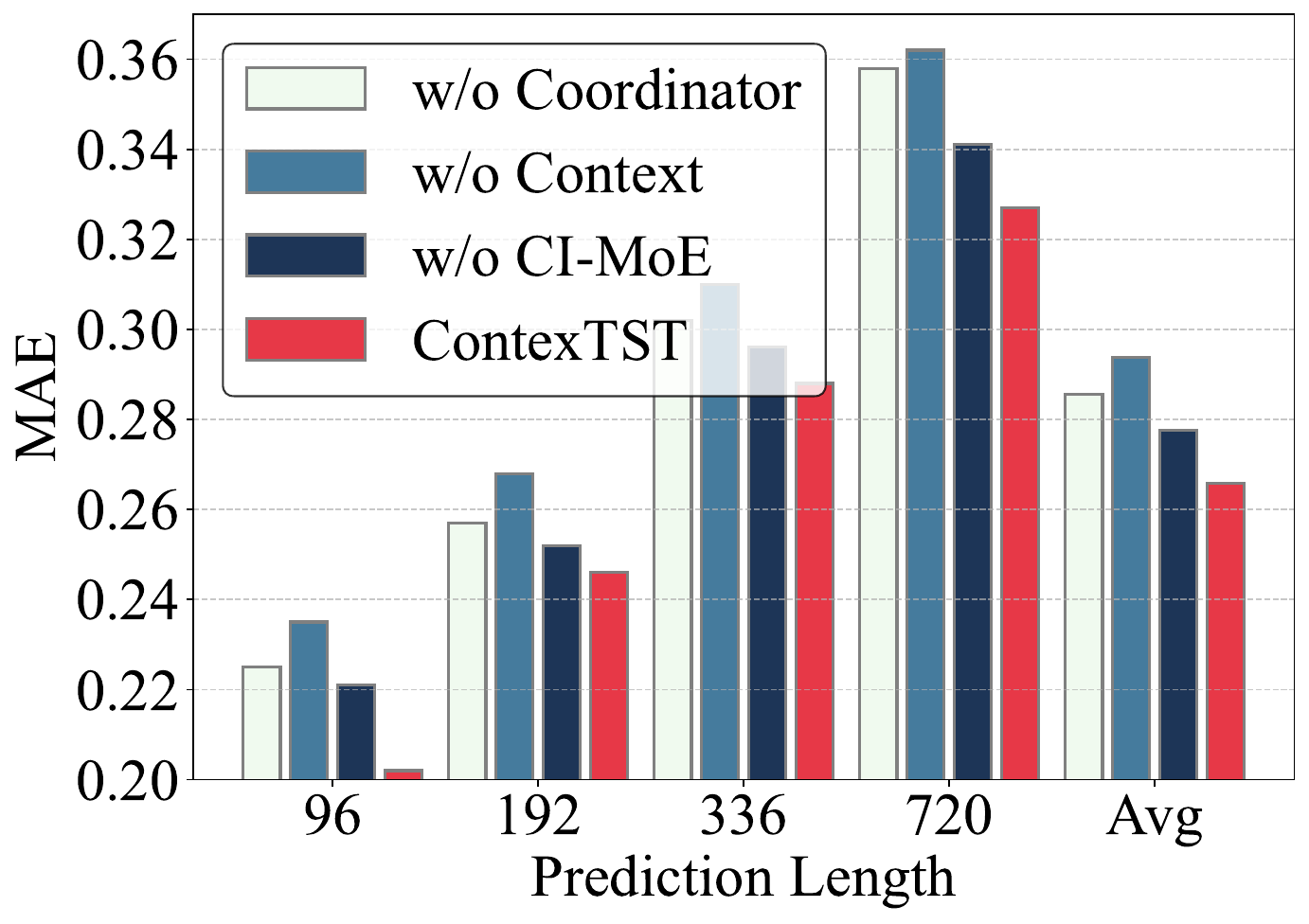}\label{fig:ab_weather_mae}}
  \subfigure[Traffic (MAE)]{\includegraphics[width=0.24\textwidth]{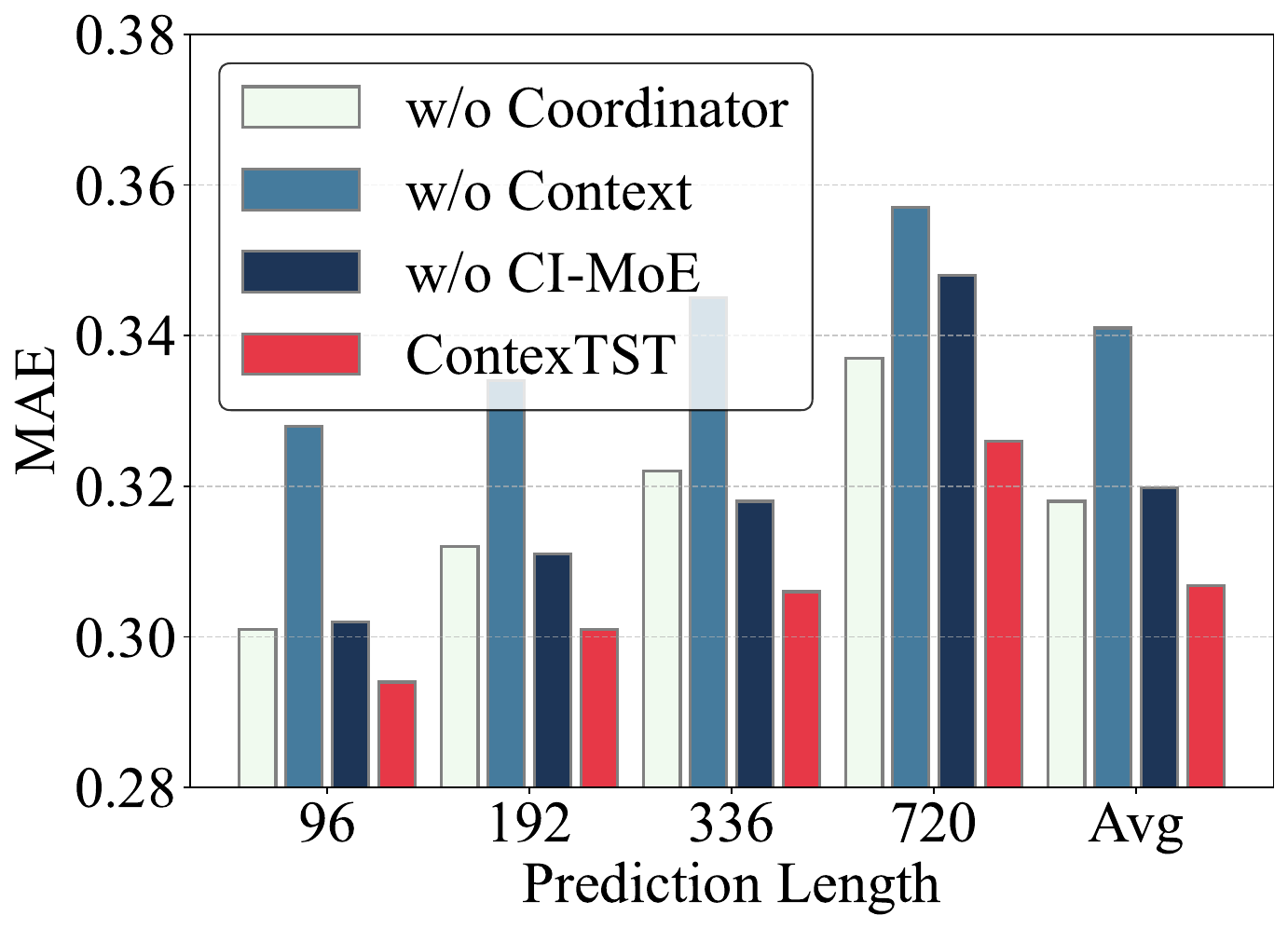}\label{fig:ab_traffic_mae}}
  
    \caption{The ablation studies about our proposed coordinator, context, and CI-MoE modules for in-domain forecasting scenario.}

    \label{fig:ablation_studies}
\end{figure*} 

\begin{figure*}[th]
    \centering
    
    \subfigure[ETTh1$\rightarrow$ETTh2]{\includegraphics[width=0.24\textwidth]{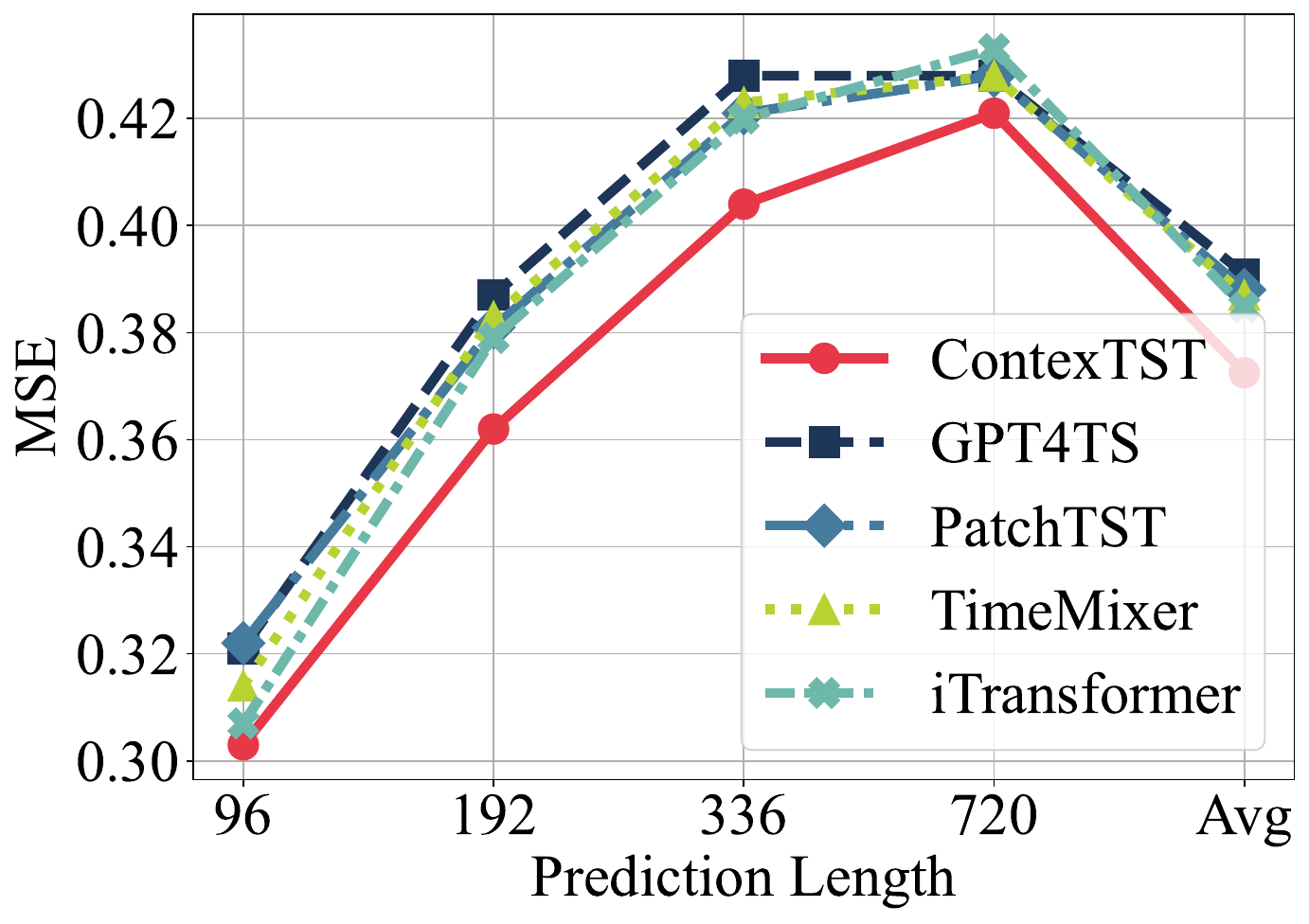}\label{fig:etth1_etth2_mse}}
    \hfill
    \subfigure[ETTh1$\rightarrow$ETTm2]{\includegraphics[width=0.24\textwidth]{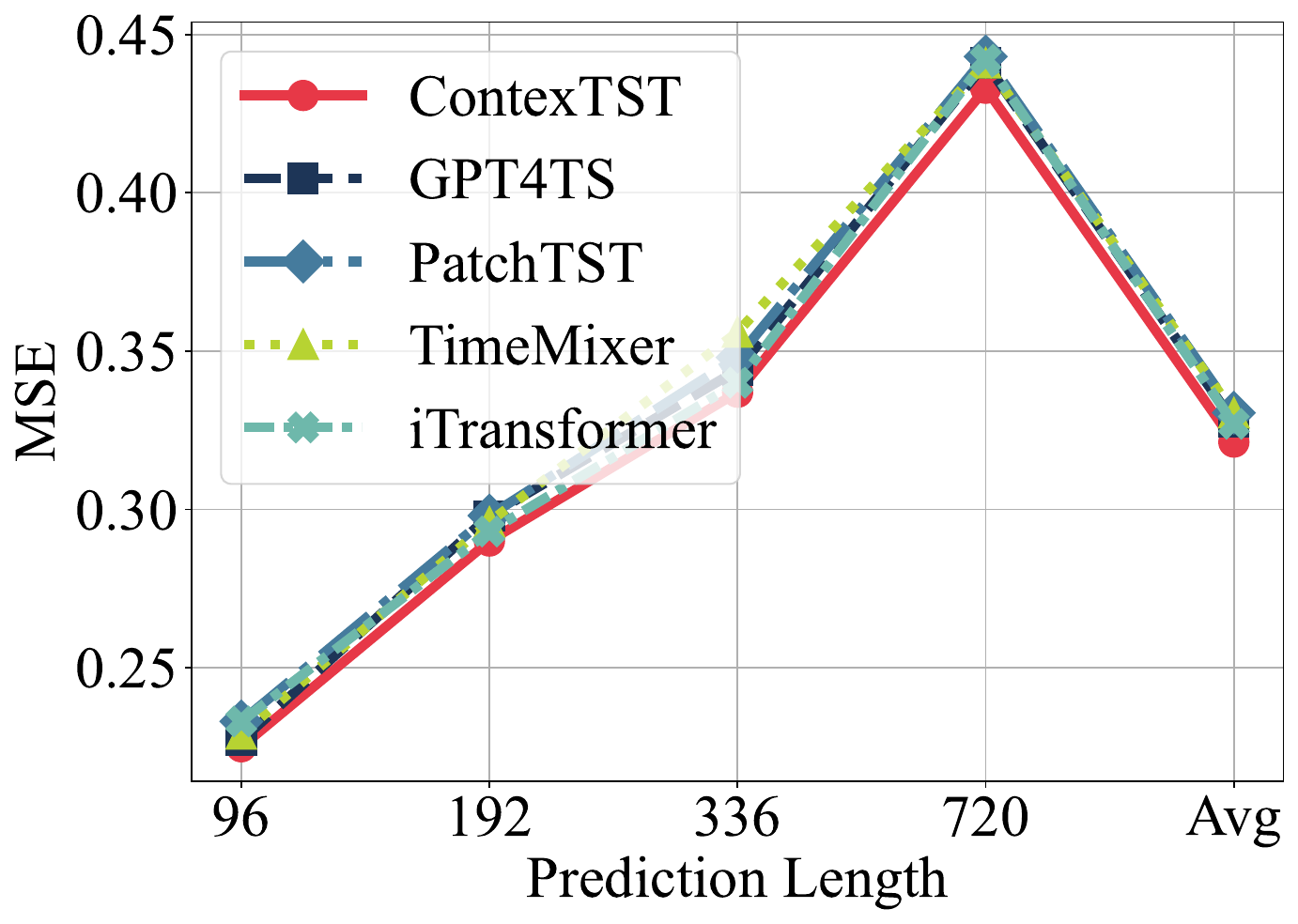}\label{fig:etth1_ettm2_mse}}
    \hfill
    \subfigure[Electricity$\rightarrow$ETTh2]{\includegraphics[width=0.24\textwidth]{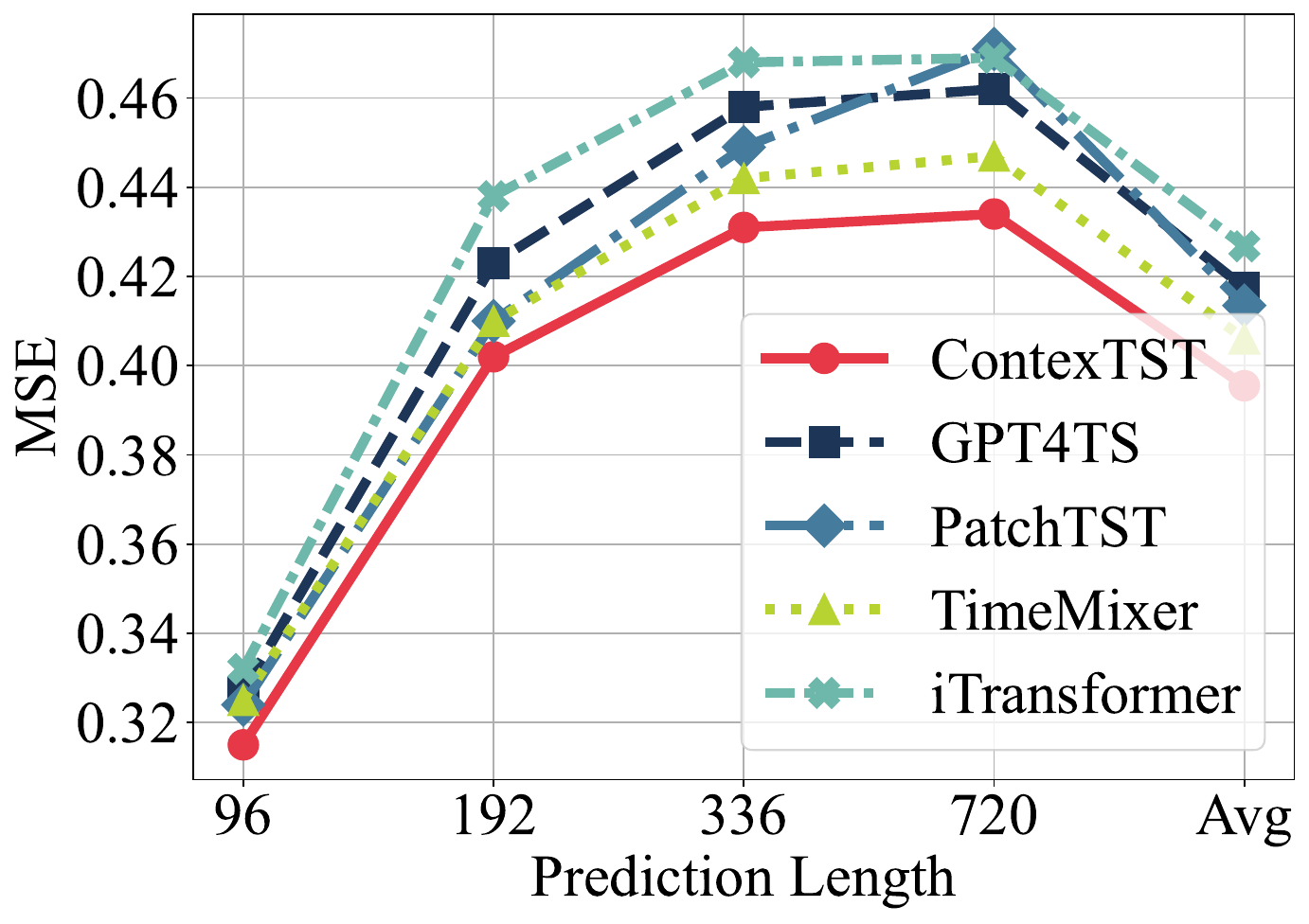}\label{fig:electricity_etth2_mse}}
    \hfil
    \subfigure[Electricity$\rightarrow$ETTm2]{\includegraphics[width=0.24\textwidth]{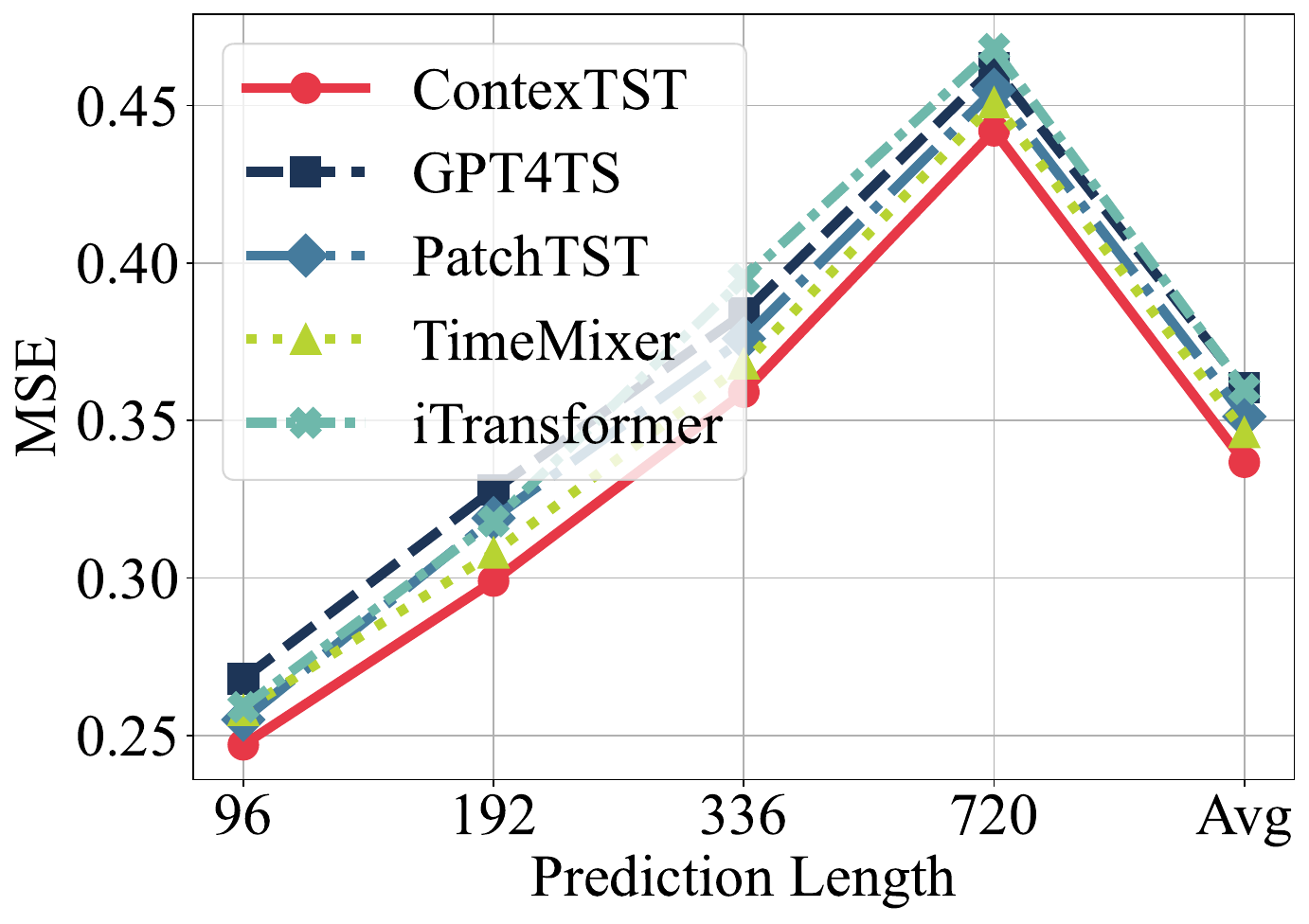}\label{fig:electricity_ettm2_mse}}
    \caption{Additional one-to-one zero-shot time series forecasting. The experimental setting is illustrated in Section~\ref{sec:exp_zero}.}

    \label{fig:ap_one-to-one_transfer}
\end{figure*}

\subsection{Full Results of In-Domain Forecasting}
We report the full results of in-domain forecasting from ContexTST and baseline methods in Tabel~\ref{tab:full_forecasting_results}. The evaluated metrics are MSE and MAE, and the results of compared approaches are either copied from TimeMixer~\cite{wang2024timemixer} and TimeXer~\cite{wang2024timexer} or reproduced by Time-Series-Library~\cite{wang2024tssurvey}.

The experimental results in Table~\ref{tab:full_forecasting_results} demonstrate that the proposed model, ContexTST, consistently outperforms state-of-the-art models, including TimeMixer, iTransformer, and PatchTST, across various datasets and prediction lengths. Specifically, ContexTST achieves the lowest average MSE and MAE across all datasets. its robustness and accuracy. Notably, its performance is particularly strong on the Electricity dataset, where it consistently achieves the best results for all prediction horizons (96, 192, 336, 720). On the Weather dataset, while models like TimeMixer occasionally show competitive results for shorter horizons, ContexTST demonstrates superior stability and accuracy, especially for longer predictions (e.g., 336 and 720). Although PatchTST performs well on the Traffic dataset for short-term predictions, ContextX/ST remains highly competitive and excels in longer-term forecasting.

Moreover, ContexTST leads in the 1st Count metric with 19 first-place rankings, significantly surpassing other models, such as TimeMixer (6) and PatchTST (5). This dominance highlights its ability to effectively capture temporal dependencies and adapt to different datasets and forecasting horizons. Overall, the results validate ContextTST as a robust and reliable solution for in-domain long-term forecasting tasks, outperforming existing methods in both accuracy and consistency.

\subsection{Ablation Studies in In-Domain Setting}
\label{sec:app_ab1}

Figure~\ref{fig:ablation_studies} demonstrates the critical role of each component in ContexTST: (1) Excluding the decomposition module significantly increases model's performance across all datasets and prediction lengths. on the ETTm1 dataset (Figure~\ref{fig:ab_ettm1_mse}), the average MSE decreases from 0.436 to 0.380, highlighting the effectiveness of the time series coordinator design in capturing underlying temporal dependency and improving forecast. (2) The absence of context information leads to degraded model performance, as shown across all datasets. For instance, in the Weather dataset (Figure~\ref{fig:ab_weather_mae}), the average MAE increases from 0.307 to 0.365, demonstrating that incorporating context-aware representations enhances the forecasting ability of ContexTST. (3) Removing the MoE module results in consistently worse performance, particularly on datasets with diverse patterns, such as Traffic (Figure~\ref{fig:ab_traffic_mse}). These results show that the decomposition, context information, and MoE modules each play a crucial role in enhancing ContexTST's forecasting capability. The full ContexTST model achieves the best performance across all datasets and prediction horizons, underscoring the synergy of these components in capturing complex temporal patterns and generalizing across diverse forecasting tasks.

\subsection{Full Results of Few-shot Forecasting}
\label{sec:full_few_shot_indomain}
To assess the models' performance in predicting future series when trained with only a few samples, we conduct the in-domain few-shot time series forecasting experiments. For the experimental setup, we use only a portion (5\% or 10\%) of the training instances for model training and evaluate the models' MSE and MAE on the full test set. The results across 7 datasets of 5\% setting are reported in Table~\ref{tab:full_few_shot_5}, and 10\% are reported in Table~\ref{tab:full_few_shot_10}.

The results indicate that ContexTST consistently outperforms other models in both the 5\% and 10\% few-shot settings across most datasets and prediction lengths. In the 5\% setting, ContexTST achieves the best average MSE and MAE across all datasets. Specifically, it demonstrates superior performance on challenging datasets like ETTm1, ETTm2, and Electricity, where it achieves the lowest errors across nearly all prediction lengths. Similarly, in the 10\% setting, ContexTST maintains its dominance, particularly excelling in datasets such as Weather and Traffic, where it achieves the best results in both short- and long-term forecasting horizons. The 1st Count metric further emphasizes this trend, with ContexTST securing 31 first-place rankings in the 5\% setting and 28 first-place rankings in the 10\% setting, far surpassing other models like TimeMixer and PatchTST.

These findings highlight the robustness and data efficiency of ContexTST, as it remains effective even with limited training data. Its consistent superiority in both accuracy metrics (MSE and MAE) across diverse datasets and prediction horizons underscores its capability to generalize well and adapt to various forecasting tasks, making it a reliable choice for few-shot time series forecasting scenarios. Furthermore, its ability to achieve competitive performance with minimal data demonstrates its potential for practical applications where labeled data is scarce or costly to obtain.

\begin{table*}[htbp]
\caption{The forecasting results from 4 different prediction lengths \{96, 192, 336, 720\} in few-shot (Label Rate = 5\%) setting. The input length is set to 96.}
\label{tab:full_few_shot_5}
\centering
% \resizebox{\linewidth}{!}{
\centering
\begin{threeparttable}
\renewcommand{\multirowsetup}{\centering}
\begin{tabular}{c|c|cc|cc|cc|cc|cc|cc}
\toprule

\multicolumn{2}{c}{\multirow{2}{*}{Models}} &
    \multicolumn{2}{c}{\rotatebox{0}{\textbf{ContexTST}}} &
    \multicolumn{2}{c}{\rotatebox{0}{TimeXer}} &
    \multicolumn{2}{c}{\rotatebox{0}{TimeMixer}} &
    \multicolumn{2}{c}{\rotatebox{0}{iTransformer}} &
    \multicolumn{2}{c}{\rotatebox{0}{PatchTST}} &
    \multicolumn{2}{c}{\rotatebox{0}{TimesNet}} \\

     \multicolumn{2}{c}{} & \multicolumn{2}{c}{(\textbf{Ours})} &
    \multicolumn{2}{c}{\citeyear{wang2024timexer}} &
    \multicolumn{2}{c}{\citeyear{wang2024timemixer}}&
    \multicolumn{2}{c}{\citeyear{liu2023itransformer}}&
    \multicolumn{2}{c}{\citeyear{patchtst}}&
    \multicolumn{2}{c}{\citeyear{wu2022timesnet}} \\

    \cmidrule(lr){3-4} \cmidrule(lr){5-6}\cmidrule(lr){7-8} \cmidrule(lr){9-10}\cmidrule(lr){11-12}\cmidrule(lr){13-14}
    \multicolumn{2}{c}{Metric} & MSE & MAE & MSE & MAE & MSE & MAE & MSE & MAE & MSE & MAE & MSE & MAE\\
\midrule

%ETTh1-96:
\multirow{5}{*}{\rotatebox{90}{ETTh1}}& 96&\boldres{0.679}&\boldres{0.548}&0.924&0.638&0.997&0.663&\secondres{0.886}&\secondres{0.628}&0.916&0.639&0.909&0.635\\
%ETTh1-192:
& 192&\boldres{0.708}&\boldres{0.566}&0.932&0.649&1.305&0.789&\secondres{0.921}&\secondres{0.646}&0.942&0.652&0.945&0.667\\
%ETTh1-336:
& 336&\boldres{0.720}&\boldres{0.579}&0.931&0.657&1.557&0.875&\secondres{0.924}&0.654&0.927&\secondres{0.653}&0.957&0.659\\
%ETTh1-720:
& 720&\boldres{0.713}&\boldres{0.596}&0.943&0.675&1.861&0.940&0.926&0.668&0.926&0.668&\secondres{0.893}&\secondres{0.650}\\
%ETTh1-Avg:
\cmidrule(lr){2-14}& Avg&\boldres{0.705}&\boldres{0.572}&0.932&0.655&1.430&0.817&\secondres{0.914}&\secondres{0.649}&0.928&0.653&0.926&0.653\\
\midrule

%ETTh2-96:
\multirow{5}{*}{\rotatebox{90}{ETTh2}}& 96&\boldres{0.348}&\boldres{0.384}&0.417&0.427&0.449&0.442&0.412&\secondres{0.422}&\secondres{0.411}&0.424&0.432&0.433\\
%ETTh2-192:
& 192&\boldres{0.429}&\boldres{0.430}&0.489&0.465&0.637&0.537&0.486&\secondres{0.463}&0.511&0.486&\secondres{0.482}&0.464\\
%ETTh2-336:
& 336&\boldres{0.455}&\boldres{0.453}&0.512&\secondres{0.486}&0.866&0.630&0.511&0.486&\secondres{0.504}&0.489&0.515&0.487\\
%ETTh2-720:
& 720&\boldres{0.451}&\boldres{0.460}&0.518&0.491&0.691&0.575&0.523&0.491&0.511&\secondres{0.485}&\secondres{0.510}&0.492\\
%ETTh2-Avg:
\cmidrule(lr){2-14}& Avg&\boldres{0.421}&\boldres{0.432}&0.484&0.467&0.661&0.546&\secondres{0.483}&\secondres{0.466}&0.484&0.471&0.485&0.469\\
\midrule

%ETTm1-96:
\multirow{5}{*}{\rotatebox{90}{ETTm1}}& 96&\secondres{0.691}&\boldres{0.547}&0.712&\secondres{0.550}&1.029&0.647&0.880&0.607&\boldres{0.638}&0.577&0.873&0.606\\
%ETTm1-192:
& 192&\secondres{0.710}&\boldres{0.557}&0.742&\secondres{0.563}&1.391&0.772&0.871&0.612&\boldres{0.685}&0.602&0.912&0.631\\
%ETTm1-336:
& 336&\secondres{0.724}&\boldres{0.567}&0.762&\secondres{0.574}&1.530&1.020&0.867&0.622&\boldres{0.689}&0.607&0.940&0.632\\
%ETTm1-720:
& 720&\boldres{0.748}&\boldres{0.581}&0.810&\secondres{0.601}&1.677&1.106&0.912&0.635&\secondres{0.756}&0.641&0.918&0.633\\
%ETTm1-Avg:
\cmidrule(lr){2-14}& Avg&\secondres{0.718}&\boldres{0.563}&0.757&\secondres{0.572}&1.407&0.886&0.882&0.619&\boldres{0.692}&0.607&0.911&0.626\\
\midrule

%ETTm2-96:
\multirow{5}{*}{\rotatebox{90}{ETTm2}}& 96&\boldres{0.229}&\secondres{0.307}&0.242&0.316&0.293&0.352&0.264&0.335&\secondres{0.231}&\boldres{0.301}&0.275&0.340\\
%ETTm2-192:
& 192&\boldres{0.284}&\secondres{0.337}&0.299&0.349&0.426&0.429&0.317&0.362&\secondres{0.289}&\boldres{0.335}&0.316&0.363\\
%ETTm2-336:
& 336&\boldres{0.338}&\boldres{0.368}&0.362&0.386&0.681&0.551&0.373&0.394&\secondres{0.344}&\secondres{0.371}&0.380&0.401\\
%ETTm2-720:
& 720&\boldres{0.433}&\boldres{0.419}&0.454&0.433&0.627&0.528&0.468&0.442&\secondres{0.436}&\secondres{0.420}&0.477&0.449\\
%ETTm2-Avg:
\cmidrule(lr){2-14}& Avg&\boldres{0.321}&\secondres{0.358}&0.339&0.371&0.507&0.465&0.355&0.383&\secondres{0.325}&\boldres{0.357}&0.362&0.388\\
\midrule

%Electricity-96:
\multirow{5}{*}{\rotatebox{90}{Electricity}}& 96&\boldres{0.278}&\boldres{0.357}&0.288&0.372&\secondres{0.279}&\secondres{0.360}&0.280&0.361&0.280&0.361&0.305&0.412\\
%Electricity-192:
& 192&\boldres{0.283}&\boldres{0.371}&0.296&0.399&0.297&0.382&\secondres{0.292}&\secondres{0.373}&0.292&0.375&0.385&0.453\\
%Electricity-336:
& 336&\boldres{0.301}&\boldres{0.379}&0.335&0.413&0.338&0.402&0.317&0.394&\secondres{0.303}&\secondres{0.383}&0.392&0.473\\
%Electricity-720:
& 720&\boldres{0.348}&\boldres{0.397}&0.379&0.418&0.382&0.416&0.374&0.435&\secondres{0.351}&\secondres{0.415}&0.407&0.489\\
%Electricity-Avg:
\cmidrule(lr){2-14}& Avg&\boldres{0.302}&\boldres{0.376}&0.325&0.400&0.324&0.390&0.316&0.391&\secondres{0.306}&\secondres{0.384}&0.372&0.457\\
\midrule

%Weather-96:
\multirow{5}{*}{\rotatebox{90}{Weather}}& 96&\boldres{0.215}&\boldres{0.263}&0.232&0.277&\secondres{0.227}&0.284&0.240&0.293&0.231&\secondres{0.275}&0.297&0.307\\
%Weather-192:
& 192&\boldres{0.259}&\boldres{0.298}&0.280&0.320&\secondres{0.268}&\secondres{0.299}&0.302&0.325&0.282&0.311&0.321&0.342\\
%Weather-336:
& 336&\boldres{0.310}&\boldres{0.332}&0.330&0.341&\secondres{0.322}&0.339&0.331&0.346&0.322&\secondres{0.335}&0.365&0.378\\
%Weather-720:
& 720&\boldres{0.380}&\secondres{0.375}&0.396&0.382&\secondres{0.385}&0.379&0.402&0.396&0.389&\boldres{0.374}&0.401&0.413\\
%Weather-Avg:
\cmidrule(lr){2-14}& Avg&\boldres{0.291}&\boldres{0.317}&0.309&0.330&\secondres{0.300}&0.325&0.319&0.340&0.306&\secondres{0.324}&0.346&0.360\\
\midrule

%Traffic-96:
\multirow{5}{*}{\rotatebox{90}{Traffic}}& 96&\boldres{0.754}&\boldres{0.517}&0.845&0.539&\secondres{0.835}&0.623&0.862&\secondres{0.525}&0.907&0.959&1.045&0.923\\
%Traffic-192:
& 192&\boldres{0.786}&\secondres{0.539}&0.877&0.547&0.903&0.668&\secondres{0.818}&\boldres{0.517}&1.069&0.987&1.105&0.962\\
%Traffic-336:
& 336&\boldres{0.826}&\secondres{0.547}&\secondres{0.873}&0.549&0.914&0.692&0.875&\boldres{0.542}&1.190&1.052&1.139&0.970\\
%Traffic-720:
& 720&\boldres{0.885}&\boldres{0.552}&\secondres{0.912}&\secondres{0.558}&0.923&0.685&0.933&0.558&1.285&1.153&1.202&0.963\\
%Traffic-Avg:
\cmidrule(lr){2-14}& Avg&\boldres{0.813}&\secondres{0.539}&0.877&0.548&0.894&0.667&\secondres{0.872}&\boldres{0.536}&1.113&1.038&1.123&0.955\\
\midrule

\multicolumn{2}{c|}{\rotatebox{0}{\scalebox{0.9}{$1^{\text{st}}$ Count}}} & \boldres{31} & \boldres{30} & 0 & 0 & 0 & 0 & 0 & 3 & \secondres{4} & \secondres{4} & 0 & 0\\

\bottomrule

\end{tabular}
\end{threeparttable}
% }
\end{table*}

% \begin{tabular}{c|c|cc|cc|cc|cc|cc|cc}
% \toprule

% \multicolumn{2}{c}{\multirow{2}{*}{Models}} &
%     \multicolumn{2}{c}{\rotatebox{0}{\scalebox{0.8}{\textbf{ContexTST}}}} &
%     \multicolumn{2}{c}{\rotatebox{0}{\scalebox{0.8}{TimeXer}}} &
%     \multicolumn{2}{c}{\rotatebox{0}{\scalebox{0.8}{TimeMixer}}} &
%     \multicolumn{2}{c}{\rotatebox{0}{\scalebox{0.8}{iTransformer}}} &
%     \multicolumn{2}{c}{\rotatebox{0}{\scalebox{0.8}{PatchTST}}} &
%     \multicolumn{2}{c}{\rotatebox{0}{\scalebox{0.8}{TimesNet}}} \\

%      \multicolumn{2}{c}{} & \multicolumn{2}{c}{\scalebox{0.8}{(\textbf{Ours})}} &
%     \multicolumn{2}{c}{\scalebox{0.8}{\citeyear{wang2024timexer}}} &
%     \multicolumn{2}{c}{\scalebox{0.8}{\citeyear{wang2024timemixer}}}&
%     \multicolumn{2}{c}{\scalebox{0.8}{\citeyear{liu2023itransformer}}}&
%     \multicolumn{2}{c}{\scalebox{0.8}{\citeyear{patchtst}}}&
%     \multicolumn{2}{c}{\scalebox{0.8}{\citeyear{wu2022timesnet}}} \\

%     \cmidrule(lr){3-4} \cmidrule(lr){5-6}\cmidrule(lr){7-8} \cmidrule(lr){9-10}\cmidrule(lr){11-12}\cmidrule(lr){13-14}
%     \multicolumn{2}{c}{\scalebox{0.78}{Metric}} & \scalebox{0.78}{MSE} & \scalebox{0.78}{MAE} & \scalebox{0.78}{MSE} & \scalebox{0.78}{MAE} & \scalebox{0.78}{MSE} & \scalebox{0.78}{MAE} & \scalebox{0.78}{MSE} & \scalebox{0.78}{MAE} & \scalebox{0.78}{MSE} & \scalebox{0.78}{MAE} & \scalebox{0.78}{MSE} & \scalebox{0.78}{MAE}\\
% \midrule

% \end{tabular}

\begin{table*}[htbp]
\caption{The forecasting results from 4 different prediction lengths \{96, 192, 336, 720\} in few-shot (Label Rate = 10\%) setting. The input length is set to 96.}
\label{tab:full_few_shot_10}
\centering
% \resizebox{\linewidth}{!}{
\centering
\begin{threeparttable}
\renewcommand{\multirowsetup}{\centering}
\begin{tabular}{c|c|cc|cc|cc|cc|cc|cc}
\toprule

\multicolumn{2}{c}{\multirow{2}{*}{Models}} &
    \multicolumn{2}{c}{\rotatebox{0}{\textbf{ContexTST}}} &
    \multicolumn{2}{c}{\rotatebox{0}{TimeXer}} &
    \multicolumn{2}{c}{\rotatebox{0}{TimeMixer}} &
    \multicolumn{2}{c}{\rotatebox{0}{iTransformer}} &
    \multicolumn{2}{c}{\rotatebox{0}{PatchTST}} &
    \multicolumn{2}{c}{\rotatebox{0}{TimesNet}} \\

     \multicolumn{2}{c}{} & \multicolumn{2}{c}{(\textbf{Ours})} &
    \multicolumn{2}{c}{\citeyear{wang2024timexer}} &
    \multicolumn{2}{c}{\citeyear{wang2024timemixer}}&
    \multicolumn{2}{c}{\citeyear{liu2023itransformer}}&
    \multicolumn{2}{c}{\citeyear{patchtst}}&
    \multicolumn{2}{c}{\citeyear{wu2022timesnet}} \\

    \cmidrule(lr){3-4} \cmidrule(lr){5-6}\cmidrule(lr){7-8} \cmidrule(lr){9-10}\cmidrule(lr){11-12}\cmidrule(lr){13-14}
    \multicolumn{2}{c}{Metric} & MSE & MAE & MSE & MAE & MSE & MAE & MSE & MAE & MSE & MAE & MSE & MAE\\
\midrule

%ETTh1-96:
\multirow{5}{*}{\rotatebox{90}{ETTh1}}& 96&\boldres{0.667}&\boldres{0.537}&0.924&0.638&0.996&0.662&\secondres{0.873}&\secondres{0.619}&0.875&0.627&0.892&0.624\\
%ETTh1-192:
& 192&\boldres{0.704}&\boldres{0.558}&0.932&0.649&1.305&0.789&0.917&\secondres{0.639}&\secondres{0.915}&0.648&0.937&0.643\\
%ETTh1-336:
& 336&\boldres{0.712}&\boldres{0.564}&0.931&0.657&1.406&0.811&\secondres{0.920}&\secondres{0.647}&0.927&0.653&0.956&0.658\\
%ETTh1-720:
& 720&\boldres{0.707}&\boldres{0.571}&0.934&0.665&1.547&0.875&\secondres{0.922}&\secondres{0.654}&0.931&0.658&0.967&0.660\\
%ETTh1-Avg:
\cmidrule(lr){2-14}& Avg&\boldres{0.698}&\boldres{0.558}&0.930&0.652&1.313&0.825&\secondres{0.908}&\secondres{0.640}&0.912&0.646&0.938&0.646\\
\midrule

%ETTh2-96:
\multirow{5}{*}{\rotatebox{90}{ETTh2}}& 96&\boldres{0.338}&\boldres{0.376}&0.408&\secondres{0.413}&0.432&0.421&0.410&0.416&\secondres{0.405}&0.416&0.425&0.427\\
%ETTh2-192:
& 192&\boldres{0.416}&\boldres{0.425}&0.478&\secondres{0.456}&0.598&0.511&\secondres{0.475}&0.458&0.491&0.465&0.476&0.459\\
%ETTh2-336:
& 336&\boldres{0.449}&\boldres{0.437}&0.511&0.485&0.786&0.573&0.507&0.487&\secondres{0.503}&0.485&0.511&\secondres{0.473}\\
%ETTh2-720:
& 720&\boldres{0.442}&\boldres{0.451}&0.508&0.489&0.766&0.562&\secondres{0.503}&0.490&0.507&0.489&0.508&\secondres{0.481}\\
%ETTh2-Avg:
\cmidrule(lr){2-14}& Avg&\boldres{0.411}&\boldres{0.422}&0.476&0.461&0.645&0.517&\secondres{0.474}&0.463&0.477&0.464&0.480&\secondres{0.460}\\
\midrule

%ETTm1-96:
\multirow{5}{*}{\rotatebox{90}{ETTm1}}& 96&\secondres{0.572}&\boldres{0.487}&0.576&\secondres{0.497}&0.996&0.587&0.811&0.584&\boldres{0.518}&0.517&0.861&0.602\\
%ETTm1-192:
& 192&\secondres{0.599}&\boldres{0.496}&0.604&\secondres{0.509}&1.239&0.734&0.828&0.593&\boldres{0.559}&0.536&0.887&0.622\\
%ETTm1-336:
& 336&0.630&\boldres{0.519}&\secondres{0.629}&\secondres{0.521}&1.433&0.971&0.843&0.605&\boldres{0.585}&0.553&0.926&0.627\\
%ETTm1-720:
& 720&\secondres{0.665}&\boldres{0.534}&0.677&\secondres{0.548}&1.520&0.988&0.875&0.621&\boldres{0.656}&0.590&0.907&0.630\\
%ETTm1-Avg:
\cmidrule(lr){2-14}& Avg&\secondres{0.616}&\boldres{0.509}&0.621&\secondres{0.519}&1.297&0.820&0.839&0.601&\boldres{0.580}&0.549&0.895&0.620\\
\midrule

%ETTm2-96:
\multirow{5}{*}{\rotatebox{90}{ETTm2}}& 96&0.227&0.305&\secondres{0.222}&\secondres{0.301}&0.278&0.344&0.273&0.339&\boldres{0.205}&\boldres{0.286}&0.273&0.339\\
%ETTm2-192:
& 192&\secondres{0.273}&\boldres{0.256}&0.282&0.336&0.417&0.408&0.311&0.358&\boldres{0.272}&\secondres{0.327}&0.314&0.362\\
%ETTm2-336:
& 336&\boldres{0.327}&\boldres{0.356}&0.347&0.376&0.642&0.519&0.369&0.391&\secondres{0.328}&\secondres{0.360}&0.378&0.399\\
%ETTm2-720:
& 720&\secondres{0.423}&\secondres{0.412}&0.444&0.426&0.584&0.495&0.464&0.440&\boldres{0.422}&\boldres{0.408}&0.475&0.448\\
%ETTm2-Avg:
\cmidrule(lr){2-14}& Avg&\secondres{0.312}&\boldres{0.332}&0.324&0.360&0.480&0.442&0.354&0.382&\boldres{0.307}&\secondres{0.345}&0.360&0.387\\
\midrule

%Electricity-96:
\multirow{5}{*}{\rotatebox{90}{Electricity}}& 96&\boldres{0.225}&\boldres{0.289}&0.268&0.352&0.254&0.338&0.244&0.330&\secondres{0.232}&\secondres{0.317}&0.298&0.407\\
%Electricity-192:
& 192&\boldres{0.231}&\boldres{0.296}&0.275&0.362&0.265&0.351&0.253&0.341&\secondres{0.237}&\secondres{0.324}&0.375&0.447\\
%Electricity-336:
& 336&\boldres{0.257}&\boldres{0.338}&0.302&0.385&0.276&0.372&0.279&0.363&\secondres{0.258}&\secondres{0.343}&0.385&0.463\\
%Electricity-720:
& 720&\boldres{0.297}&\boldres{0.356}&0.346&0.401&0.331&0.402&0.329&0.398&\secondres{0.302}&\secondres{0.375}&0.391&0.475\\
%Electricity-Avg:
\cmidrule(lr){2-14}& Avg&\boldres{0.253}&\boldres{0.320}&0.298&0.375&0.282&0.366&0.276&0.358&\secondres{0.257}&\secondres{0.340}&0.362&0.448\\
\midrule

%Weather-96:
\multirow{5}{*}{\rotatebox{90}{Weather}}& 96&\boldres{0.208}&\boldres{0.256}&0.218&0.264&\secondres{0.217}&0.263&0.233&0.287&0.219&\secondres{0.260}&0.286&0.305\\
%Weather-192:
& 192&\boldres{0.256}&\boldres{0.293}&\secondres{0.263}&\secondres{0.296}&0.269&0.304&0.285&0.316&0.274&0.301&0.315&0.334\\
%Weather-336:
& 336&\boldres{0.306}&\boldres{0.327}&0.316&0.331&0.313&0.336&\secondres{0.312}&0.331&0.322&\secondres{0.330}&0.358&0.369\\
%Weather-720:
& 720&\secondres{0.378}&0.373&0.382&\secondres{0.371}&\boldres{0.369}&0.375&0.397&0.392&0.385&\boldres{0.369}&0.392&0.406\\
%Weather-Avg:
\cmidrule(lr){2-14}& Avg&\boldres{0.287}&\boldres{0.312}&0.295&0.316&\secondres{0.292}&0.320&0.307&0.332&0.300&\secondres{0.315}&0.338&0.354\\
\midrule

%Traffic-96:
\multirow{5}{*}{\rotatebox{90}{Traffic}}& 96&\boldres{0.715}&\boldres{0.473}&0.798&0.490&\secondres{0.763}&0.608&0.768&\secondres{0.486}&1.081&0.638&0.906&0.836\\
%Traffic-192:
& 192&\secondres{0.744}&0.483&0.751&\secondres{0.472}&0.771&0.612&\boldres{0.730}&\boldres{0.471}&1.064&0.635&0.912&0.875\\
%Traffic-336:
& 336&\boldres{0.766}&\secondres{0.495}&\secondres{0.768}&\boldres{0.483}&0.805&0.634&0.786&0.500&1.078&0.637&0.927&0.883\\
%Traffic-720:
& 720&\boldres{0.791}&\secondres{0.503}&\secondres{0.792}&\boldres{0.488}&0.817&0.635&0.837&0.518&1.117&0.648&0.973&0.901\\
%Traffic-Avg:
\cmidrule(lr){2-14}& Avg&\boldres{0.754}&\secondres{0.489}&\secondres{0.777}&\boldres{0.483}&0.789&0.622&0.780&0.494&1.085&0.640&0.929&0.874\\
\midrule

\multicolumn{2}{c|}{\rotatebox{0}{\scalebox{0.9}{$1^{\text{st}}$ Count}}} & \boldres{24} & \boldres{28} & 0 & 3 & 1 & 0 & 1 & 1 & \secondres{9} & \secondres{3} & 0 & 0\\

\bottomrule

\end{tabular}
\end{threeparttable}
% }
\end{table*}

\section{Full Results in Cross-Domain Setting}
\label{app:cross_domain}
Due to the space limitation of the main text, we show the full experimental results under the cross-domain setup in this section, including (1) the results of zero-shot transfer to other datasets, (2) the comparison of the results with TSFMs under MAE metrics, and (3) the ablation experiments under cross-domain transfer.

\subsection{Additional Results of Cross-Domain Transfer}
\label{sec:ap_cross_domain}

Figure~\ref{fig:ap_one-to-one_transfer} presents the results of one-to-one zero-shot time series forecasting experiments, evaluating models on unseen target datasets across four scenarios: ETTh1 $\rightarrow$ ETTh2, ETTh1 $\rightarrow$ ETTm2, Electricity $\rightarrow$ ETTh2, and Electricity $\rightarrow$ ETTm2. The performance is measured at different prediction lengths (96, 192, 336, 720) and as an overall average. The results demonstrate that the proposed ContextTST consistently achieves competitive or superior performance compared to other models, particularly excelling in shorter prediction lengths (96 and 192). Its ability to generalize effectively highlights its robustness in transferring knowledge across datasets. 

For similar dataset transfers (ETTh1 $\rightarrow$ ETTh2 and ETTh1 $\rightarrow$ ETTm2), ContexTST achieves the best or near-best results across all prediction lengths, effectively capturing temporal dependencies within similar domains. In more challenging cross-domain transfers (Electricity $\rightarrow$ ETTh2 and Electricity $\rightarrow$ ETTm2), where differences between the source and target datasets are more significant, ContexTST maintains competitive performance, outperforming models such as GPT4TS and TimesNet and matching the best-performing models like PatchTST in certain cases. Notably, ContextTST demonstrates remarkable robustness for longer prediction horizons (336 and 720), where other models often experience a significant drop in performance.

In summary, the results validate the strong generalization ability of ContexTST in zero-shot forecasting tasks. Its consistent performance across diverse datasets and prediction horizons underscores its effectiveness and reliability in adapting to unseen time series data. These findings highlight ContexTST as a robust and versatile model for zero-shot time series forecasting.

\begin{figure*}[th]
    \centering
    % 子图
    \subfigure[ETTh1]{\includegraphics[width=0.24\textwidth]{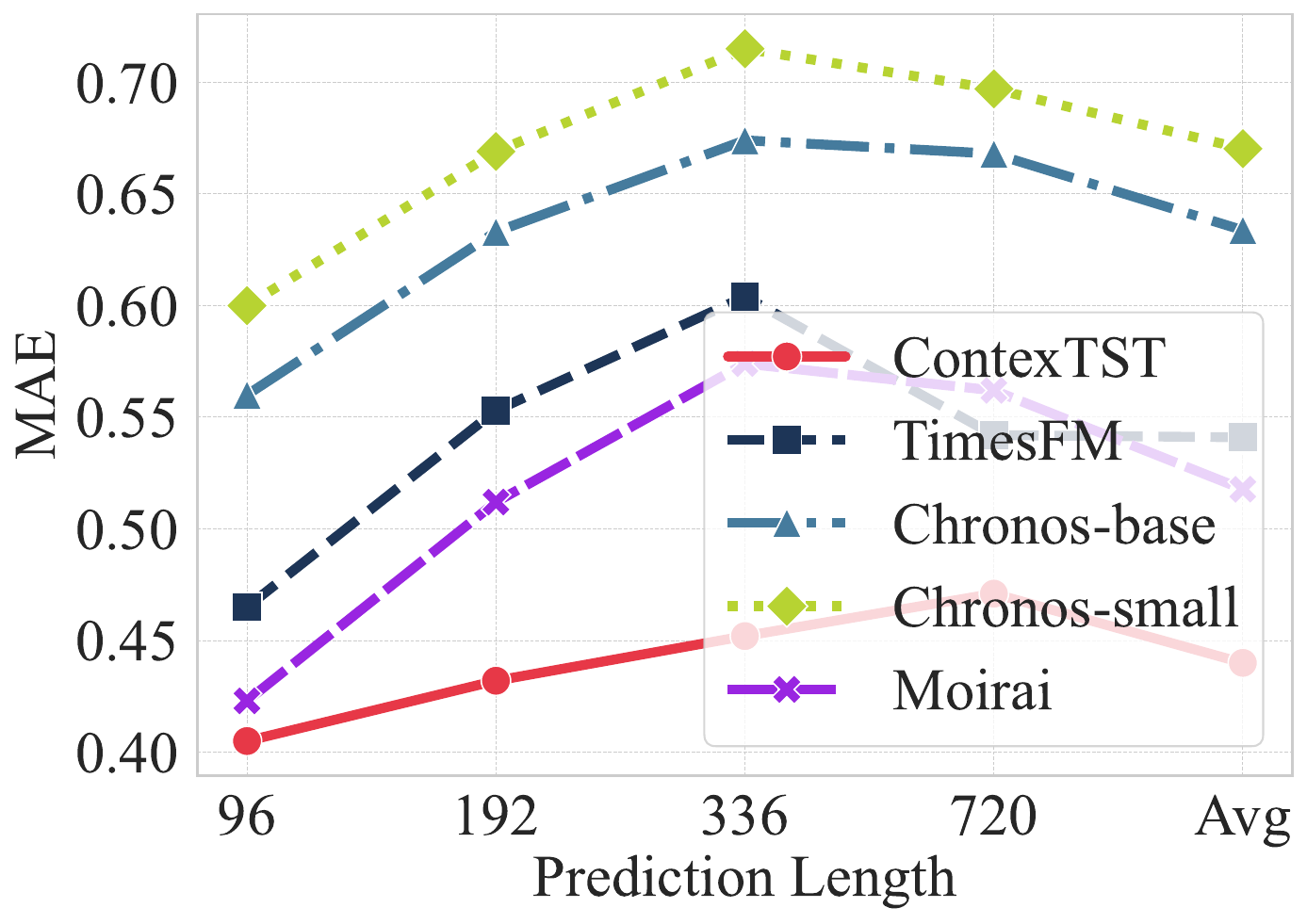}\label{fig:etth1_mae}}
    \hfill
    \subfigure[ETTh2]{\includegraphics[width=0.24\textwidth]{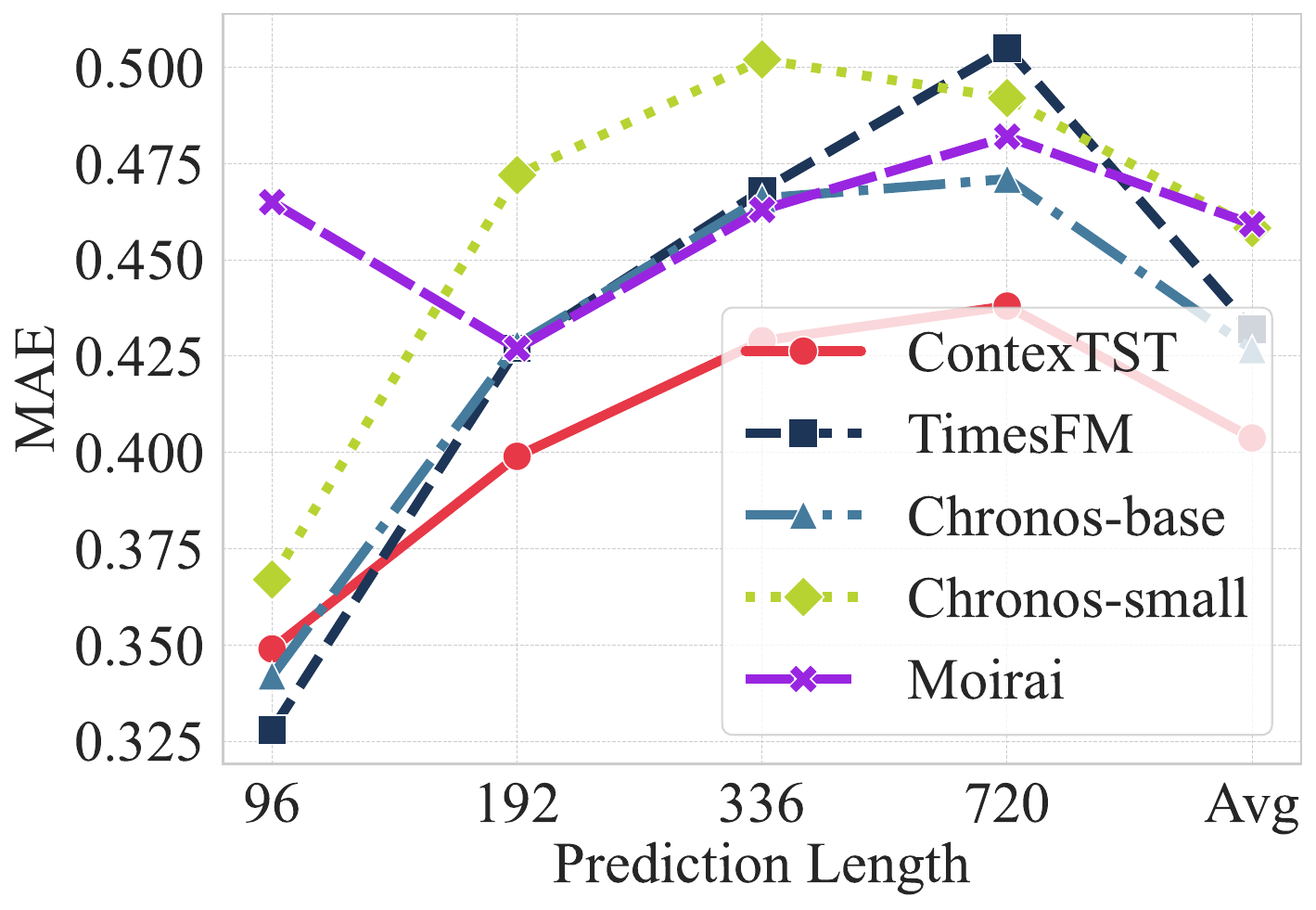}\label{fig:etth2_mae}}
    \hfill
    \subfigure[ETTm1]{\includegraphics[width=0.24\textwidth]{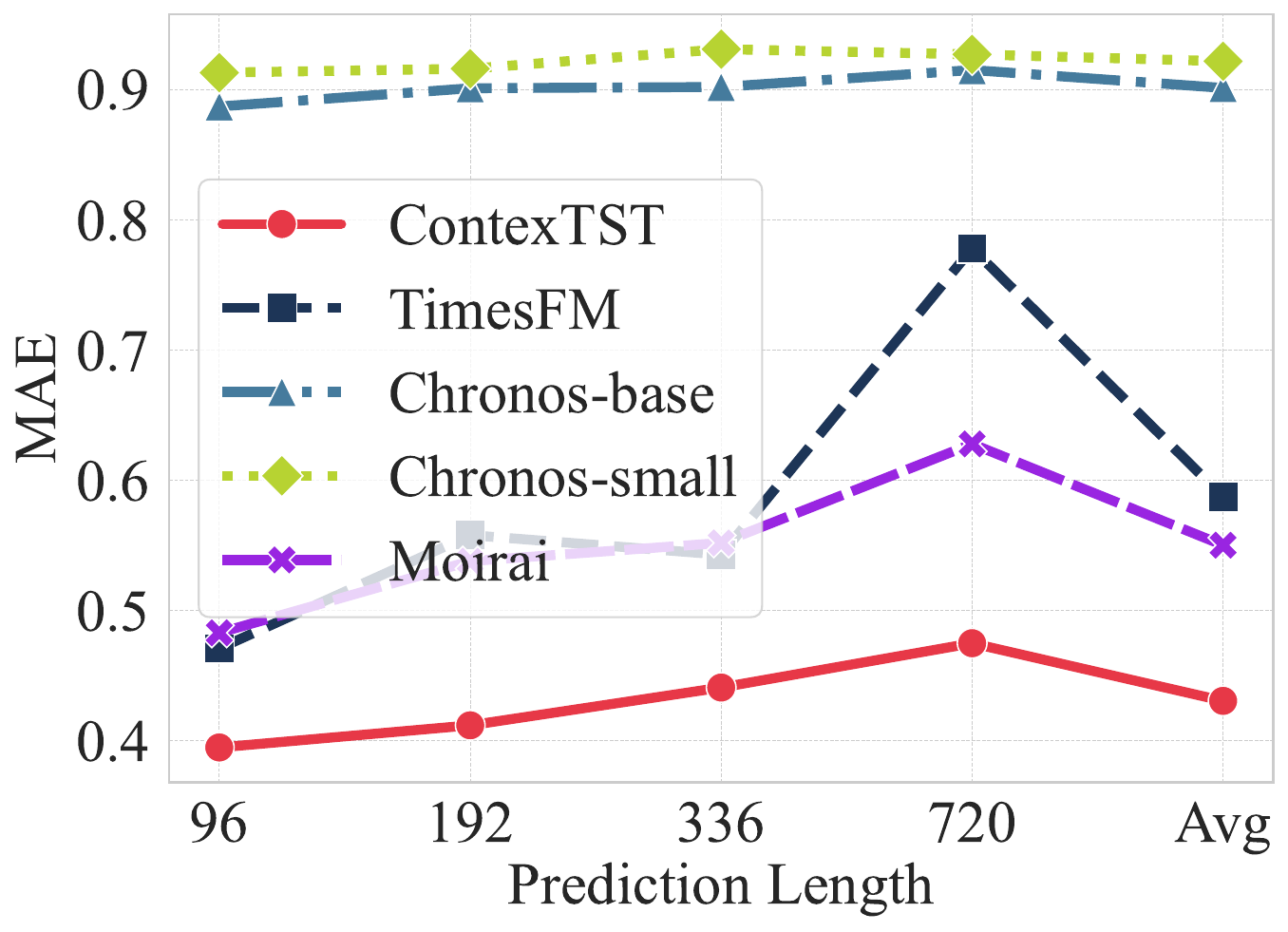}\label{fig:ettm1_mae}}
    \hfill
    \subfigure[ETTm2]{\includegraphics[width=0.24\textwidth]{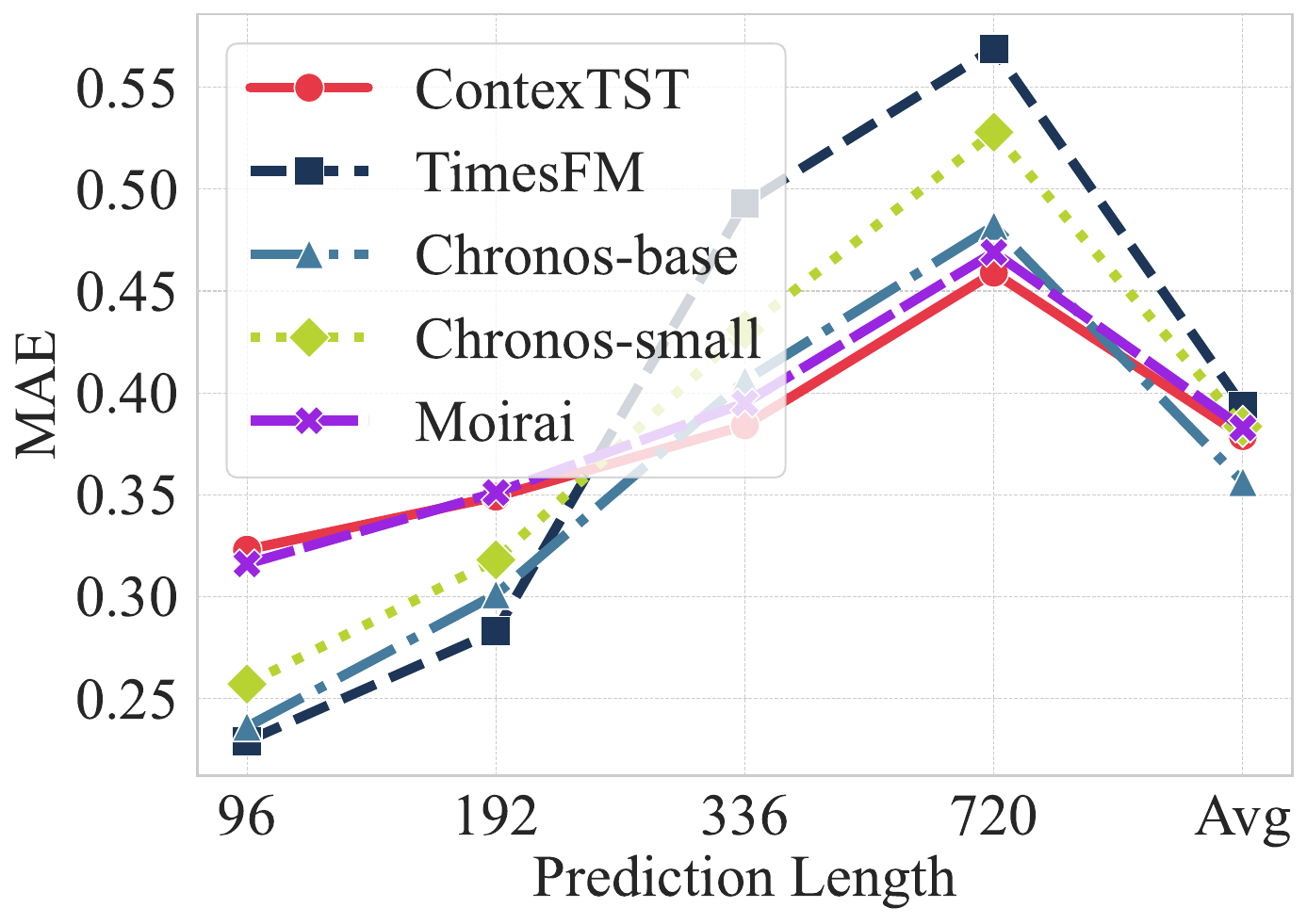}\label{fig:ettm2_mae}}\\
  
    \caption{Comparison with foundational time series models on diverse prediction horizons. The input sequence length is set to 96 for all models. For each foundation model, we exclude the evaluation results on its pre-trained datasets, and ContexTST is pre-trained on Electricity datasets then zero-shot inference in 4 target domains.}

    \label{fig:ap_foundation_res}
\end{figure*}

\begin{figure*}[th]
    \centering
    
    \subfigure[ETTh1$\rightarrow$ETTh2]{\includegraphics[width=0.24\textwidth]{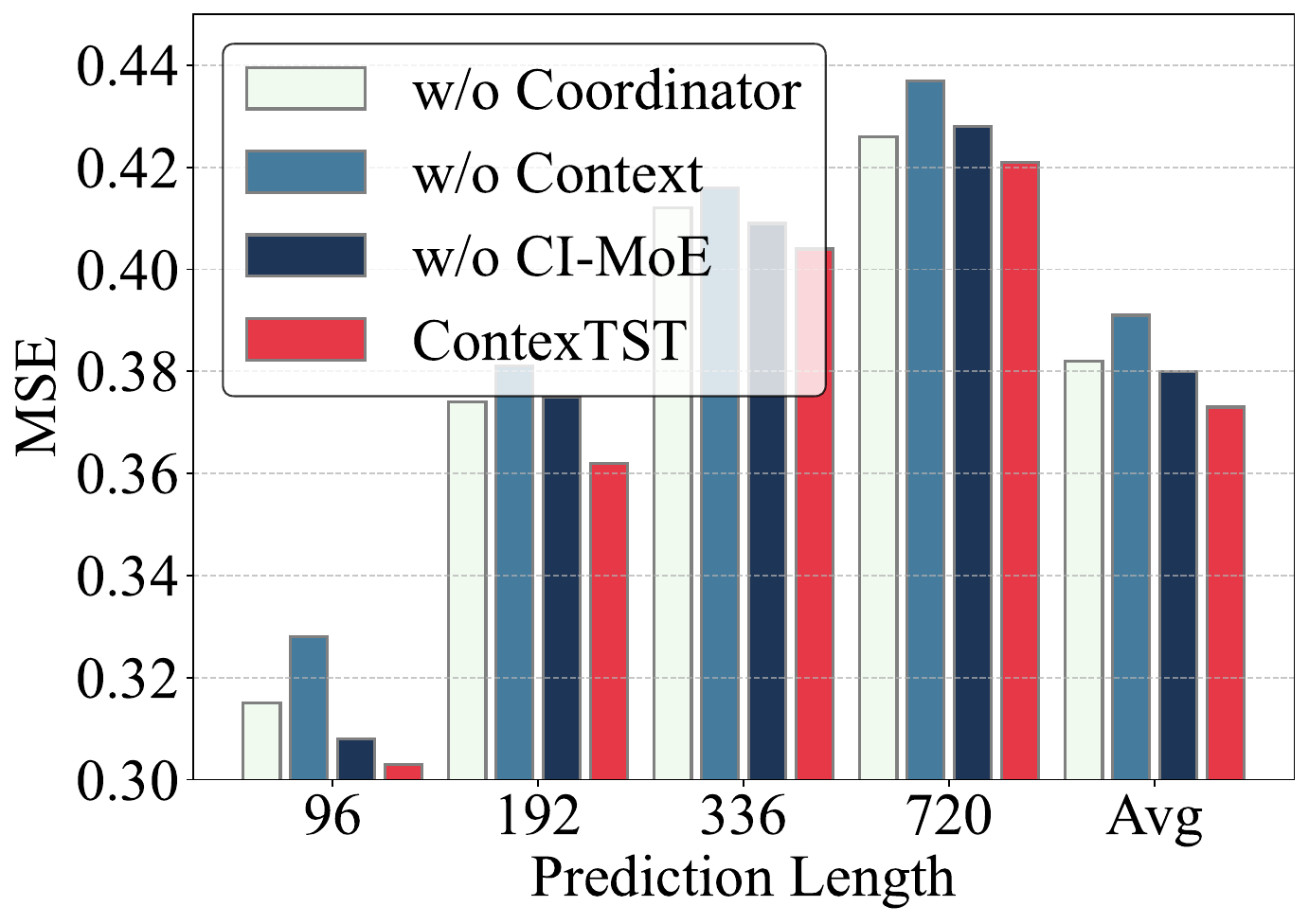}\label{fig:ab_etth1_etth2_mse}}
    \hfill
    \subfigure[ETTh1$\rightarrow$ETTm2]{\includegraphics[width=0.24\textwidth]{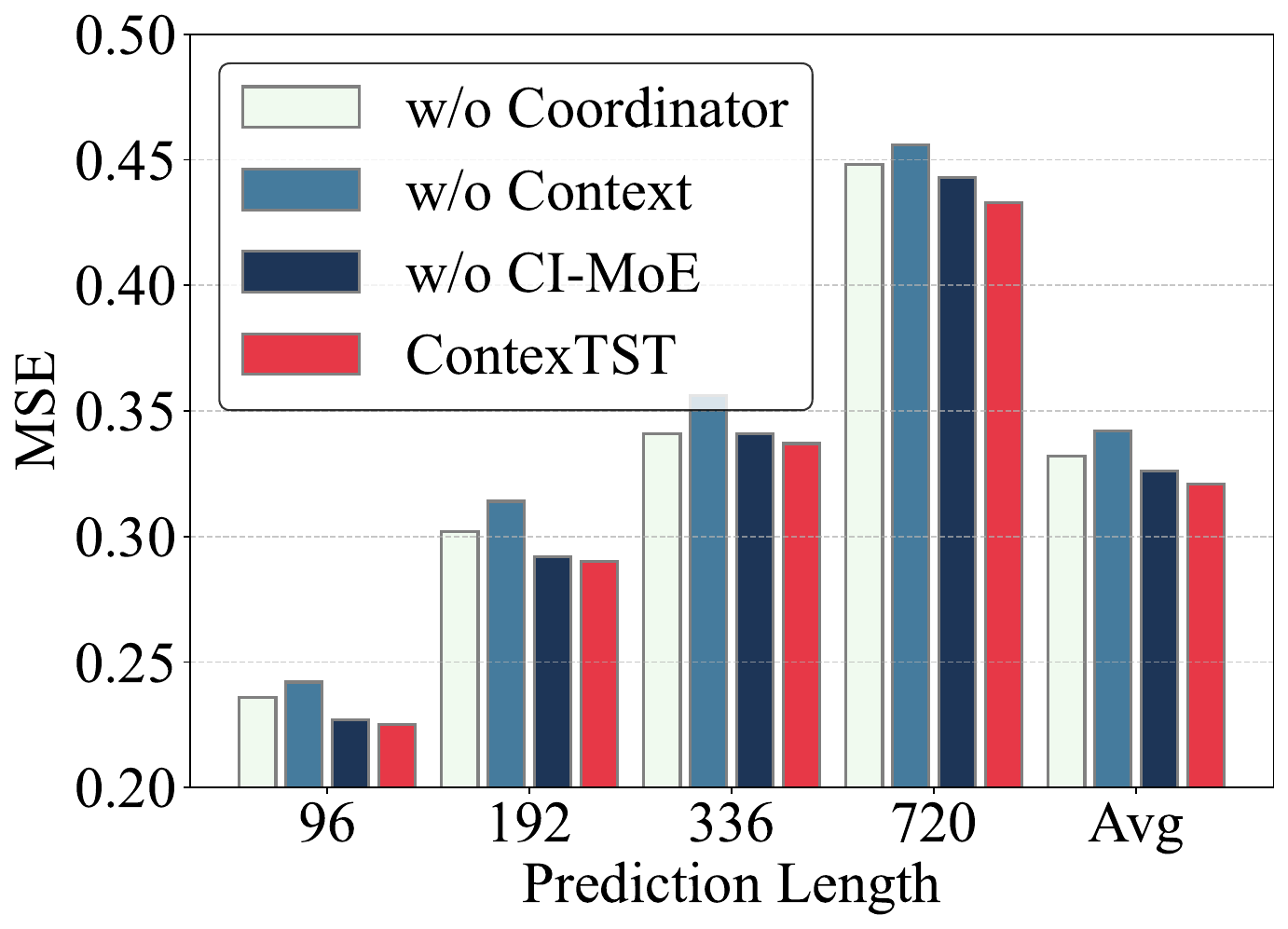}\label{fig:ab_etth1_ettm2_mse}}
    \hfill
    \subfigure[Electricity$\rightarrow$ETTh2]{\includegraphics[width=0.24\textwidth]{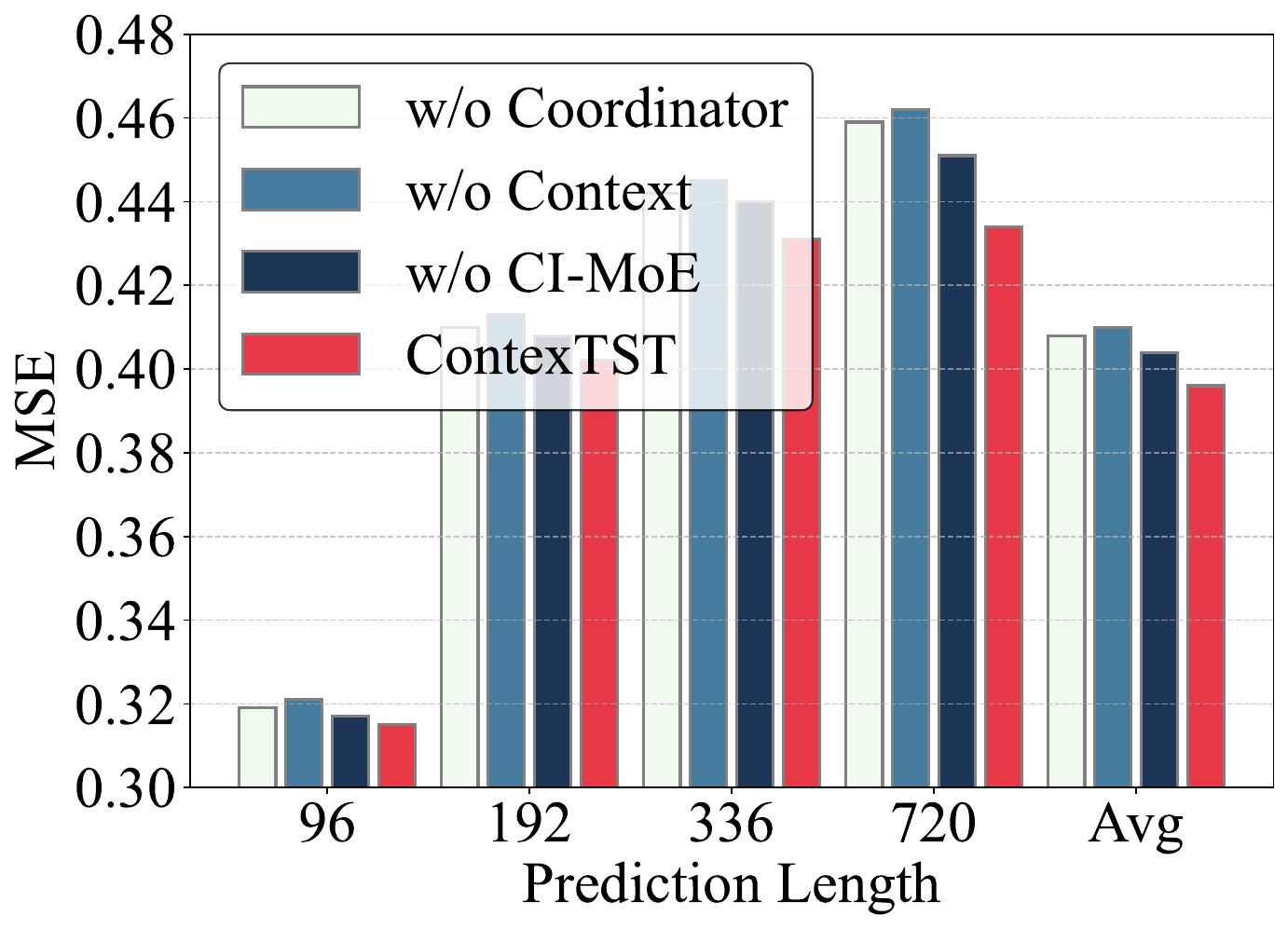}\label{fig:ab_ecl_etth2_mse}}
      \subfigure[Electricity$\rightarrow$ETTm2]{\includegraphics[width=0.24\textwidth]{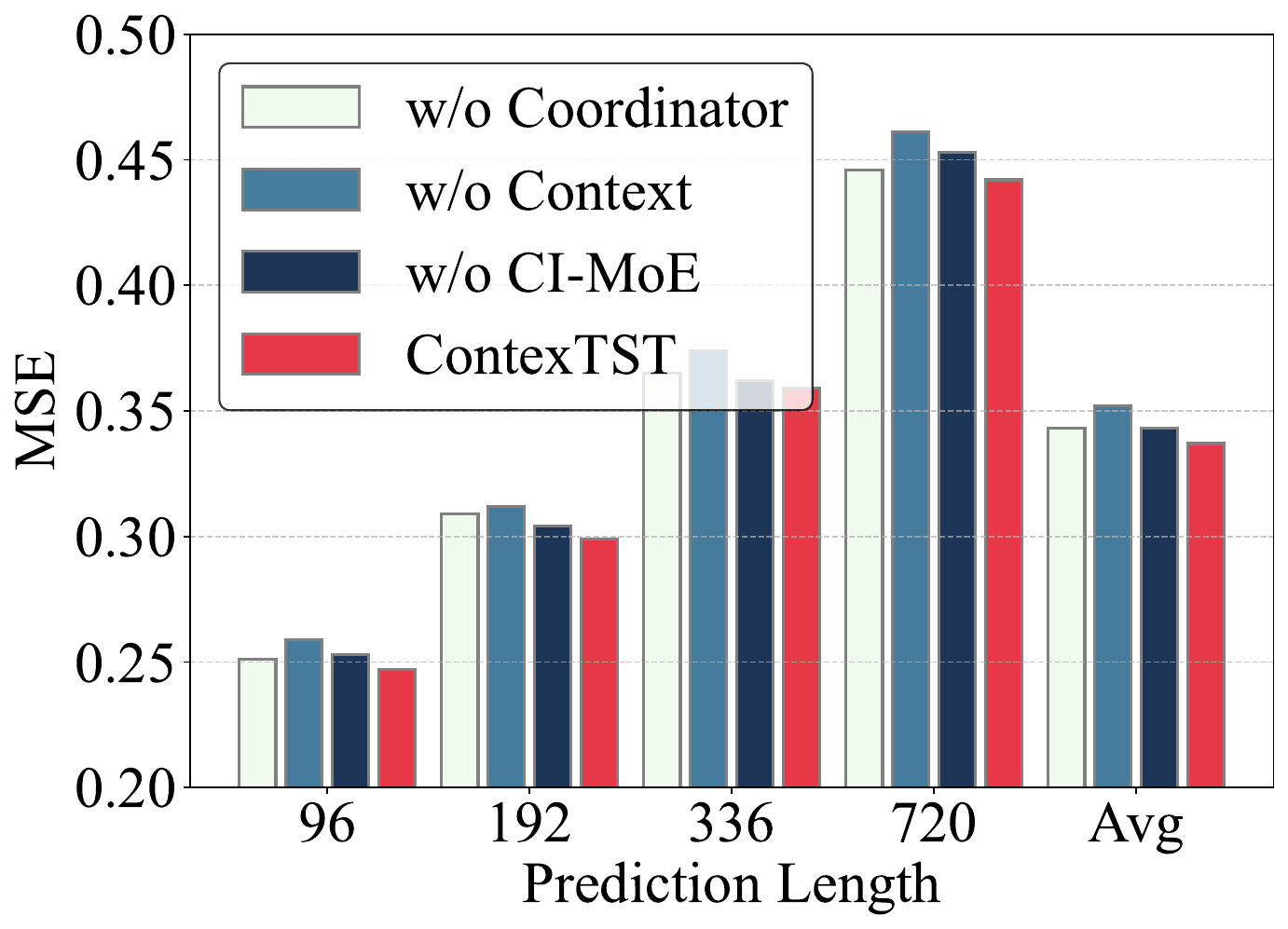}\label{fig:ab_ecl_ettm2_mse}}
  
    \caption{Additional ablation studies about our proposed time series coordinator, context, and CI-MoE modules in cross-domain transfer scenario. The experimental setting is illustrated in Section~\ref{sec:exp_ablation}}

    \label{fig:ap_cross_ablation_studies}
\end{figure*}

\begin{figure*}[th]
    \centering
    \subfigure[ETTh1]
    {\includegraphics[width=0.23\textwidth]{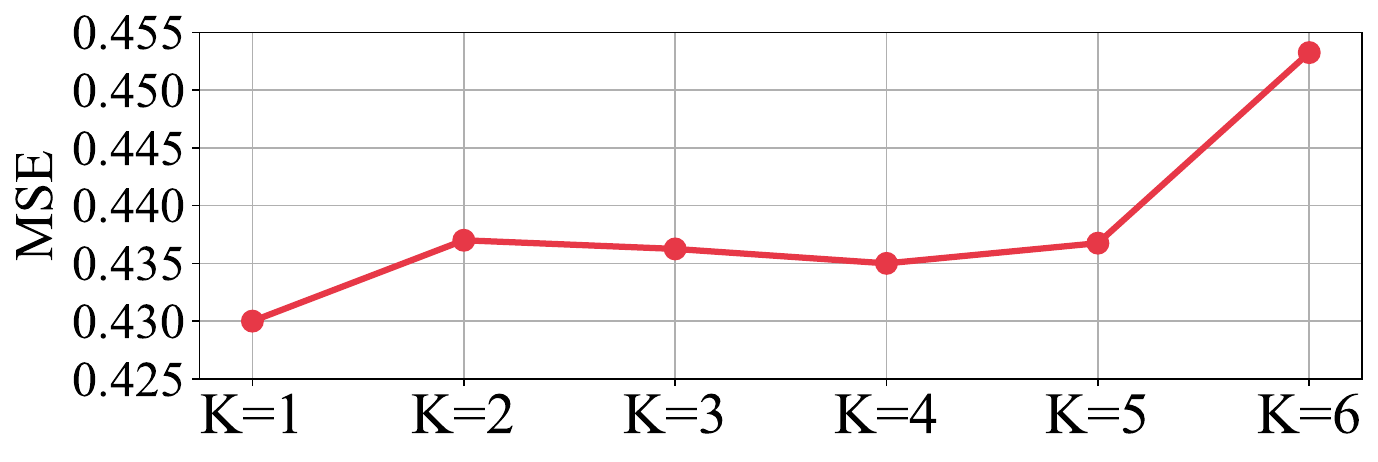}\label{fig:k_etth1_mse}
    }
    \hfill
    \subfigure[ETTm1]
    {\includegraphics[width=0.23\textwidth]{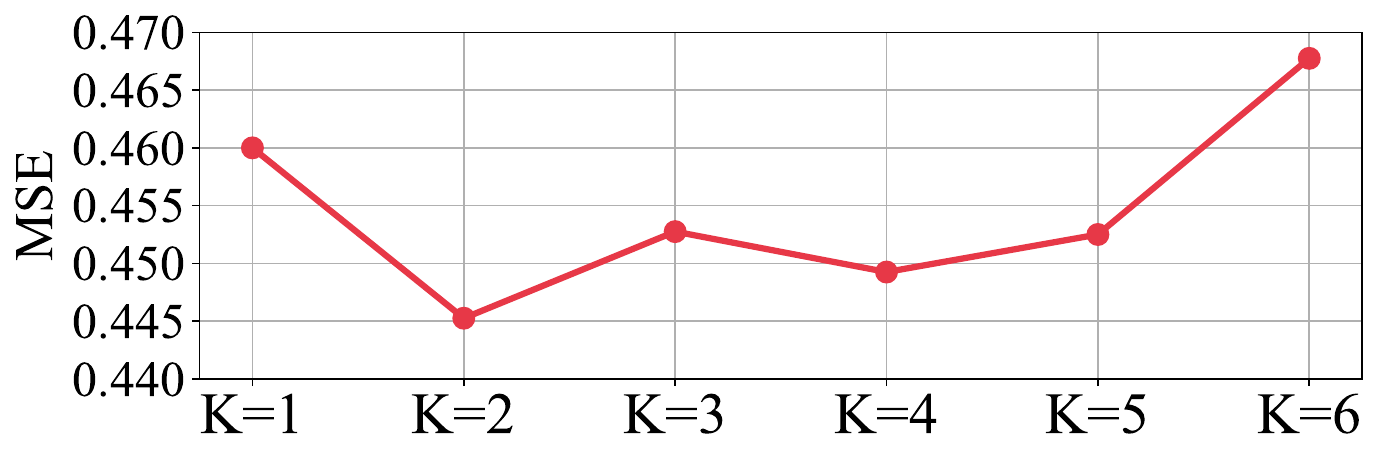}\label{fig:k_ettm1_mse}
    }
    \hfill
    \subfigure[Electricity]
    {\includegraphics[width=0.23\textwidth]{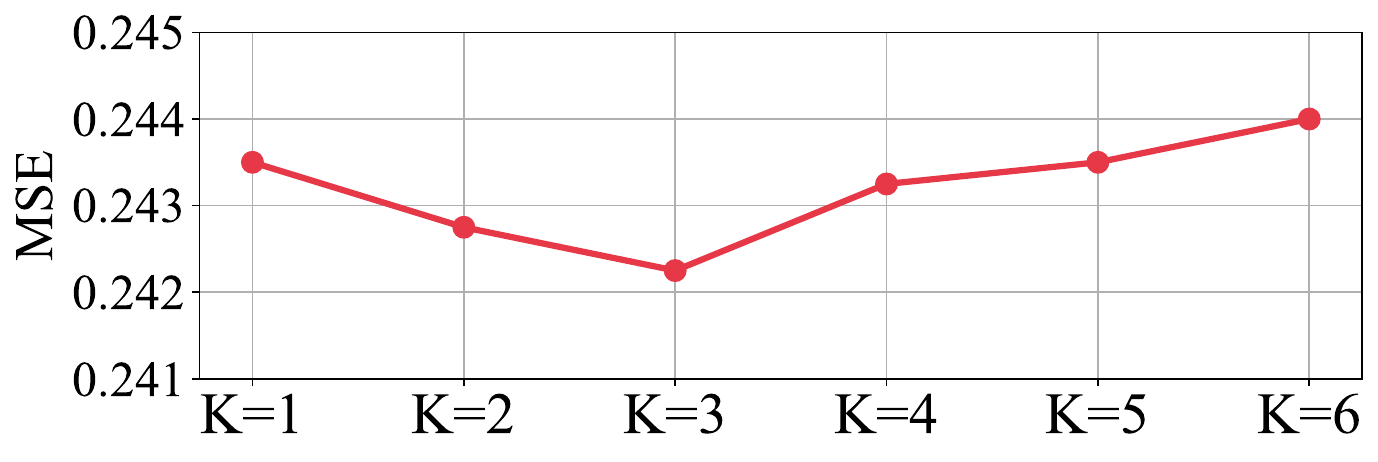}\label{fig:k_ecl_mse}
    }
    \hfill
    \subfigure[Weather]
    {\includegraphics[width=0.23\textwidth]{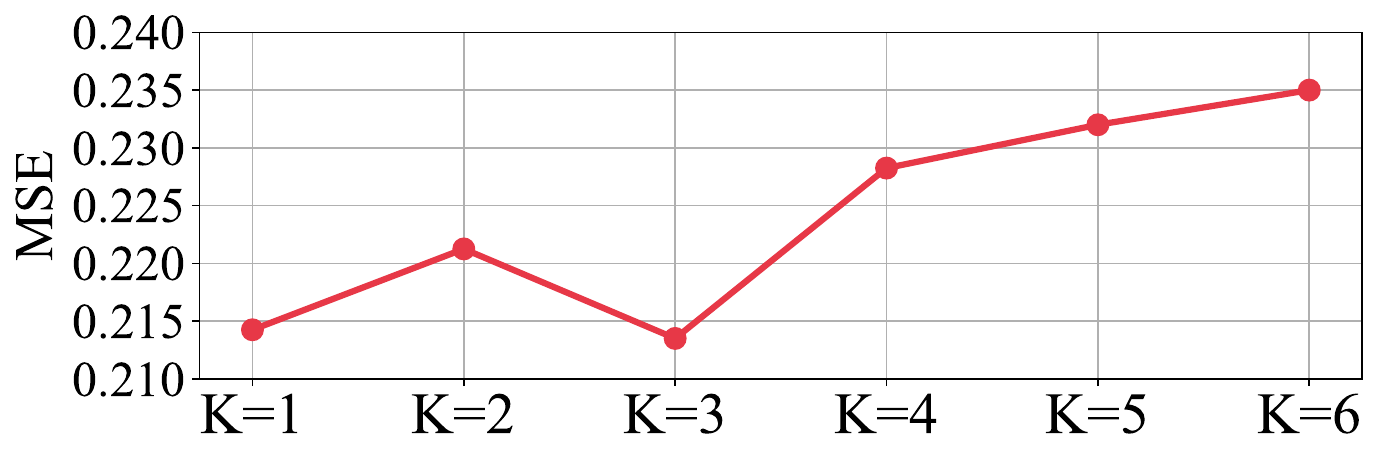}\label{fig:k_weather_mse}
    }
    
    \caption{The sensitive of decomposition levels $K$ on 4 datasets.}
    \label{fig:k_sense}
\end{figure*}

\subsection{Results Compared with TSFMs in MAE}
Figure~\ref{fig:ap_foundation_res} compares the performance of ContexTST with foundational time series models (TimesFM, Chronos-base, Chronos-small, and Moirai) across four datasets (ETTh1, ETTh2, ETTm1, and ETTm2) and prediction horizons. ContexTST consistently achieves the lowest MAE, outperforming other models across all datasets and prediction lengths. It demonstrates significant advantages on datasets like ETTh1 and ETTh2, particularly for longer horizons (336 and 720), where competing models such as Chronos-base and TimesFM show higher errors. Similarly, on ETTm1 and ETTm2, ContexTST maintains superior performance, with smaller average errors even for long-term predictions, where models like Chronos-small and TimesFM struggle. These results highlight ContexTST's strong generalization ability and robustness, making it highly effective for zero-shot inference tasks across diverse domains.

\begin{figure*}[ht]
    \centering
    \includegraphics[width=1.0\textwidth]{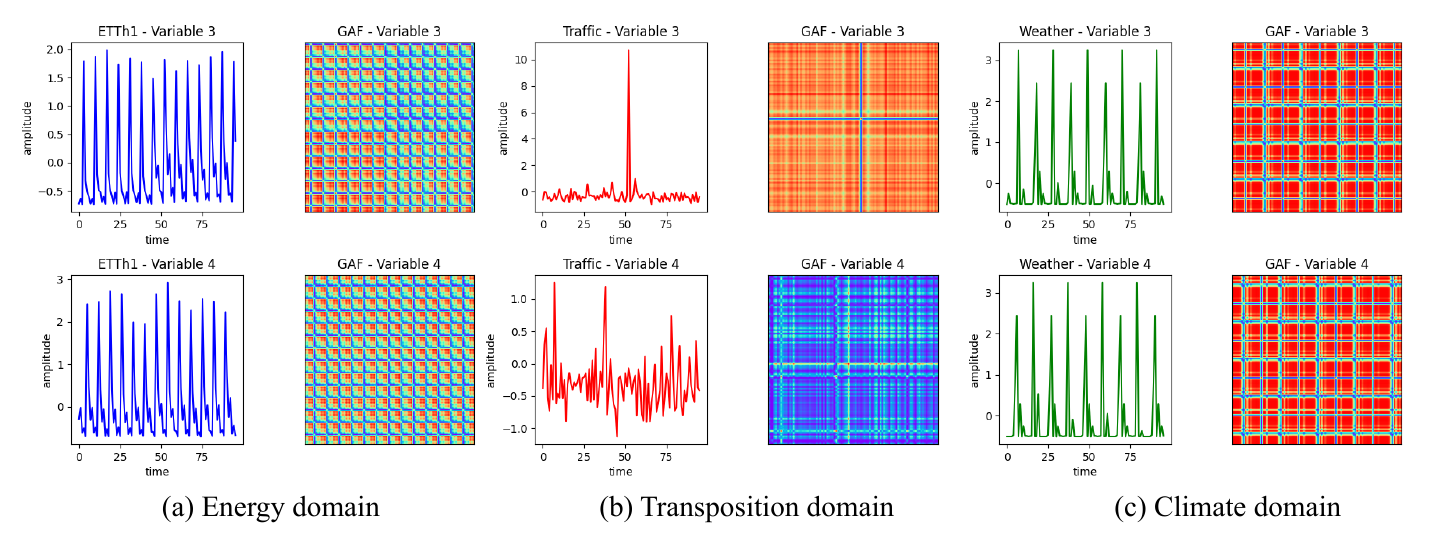}
    \caption{The series self-correlations in different domains.}
    \label{fig:multi_corre}
\end{figure*}

\begin{figure*}[ht]
    \subfigure[Series decomposition and self-correlation in energy domain.]
    {\includegraphics[width=0.9\textwidth]{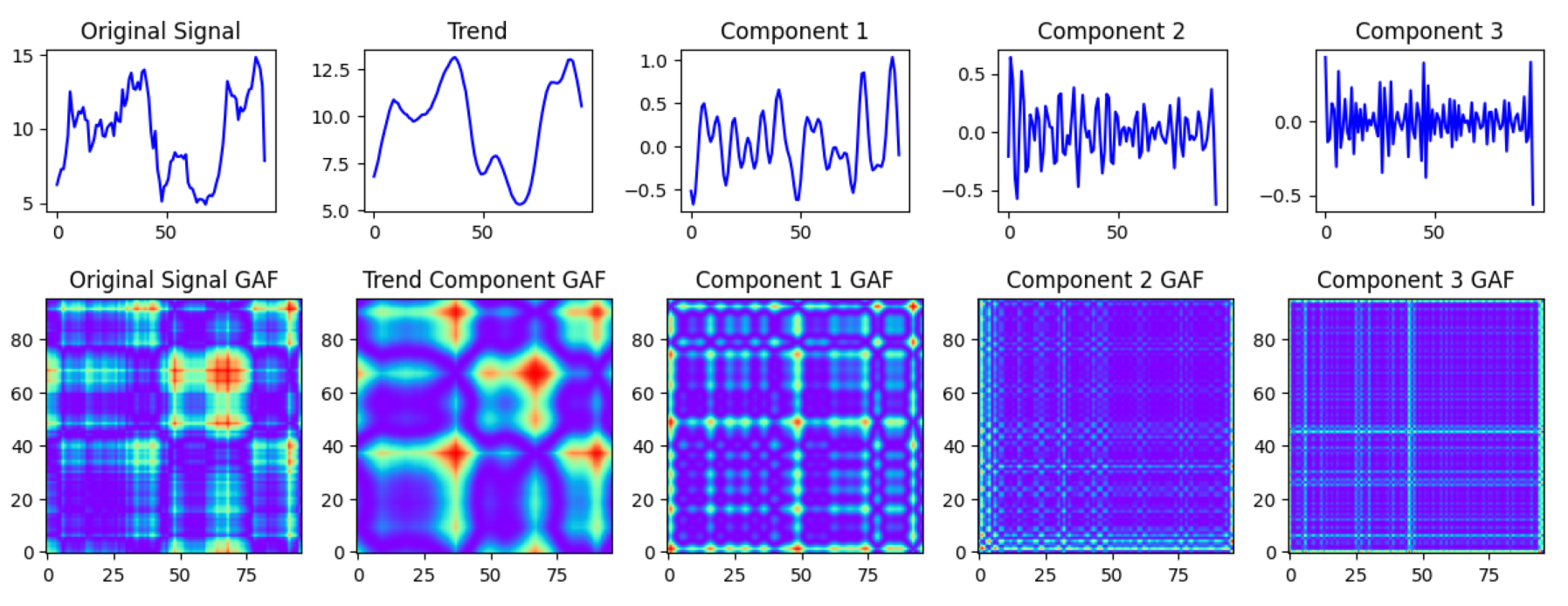}\label{fig:etth1_corre}
    }
    \vspace{-4mm}
    \subfigure[Series decomposition and self-correlation in transposition domain.]
    {\includegraphics[width=0.9\textwidth]{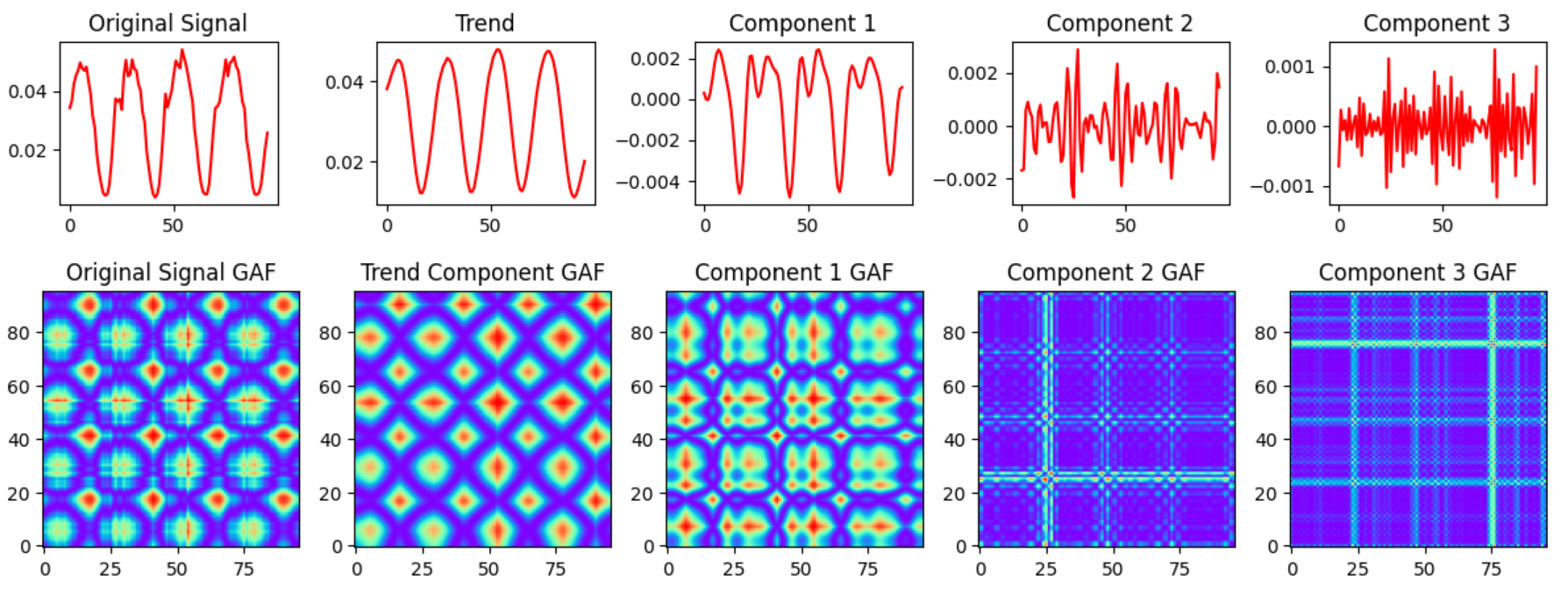}\label{fig:traffic_corre}
    }
    \vspace{-4mm}
    \subfigure[Series decomposition and self-correlation in climate domain.]
    {\includegraphics[width=0.9\textwidth]{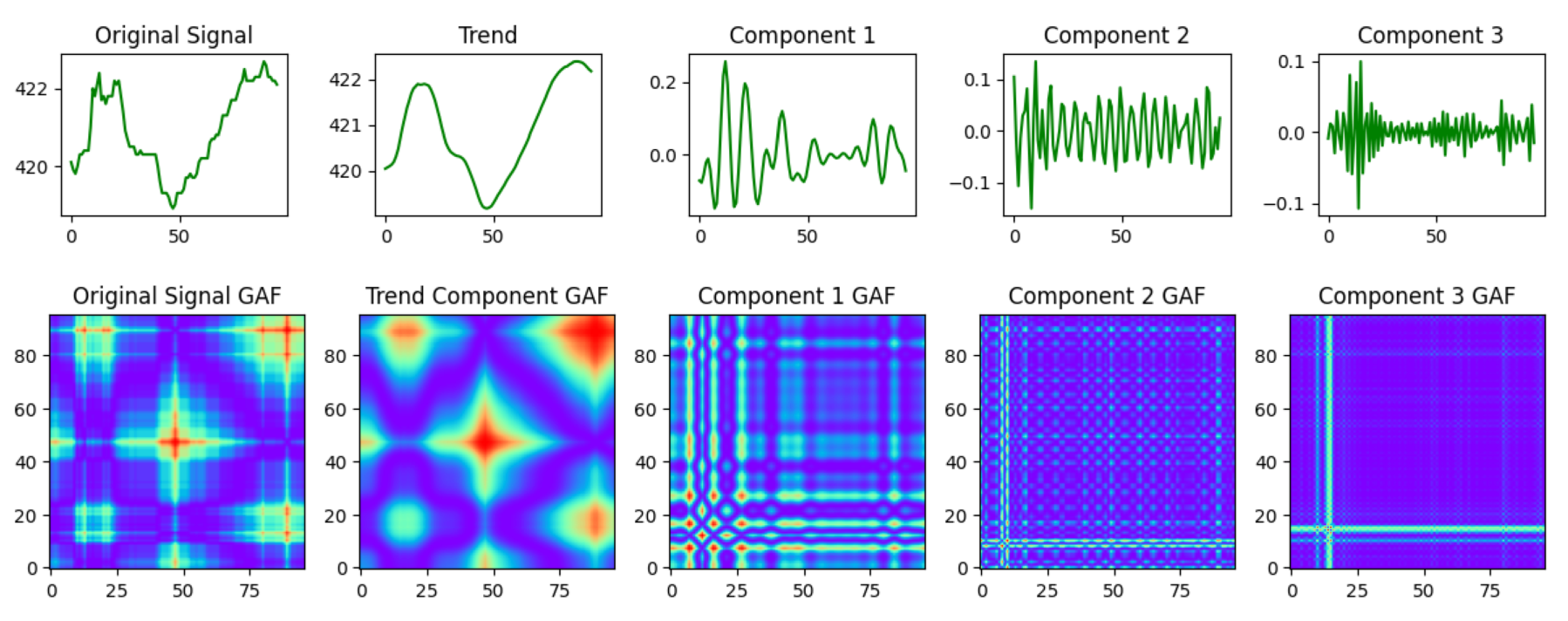}\label{fig:weather_corre}
    }
    \caption{The time series from different domains show distinct self-correlation patterns. While reconstructed series for the spectral decomposed components present similar self-correlation patterns, which inspires our time series coordinator insight.}
    \label{fig:decompose_corre}
    
\end{figure*}

\subsection{Additional Ablation Studies in Cross-Domain Setting}

Figure~\ref{fig:ap_cross_ablation_studies} presents the results of an ablation study analyzing the contributions of the decompose, context, and MoE modules in the proposed ContexTST model under cross-domain transfer scenarios. Across all datasets and prediction horizons, the full ContexTST model consistently achieves the lowest MSE, demonstrating the effectiveness of the combined modules in improving forecasting performance.

The absence of the decompose module results in a noticeable increase in MSE, particularly for longer prediction lengths (336 and 720), as seen in all transfer scenarios (e.g., ETTh1 $\rightarrow$ ETTh2 and Electricity $\rightarrow$ ETTh2). This highlights the module’s critical role in separating temporal patterns and reducing complexity. Similarly, removing the context module leads to a significant performance drop, especially in datasets with more complex temporal patterns, such as ETTm2, indicating its importance in capturing domain-specific temporal dependencies. The removal of the MoE module also negatively impacts performance, but its effect is more prominent for shorter prediction lengths (96 and 192), suggesting its role in enhancing adaptability and fine-grained prediction. Overall, the results confirm that each module contributes uniquely and significantly to the superior performance of ContexTST, with their integration being essential for robust and accurate forecasting in cross-domain settings.

\begin{table*}[ht]
    \centering
    \caption{Experiment configuration of ContexTST.}
    \resizebox{\linewidth}{!}{
    \begin{tabular}{c|ccccccc}
    \toprule
       Dataset  & Decompose levels $K$ & Patch size $P$ & Model dim $D$ & Attention heads $H$ & Transformer blocks $J$ & Num experts $M$ & Activated expert $r$ \\
    \midrule
    ETTh1 & 1 &	24 & 256 & 2 & 1 & 4 & 2 \\
    ETTh2 &	2 &	24 &	256 &	2&	1&	4&	2\\
    ETTm1 &	2 &	24 &	256 &	4&	2&	4&	2\\
    ETTm2 &	2 &	24 &	256 &	4&	2&	4&	2\\
    Electricity &	3 &	24 &	512&	8&	4&	4&	2\\
    Weather &	3 &	24 &	512 &	8&	4&	4&	2\\
    Traffic &	4 &	24 &	512 &	8&	4&	4&	2\\
    \bottomrule
    \end{tabular}}
    \label{tab:model_parameters}
\end{table*}

\begin{table*}[ht]
    \centering
    \caption{The computation memory. The batch size is 1 and the prediction horizon is set to 96.}
    % \resizebox{\linewidth}{!}{
    \begin{tabular}{c|c|cccccc}
    \toprule
     Metric & Dataset  & ContexTST & TimeMixer & TimeXer & iTransformer & PatchTST & Autoformer \\
    \midrule
       \multirow{4}{*}{NPARAMS (MB)} & ETTm1  & 8.952 & 0.288 &	6.798	& 0.855 & 	14.311 &	40.191 \\
        & Weather	& 13.988 &	0.398 &	13.806 &	18.440 &	26.336 &	40.465 \\
       & Traffic	& 20.975 &	0.461 &	33.315 &	24.460 &	14.325 &	56.894 \\
       & Electricity &	27.078 &	0.408 &	42.289 &	18.440 &	26.336 &	46.325 \\
    \midrule
        \multirow{4}{*}{Max CPU Mem. (GB)} & ETTm1 &	0.026 &	0.012 &	0.022 &	0.009 &	0.036 &	0.112 \\
        & Weather &	0.052 &	1.112 &	0.124 &	0.029 &	 0.062 &	0.113 \\
        & Traffic &	1.053 &	0.712 &	0.694 &	0.322 &	0.539 &	0.136 \\
        & Electricity &	1.289 &	0.141 & 0.326 &	0.072 &	0.350 &	0.121\\
    \bottomrule
    \end{tabular}
    \label{tab:model_mem}
\end{table*}

\section{Model Efficiency Analysis}
\label{app:model_mem}

Table~\ref{tab:model_mem} presents the parameter memory and the max GPU memory cost of various models under a batch size of 1 and a prediction horizon of 96. Our proposed method, ContexTST, demonstrates competitive memory efficiency, particularly for smaller datasets such as ETTm1 (8.952 MB) and Electricity (27.078 MB). It significantly outperforms other models like Autoformer and TimeXer in terms of model parameter size. For the max GPU memory cost, ContexTST has a similar footprint as baselines during model forwarding.

Notably, ContexTST strikes a balance between efficiency and scalability, achieving lower memory usage than most baseline methods for larger datasets such as Traffic. These results highlight the practicality of ContexTST for deployment in environments with limited computational resources, while still maintaining competitive performance across diverse datasets.

\section{Parameter Sensitivity Analysis}
\label{app:param_analysis}

Figure~\ref{fig:k_sense} illustrates the sensitivity of the model to the decomposition level K across four datasets: ETTh1, ETTm1, Electricity, and Weather. The analysis highlights the impact of varying the decomposition levels on the model's performance, as measured by the MSE.
From the results, the performance of the model exhibits clear dataset-specific trends. For instance, in ETTh1 and ETTm1, the model achieves the lowest MSE at intermediate decomposition levels (e.g., K=2 or K=3), while performance deteriorates significantly when K increases to 5. This suggests that excessive decomposition may overcomplicate the temporal patterns and hinder the model's ability to capture dependencies effectively. Overall, the results suggest that selecting an appropriate decomposition level is critical to achieving optimal performance, with K=2 or K=3 generally providing the best results. These findings emphasize the importance of tuning K based on the characteristics of the dataset to balance model complexity and forecasting accuracy. For the detailed model configurations, please refer to Table~\ref{tab:model_parameters}.

\section{Visualization Analysis}
\label{app:viz_analysis}

Figure~\ref{fig:multi_corre} demonstrates the self-correlation characteristics of time series data across three distinct domains: Energy, Transportation, and Climate. For each domain, the raw time series and their corresponding Gramian Angular Field (GAF)~\cite{wang2015encoding} representations are shown side by side, highlighting the transformation from temporal data to structured visual form. The purpose of this analysis is to showcase the diverse and complex self-correlation patterns present in time series from different domains, which pose significant challenges for cross-domain transfer in time series analysis.

To construct the GAF, each time series is first normalized to the range [-1, 1] with $\tilde{x}_t=\frac{2\cdot(x_{t,i}-\min(x_{t}))}{\max(x_{t})-\min(x_{t})}-1$. The normalized values are then mapped to angular via $\phi_i = \arccos (\tilde{x}_{t})$, and the GAF matrix is computed as:
\begin{align}
    \text{GAF}(i,j) = \cos(\phi_i + \phi_j)=\tilde{x}_{t,i}\cdot \tilde{x}_{t,j} - \sqrt{1-\tilde{x}_{t,i}^{2}} \cdot \sqrt{1-\tilde{x}_{t,j}^{2}}.
\end{align}
This transformation encodes temporal dependencies between all data points into a symmetric matrix, which is visualized as a heatmap to reflect the self-correlation structure of the time series.

The results reveal significant differences in self-correlation patterns across domains. In the Energy domain, the time series exhibits strong periodicity, with highly regular oscillations leading to repetitive and structured diagonal patterns in the GAF. In contrast, the Transportation domain shows irregular peaks and abrupt changes, resulting in more fragmented and localized GAF patterns that indicate weaker and less consistent temporal dependencies. The Climate domain, while also characterized by periodic trends, displays more complex and grid-like GAF patterns, reflecting the seasonal and cyclical variability inherent in weather data. These differences suggest that time series from different domains exhibit distinct temporal dynamics, and their corresponding GAF representations encode diverse self-correlation structures. This diversity poses challenges for cross-domain transfer, as models trained on one domain may struggle to adapt to the complex and domain-specific self-correlation patterns of another.

Figure~\ref{fig:decompose_corre} demonstrates the effectiveness of our proposed time series coordinate, which decomposes original signals into simpler spectral components and visualizes their corresponding GAF patterns. While the original signals exhibit distinct domain-specific self-correlation structures, their decomposed components, such as trends and oscillatory behaviors, display more consistent patterns across domains. For instance, the trend components in all domains show smooth, low-frequency behavior with regular GAF patterns, while higher-frequency components (e.g., Component 1 and Component 2) reveal repetitive, localized self-correlations. This suggests that spectral decomposition isolates fundamental features of time series, enabling a unified representation in a shared coordinate space and simplifying domain-specific complexities for cross-domain analysis.

\end{document}